\pdfoutput=1

\documentclass[11pt]{article}

\usepackage{amsmath,amsfonts,bm}

\def\eqref#1{equation~\ref{#1}}

\def\1{\bm{1}}

\DeclareMathAlphabet{\mathsfit}{\encodingdefault}{\sfdefault}{m}{sl}
\SetMathAlphabet{\mathsfit}{bold}{\encodingdefault}{\sfdefault}{bx}{n}

\newcommand{\softmax}{\mathrm{softmax}}

\usepackage[final]{acl}

\usepackage{times}
\usepackage{latexsym}
\usepackage[T1]{fontenc}
\usepackage[utf8]{inputenc}
\usepackage{microtype}
\usepackage{inconsolata}
\usepackage{graphicx}
\usepackage{booktabs}
\usepackage{amsmath}
\usepackage{tcolorbox}
\usepackage{amssymb}
\usepackage{multirow}
\usepackage{enumitem}
\usepackage{xspace}
\usepackage{colortbl}
\usepackage{arydshln}
\usepackage{pifont}
\usepackage{amsthm}

\definecolor{myblue}{rgb}{0.0, 0.0, 0.8}
\definecolor{myorange}{rgb}{1.0, 0.647, 0.0}
\usepackage{subcaption}

\newtheorem{definition}{Definition}

\newtheorem{theorem}{Theorem}
\newtheorem{lemma}[theorem]{Lemma}
\newtheorem{corollary}[theorem]{Corollary}

\newcommand{\Mat}{\boldsymbol}

\newcommand{\real}{\mathbb{R}}

\newcommand{\FT}{\mathcal{F}}

\DeclareMathOperator{\diag}{diag}

\newcommand{\DC}[1]{\mathcal{DC}\left[ {#1} \right]}
\newcommand{\HC}[1]{\mathcal{HC}\left[ {#1} \right]}

\DeclareMathOperator{\SA}{SA}

\DeclareMathOperator{\MSA}{MSA}

\newcommand{\up}{\ding{115}}
\newcommand{\down}{\ding{116}}

\title{Length-Induced Embedding Collapse in PLM-based Models}

\author{
    Yuqi Zhou,
    Sunhao Dai,
    Zhanshuo Cao,
    Xiao Zhang,
    Jun Xu\thanks{~~Corresponding author.}\\
    Gaoling School of Artificial Intelligence, Renmin University of China, Beijing, China \\
    \texttt{\{yuqizhou, sunhaodai, caozhanshuo117, zhangx89, junxu\}@ruc.edu.cn} 
}

\begin{document}
\maketitle
\begin{abstract}
Text embeddings from PLM-based models enable a wide range of applications, yet their performance often degrades on longer texts. In this paper, we introduce a phenomenon we call \textbf{Length Collapse}, where embeddings of longer texts tend to cluster together. This clustering results in a distributional inconsistency between the embeddings of short and long texts. We further investigate how these differences contribute to the performance decline observed with longer texts across various downstream tasks. Through a rigorous theoretical analysis of the self-attention mechanism, which acts as a low-pass filter in PLM-based models, we demonstrate that as text length increases, the strength of low-pass filtering intensifies, causing embeddings to retain more low-frequency components. As a result, input token features become more similar, leading to clustering and ultimately the collapse of embeddings for longer texts. To address this issue, we propose a simple method, TempScale, which mitigates the Length Collapse phenomenon. By narrowing the gap in low-pass filtering rates between long and short texts, TempScale ensures more consistent embeddings across different text lengths. This approach leads to performance improvements of \textbf{0.94\%} on MTEB and \textbf{1.10\%} on LongEmbed, which focuses specifically on long-context retrieval, providing strong evidence for the validity of our analysis. The source code is available at \textcolor{blue}{\url{https://github.com/Yuqi-Zhou/Length_Collapse}}.
\end{abstract}

\section{Introduction}\label{sec:introduction}
Text embeddings—dense vectors that capture the semantic information of texts—are essential for many NLP applications, including text analysis~\cite{aggarwal2012survey,angelov2020top2vec}, question answering~\cite{tan2023evaluation,xu2024search}, web search~\cite{zhao2022dense,yates2021pretrained}, and retrieval-augmented generation~\cite{gao2023retrieval,fan2024survey}. Typically, embeddings are generated by pre-trained language models (PLMs), which produce fixed-dimensional vectors regardless of text length. In practice, we expect PLMs to perform consistently across texts of varying lengths in downstream tasks.

\begin{figure*}[t]
\vspace{-0.4cm}
     \centering
      \begin{subfigure}[b]{0.3\textwidth}
         \centering
         \includegraphics[width=\textwidth]{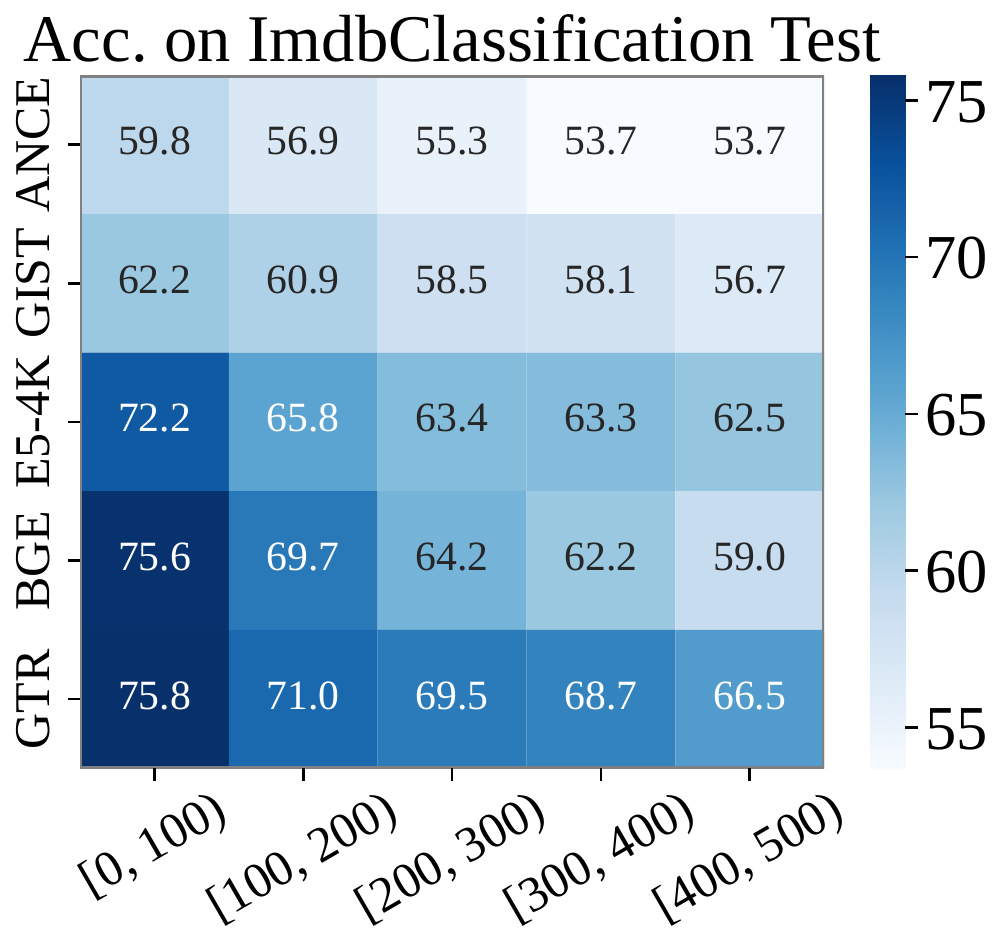}
         \caption{}
         \label{fig:intro_performance}
     \end{subfigure}
     \hfill
     \begin{subfigure}[b]{0.31\textwidth}
         \centering
         \includegraphics[width=\textwidth]{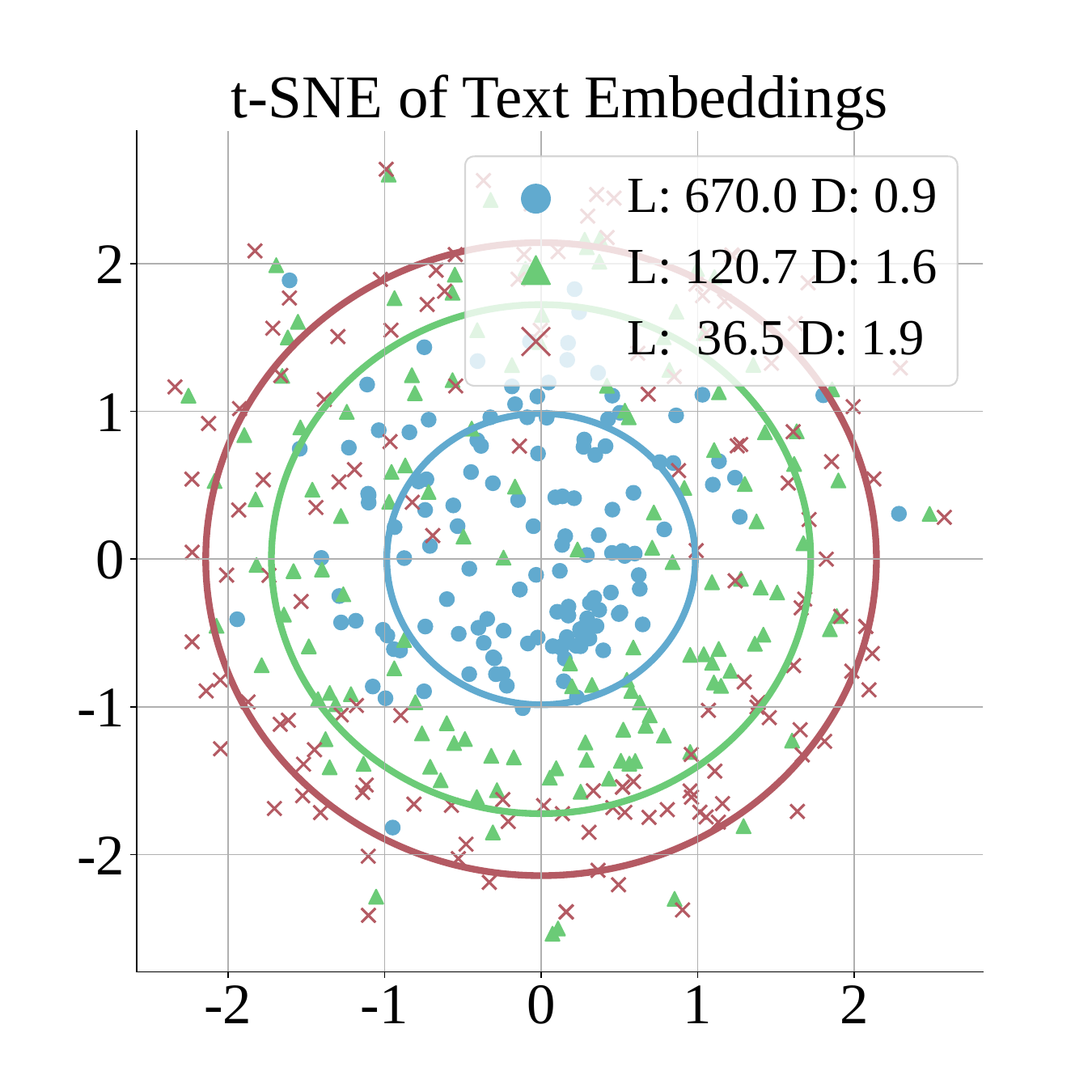}
         \caption{}
         \label{fig:intro_tsnet}
     \end{subfigure}
     \hfill
     \begin{subfigure}[b]{0.29\textwidth}
         \centering
         \includegraphics[width=\textwidth]{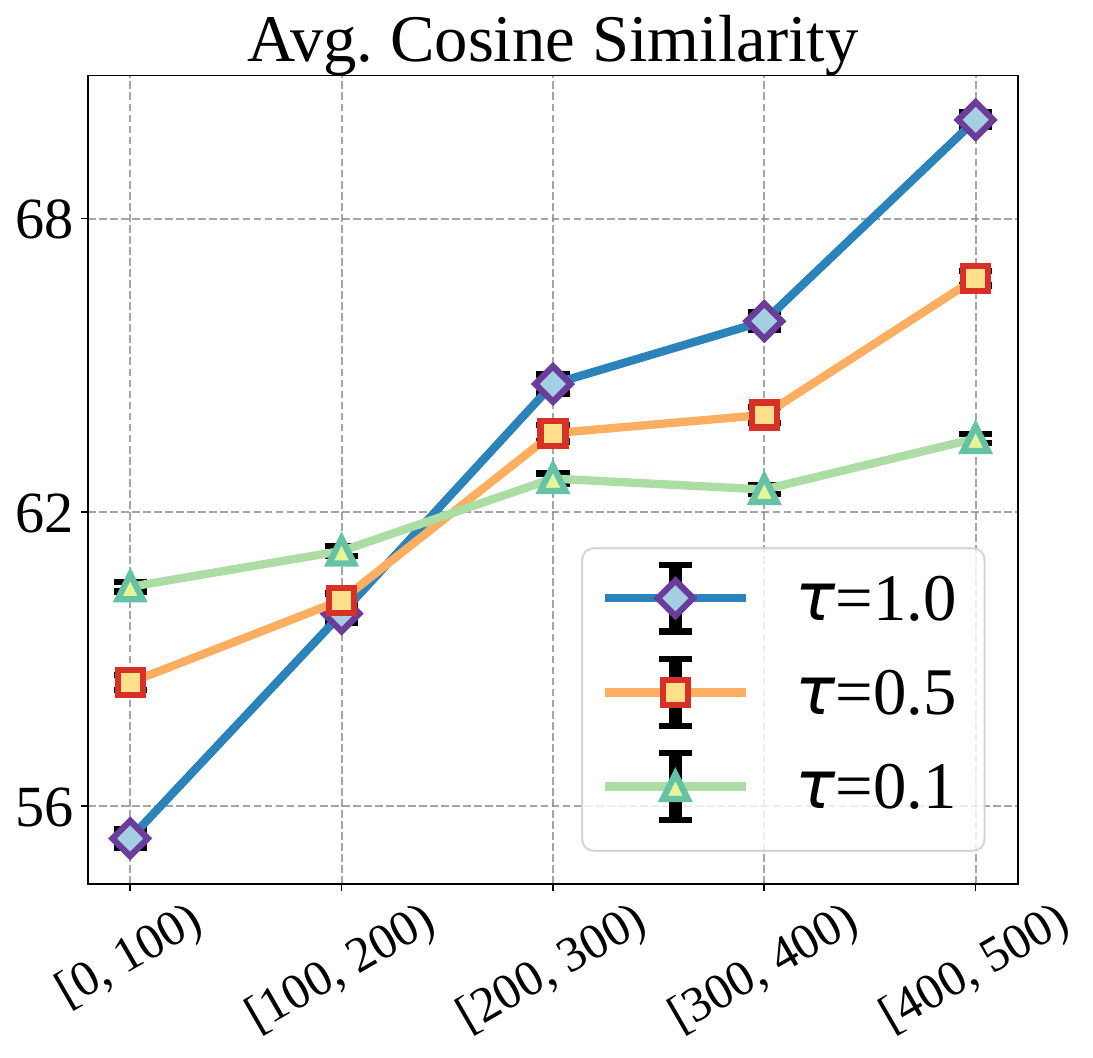}
         \caption{}
         \label{fig:intro_cosine}
     \end{subfigure} 
     \caption{\textbf{(a)} Performance of embedding models on IMDb classification across length intervals [0, 100) to [400, 500). The bluer \textcolor[HTML]{61AACF}{\rule{0.7em}{0.7em}} a cell, the higher the classification accuracy. \textbf{(b)} t-SNE visualization of embeddings from the BGE on NFCorpus dataset, with \textcolor[HTML]{61AACF}{\ding{108}} for the original dataset and \textcolor[HTML]{6BCB77}{\ding{115}} and \textcolor[HTML]{B45A63}{\text{×}} for LLM-summarized versions, retaining semantic meaning with varying lengths. \textbf{L} indicates average text length, and \textbf{D} denotes mean distance to the origin. \textbf{(c)} Mean pairwise cosine similarity of embeddings from the BGE model across text length intervals on the corpus from NFCorpus, with an X-axis for length intervals and a Y-axis for average pairwise similarity.}
     \label{fig:intro_fig}
     \vspace{-5pt}
\end{figure*}

Unfortunately, \textit{popular PLM-based embedding models perform poorly on longer texts}. As shown in Figure~\ref{fig:intro_performance}, we evaluate mainstream models on the IMDB classification task from the Massive Text Embedding Benchmark (MTEB)\cite{muennighoff2023mteb}, using test sets grouped by text length. The results show that models with varying capabilities degrade as text length increases. For example, the BGE~\cite{xiao2023c} model’s accuracy drops from 75.6\% in the [0, 100) token range to 59.0\% in the [400, 500) range, a 16.6\% point decline.

We attribute this performance degradation to a systematic tendency in embedding models: as the text length $n$ increases, the cosine similarity between embeddings of different texts also increases—a phenomenon we refer to as \textbf{Length Collapse.} To verify this, we conduct controlled experiments shown in Figures~\ref{fig:intro_tsnet} and~\ref{fig:intro_cosine}. Figure~\ref{fig:intro_tsnet} shows that embeddings of longer texts cluster more densely near the origin in the reduced embedding space, indicating a collapse that reduces variance. Figure~\ref{fig:intro_cosine} further shows that embeddings of longer texts have higher cosine similarity to each other, leading to smaller differences. This collapse causes distributional inconsistency between embeddings of different lengths. Furthermore, we analyze how this distributional inconsistency in embeddings of different text lengths leads to performance degradation in downstream tasks such as classification, retrieval, and STS, particularly affecting the performance of longer texts.

To study how text length affects embedding distributions, we analyze the self-attention mechanism in Fourier space~\cite{wanganti} (\S~\ref{sec:theorem_analysis}). Building on the findings that cascading self-attention blocks act as repeated low-pass filters, we prove that the attenuation rate of this filter is proportional to the largest singular value $\sigma_{a}$ of the high-frequency components in the self-attention matrix. Assuming Gaussian-distributed input keys and query tokens, we show that $\sigma_{a}$ decreases with increasing text length, causing longer texts to retain more of the Direct Component (DC) in token signals. This results in embeddings collapsing to a narrow space, as seen in Figure~\ref{fig:intro_tsnet}.

To validate our theoretical analysis, we propose a method, Temperature Scaling (\textbf{TempScale}), to mitigate Length Collapse. TempScale manipulates the attention map by dividing the scores by a parameter called temperature (smaller than 1) before applying the $\softmax(\cdot)$ operator. This reduces the gap in $\sigma_{a}$ between long and short texts in the self-attention matrix, minimizing the distributional gap between their embeddings. As shown in Figure~\ref{fig:intro_cosine}, a smaller temperature enables embeddings to exhibit lower pairwise cosine similarity, resulting in a more even distribution and alleviating Length Collapse. Additionally, TempScale improves performance by \textbf{0.94\%} and \textbf{1.10\%} in MTEB and LongEmbed~\cite{zhu2024longembed}, respectively. All results confirm the validity of our analysis.

Our contributions are as follows: ($1$) We are the first to identify \textbf{Length Collapse}, where embeddings of longer texts cluster together, and explain how this uneven distribution degrades overall performance, especially for long texts. ($2$) We provide a rigorous theoretical analysis in the spectral domain, demonstrating that Length Collapse occurs due to the increasing low-pass filtering strength of self-attention as sequence length grows, causing token signals to retain only their DC component. ($3$) To validate this theoretical analysis, we propose a simple method, TempScale, to mitigate Length Collapse. Empirical results show a \textbf{0.94\%} performance improvement across the MTEB and a \textbf{1.10\%} improvement on LongEmbed, validating the correctness of our theoretical analysis.

\section{Length Collapse}\label{sec:length_collapse}
In this section, we define the problem and introduce our notations. We then present the Transformer structure in PLM-based models and briefly explain the Fourier transform from~\cite{wanganti}. Using this framework, we show that the attention mechanism functions as a low-pass filter, with longer input sequences amplifying this effect, resulting in increasingly similar representations.

\subsection{Preliminaries and Background} 
\textbf{Notations.} Let $\bm{X} \in \mathbb{R}^{n \times d}$ denote the input feature matrix, where $n$ is the number of tokens and $d$ is the embedding dimension. Let $\bm{x}_i \in \mathbb{R}^d$ represent the vector for the $i$-th token, and $\bm{z}_j \in \mathbb{R}^n$ represent the token sequence for the $j$-th dimension, where $i \in \{1, \dots, n\}$ and $j \in \{1, \dots, d\}$.

\textbf{Transformer Architecture.} Most modern embedding models~\cite{chen2024bge, xiong2021approximate} use a bidirectional transformer architecture with attention mechanisms. These models typically consist of three components: an embedding layer, a stack of transformer encoder blocks with Multi-Head Self-Attention (MSA) and Feed-Forward Networks (FFN), and a pooling layer to generate the final embedding. The Self-Attention (SA) module encodes each token by aggregating information from other tokens based on attention scores, as defined by the equation~\cite{vaswani2017attention}: \begin{align*} \SA (\bm{X}) = \softmax\left(\frac{\Mat{X} \Mat{W}_{Q} (\Mat{X} \Mat{W}_{K})^T}{\sqrt{d}}\right) \Mat{X} \Mat{W}_{V}, \end{align*} where $\Mat{W}_K$, $\Mat{W}_Q$, and $\Mat{W}_V$ are the key, query, and value weight matrices. The function $\text{softmax}(\cdot)$ normalizes the attention scores. Multi-Head Self-Attention (MSA) aggregates the outputs of multiple SA heads, projected back to the hidden dimension: \begin{align*} \MSA(\Mat{X}) = \begin{bmatrix} \SA_1(\Mat{X}) & \cdots & \SA_H(\Mat{X}) \end{bmatrix} \Mat{W}_O, \end{align*} where $H$ is the number of heads, and $\bm{W}_O$ projects the combined outputs to the hidden dimension.

\textbf{Fourier Analysis.} 
Inspired by the recent work~\cite{wanganti}, which reveals that self-attention in Vision Transformers acts as a low-pass filter and causes feature degradation when scaling up depth, we employ Fourier analysis to further explore this phenomenon. We use the Fourier transform as the primary analytic tool, following~\cite {wanganti}. Let $\mathcal{F}: \mathbb{R}^{n} \to \mathbb{C}^{n}$ represent the Discrete Fourier Transform (DFT), with its inverse $\mathcal{F}^{-1}: \mathbb{C}^{n} \to \mathbb{R}^{n}$. Applying $\mathcal{F}$ to a token sequence $\bm{z}$ corresponds to multiplying by a DFT matrix, where the $k$-th row corresponds to the Fourier basis at frequency $\bm{f}k=[e^{2\pi j(k-1)\cdot 0}, \ldots, e^{2\pi j(k-1)\cdot(n-1)}]^{\top} / \sqrt{n} \in \mathbb{R}^{n}$, with $j$ as the imaginary unit. Let $\bm{\tilde{z}} = \mathcal{F} \bm{z}$ denote the spectrum of $\bm{z}$, with $\bm{\tilde{z}}_{dc} \in \mathbb{C}$ and $\bm{\tilde{z}}_{hc} \in \mathbb{C}^{n-1}$ as the DC and high-frequency components, respectively. We define the Direct-Current (DC) component as $\mathcal{DC}[\bm{z}] = \bm{\tilde{z}}_{dc} \bm{f}_1 \in \mathbb{C}^n$, and the high-frequency component as $\mathcal{HC}[\bm{z}] = [\bm{f}_2, \dots, \bm{f}n] \bm{\tilde{z}}{hc} \in \mathbb{C}^n$. In signal processing, a low-pass filter preserves low-frequency components while attenuating high frequencies. In this paper, we define a low-pass filter that retains only the DC component $\DC{\Mat{z}}$, while diminishing the high-frequency components $\HC{\Mat{z}}$. A precise definition is given in Definition \ref{dfn:low_pass_filter}.

\begin{definition} \label{dfn:low_pass_filter} Let $ f: \mathbb{R}^n \rightarrow \mathbb{R}^n $ be an endomorphism with $f^t$ denoting $f$ applied $t$ times. The function $f$ acts as a low-pass filter if and only if $\lim_{t \rightarrow \infty} \frac{\lVert \HC{f^t(\mathbf{z})} \rVert_2}{\lVert \DC{f^t(\mathbf{z})} \rVert_2} = 0$ for all $\Mat{z} \in \mathbb{R}^n$. \end{definition}
For additional details, refer to Appendix \ref{sec:background_fourier}.

\subsection{Theoretical Analysis on Length Collapse}~\label{sec:theorem_analysis}
\textbf{Overview.} This subsection demonstrates that increasing text length $n$ accelerates low-pass filtering in the attention matrix, making text embeddings for longer texts more similar. We justify this by analyzing self-attention’s spectral-domain effect. Building on \textbf{Lemma}\ref{thm:sa_as_low_pass}, we show that the largest singular value $\sigma_a$ of $\HC{\Mat{A}}$ influences the filtering rate, with a smaller $\sigma_a$ indicating greater high-frequency loss (\textbf{Theorem}\ref{thm:lambda2_sa_rate}). Our analysis further reveals that longer texts result in smaller $\sigma_a$ values, causing longer text embeddings to lose feature expressiveness (\textbf{Theorem}\ref{thm:length_rate}). Consequently, we infer that longer texts yield more similar representations (\textbf{Corollary\ref{thm:length_collapse}}), leading to \textbf{Length Collapse}.

Formally, the following lemma demonstrates that the attention matrix generated by a softmax function acts as a low-pass filter, independent of the specific token features or context window.

\begin{lemma} \label{thm:sa_as_low_pass}
(Attention Matrix is A Low-pass Filter) 
Let $\bm{A} = \softmax(\bm{P})$, where $ \bm{P} \in \mathbb{R}^{n \times n}$. Then $\bm{A}$ must be a low-pass filter. For all $\bm{z} \in \mathbb{R}^n$,
\begin{align*}
    \lim_{t \to \infty} \frac{\|\mathcal{HC} [\bm{A}^t \bm{z}]\|_2}{\|\mathcal{DC} [\bm{A}^t \bm{z}]\|_2} = 0.
\end{align*}
\end{lemma}

Lemma \ref{thm:sa_as_low_pass} follows from the Perron-Frobenius theorem~\cite{meyer2000matrix}. Since all elements of the self-attention matrix are positive and each row sums to $1$, the largest eigenvalue is $1$. Repeated application of the self-attention matrix represents the forward process of the embedding model, and as the number of layers increases, the output retains only the DC component. A more detailed proof can be found in Theorem $1$ of \citet{wanganti}.

Understanding that self-attention matrices act as low-pass filters, we are interested in the extent to which an SA layer suppresses high-frequency components. Additionally, we provide a filter rate to illustrate the speed at which these high-frequency components are eliminated.
\begin{theorem} \label{thm:lambda2_sa_rate}
(Filter Rate of SA) 
Let $\sigma_a$ be the largest singular value of $\HC{\Mat{A}}$, and $\SA(\Mat{X}) = \Mat{A}\Mat{X}\Mat{W}_V$ the self-attention output. We have:
\begin{align} \label{eqn:theorem2}
\lVert \HC{\SA(\Mat{X})} \rVert_F \le \sigma_a \lVert\Mat{W}_V\rVert_2 \lVert \HC{\Mat{X}} \rVert_F.
\end{align}
\end{theorem}

Theorem \ref{thm:lambda2_sa_rate} suggests the high-frequency intensity ratio to the pre- and post- attention aggregation is upper bounded by $\sigma_a \lVert\Mat{W}_V\rVert_2$. When $\sigma_a \lVert\Mat{W}_V\rVert_2 < 1$, $\HC{\Mat{X}}$ converges to zero exponentially. We further present Figure~\ref{fig:theorem2} in Appendix~\ref{sec:details_figure_theorem2} to justify our results, showing that the upper bound is consistent with the trend observed in the actual values. The proof of Theorem \ref{thm:lambda2_sa_rate} can be found in Appendix~\ref{prf:lambda2_sa_rate}. 

Based on the fact that a lower $\sigma_s$ leads to a higher filter-pass rate, we prove in the following theorem that $\sigma_s$ decreases as the input length $n$ increases.

\begin{theorem} \label{thm:length_rate}
(Filter Rate of Different Input Length $n$)
Let $\Mat{XW}_{Q}$ and $\Mat{XW}_{K}$ be a Gaussian matrix, where elements $q_{ij} \sim \mathcal{N}(0, \sigma_q^2)$ and $k_{ij} \sim \mathcal{N}(0, \sigma_k^2), \forall i, j.$ Let $p_{ij}=\Mat{q}_{i}^{\top} \Mat{k}_{j} / \sqrt{d}$ the attention score of pair $i,j$, whose variance can be expressed as $\sigma_s^2=\sigma_q^2 \sigma_k^2+C_{cross}$, where $C_{cross}$ is the cross-covariance of the squared queries and keys~\cite{goodman1960exact}. Then we have
\begin{align}
    \label{ieqn:rate_with_len} \sigma_{a} \leq \sqrt{\frac{n}{2\sqrt{1+\frac{1}{e^{2\sigma^{2}_{s}}}}(n-1)^{\frac{3}{2}}+1}},
\end{align}
where $\sigma_{a}$ decreases with $n$ increasing.
\end{theorem}

The proof of Theorem \ref{thm:length_rate} is in Appendix~\ref{prf:length_rate}. Building on the work of \citet{fenton1960sum, nahshanlinear} on log-normal variables, Theorem \ref{thm:length_rate} shows that as input length $n$ increases, $\sigma_{a}$ decreases, suppressing high-frequency information and reducing feature expressiveness due to the self-attention matrix's low-pass filtering effect. To validate this, we sample texts of varying lengths and plot the $\sigma_a$ values from the final layer's attention matrix, as shown in Figure~\ref{fig:more_results_on_theorem3} in Appendix~\ref{sec:details_figure_theorem3}. The results confirm that $\sigma_a$ decreases with text length, leading to a higher filtering rate. To facilitate further analysis, we define the temperature of the SA defined in \citet{nahshanlinear} as:
\begin{align}
    \tau_{s} = \frac{1}{\sigma_{s}} = \frac{1}{\sqrt{\sigma_q^2 \sigma_k^2+C_{cross}}} . \label{eqn:tau_sigma}
\end{align}

Then denote $\tilde{p}_{ij}=p_{ij}/ \sigma_s$ and each element in attention matrix $\Mat{A}$ can be rewritten as follows:
\begin{align} 
    \Mat{A}_{ij} = \frac{e^{\tilde{p}_{ij}/ \tau_{s}}}{\sum_{k=1}^n e^{\tilde{p}_{ik}/ \tau_{s}}}, \label{eqn:rewritten_attention}
\end{align}
where $\tilde{p}_{ij} \sim \mathcal{N}(0,1)$ and $\sigma_a$ increases with $\tau_s$ decreases. This implies that with a lower temperature $\tau_s$, the self-attention (SA) mechanism preserves more high-frequency components in the token signals, thereby preventing collapse in long texts.

\begin{figure}[t]  
    \centering    
    \includegraphics[width=0.9\linewidth]{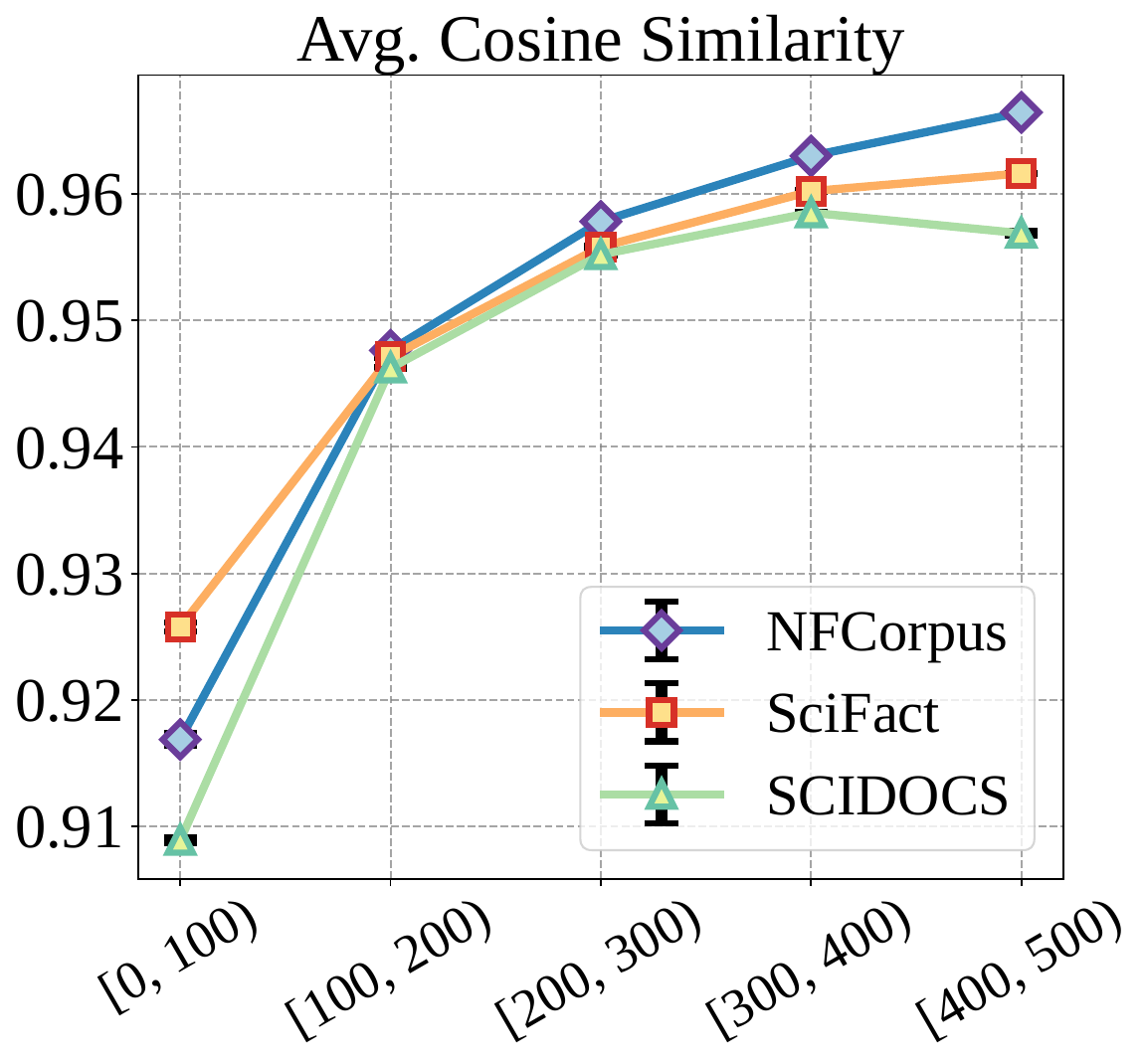}
    \caption{Mean cosine similarity of text embeddings across length intervals, with embeddings averaged from the model's word embedding matrix.}
    \label{fig:theorem4}
    \vspace{-10pt}
\end{figure}

\begin{corollary} \label{thm:length_collapse} (Length Collapse in Text Embeddings)
As the text length $n$ increases, the cosine similarity of their embeddings tends to increase.
\end{corollary}

The proof of Corollary~\ref{thm:length_collapse} in Appendix~\ref{prf:length_collapse} is based on the assumption that the mean of word embeddings in natural language texts remains consistent, with the formal definition and proof detailed therein. As shown in Figure~\ref{fig:theorem4}, we compute text embeddings by averaging word embeddings from the BGE model's embedding matrix and then assess similarity across length intervals. The results show that as text length increases, embedding similarity rises, supporting the assumption. However, we must emphasize that the increased low-pass filtering with length is the primary cause of Length Collapse. In Appendix~\ref{sec:details_theorem4_assumption}, we demonstrate that repeating two different tokens and computing the similarity between the sequences shows that the cosine similarity of their embeddings increases with length, even without any overlapping tokens.


\begin{figure*}[h]
    \vspace{-20pt}
    \centering
      \begin{subfigure}[b]{0.3\textwidth}
         \centering
         \includegraphics[width=\textwidth]{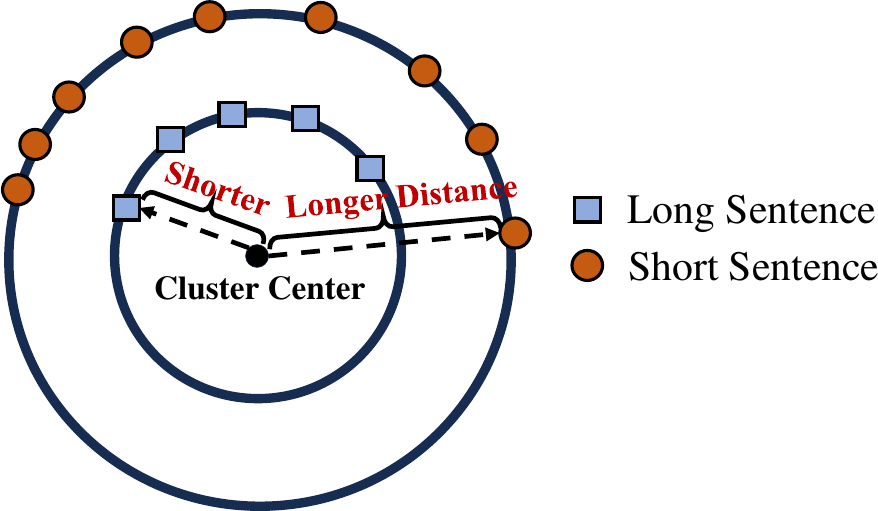}
         \label{fig:class_cause}
         \vspace{-12pt}
         \caption{{Classification and Clustering}}
     \end{subfigure}
     \hfill
      \begin{subfigure}[b]{0.3\textwidth}
         \centering
         \includegraphics[width=\textwidth]{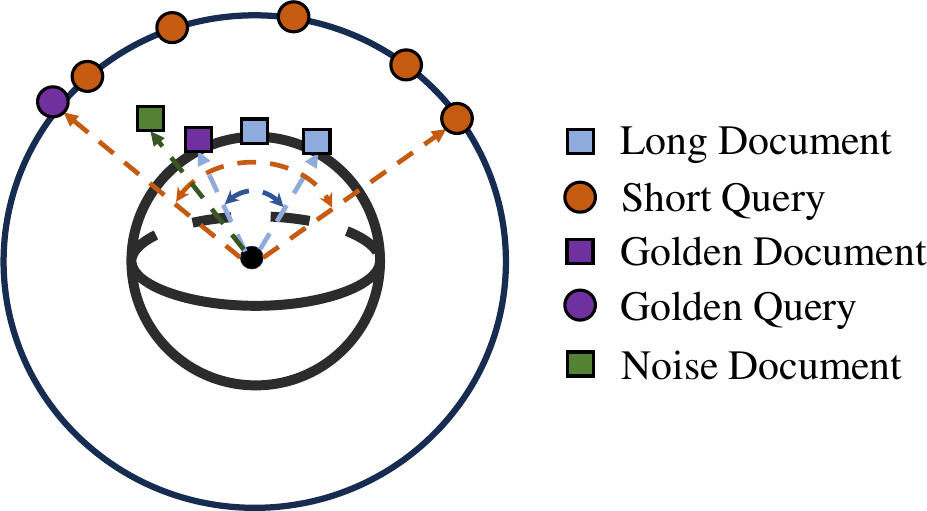}
         \label{fig:retrieval_cause}
         \vspace{-12pt}
         \caption{{Retrieval}}
     \end{subfigure}
     \hfill
      \begin{subfigure}[b]{0.32\textwidth}
         \includegraphics[width=\textwidth]{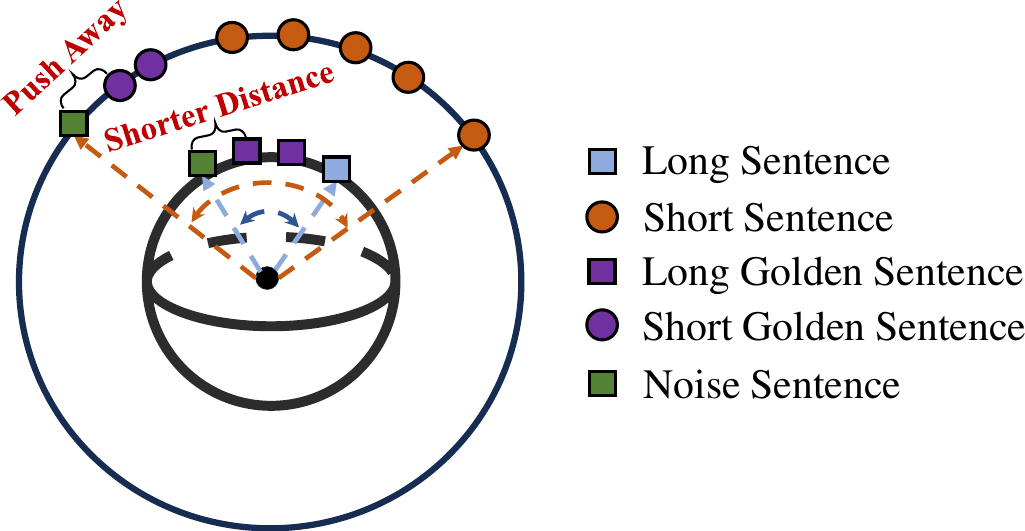}
        \label{fig:sts_cause}
         \centering
         \vspace{-12pt}
         \caption{{STS}}
     \end{subfigure}
     \hfill
    \vspace{-4pt}
    \caption{{A 3D toy example explains why Length Collapse leads to performance degradation.}}
    \label{fig:cause_of_degreation}
\end{figure*}

\subsection{Discussion}
\textbf{Other Components in Transformer.} After discussing how self-attention contributes to Length Collapse, we further examine the impact of other Transformer components, including multi-head attention, residual connections, and feed-forward networks (FFN). Fortunately, prior work~\citep{wanganti} has analyzed whether these modules can alleviate the low-pass filtering effect. The findings show that while they help preserve high-frequency signals, they do not change the fact that the MSA block, as a whole, behaves like a low-pass filter. Moreover, their ability to retain high-frequency information depends entirely on internal parameters and architecture, rather than input length. Therefore, these components do not affect our analysis of Length Collapse in practical Transformer models.

\textbf{Difference from Over-Smoothing in Deeper Layers.} In previous research~\citep{wanganti}, it has been noted that a self-attention module acts as a low-pass filter, causing input feature maps to gradually lose high-frequency signals as the model layers go deeper. Furthermore, other studies~\citep{oonograph,cai2020note,peng2024beyond} indicate that the node features of Graph Convolutional Networks (GCNs) can become exponentially trapped in the null space of the graph Laplacian matrix. The root cause of this phenomenon is that both graph Laplacian matrices and self-attention matrices consistently exhibit a dominant eigenvector, commonly referred to as the DC component. While these studies address over-smoothing in deeper layers, we focus on how the low-pass filtering process changes as the input sequence lengthens, specifically examining over-smoothing in longer sequences.

\section{How does Length Collapse Lead to Performance Degradation}\label{sec:cause_of_degradation}
Building on our analysis of how the self-attention mechanism contributes to Length Collapse, this section investigates its impact on model performance across the MTEB benchmark. The tasks in MTEB can be broadly categorized into classification/clustering and matching tasks. For classification and clustering, a classifier is trained on text embeddings. In matching tasks, performance is evaluated by computing the similarity between embeddings using cosine similarity or the dot product. These matching tasks can be further divided into retrieval tasks, which typically involve texts of significantly different lengths, and STS/summarization tasks, which involve texts of similar lengths. We examine how Length Collapse affects model performance across these different task types.

\textbf{Impact on Classification and Clustering Tasks.} As shown in Figure~\ref{fig:cause_of_degreation} (Left), Length Collapse causes long-text embeddings to cluster at the center, while short-text embeddings spread around the periphery. In KNN classification, this brings clustering centers closer to long texts, resulting in a biased influence and reducing performance.

\begin{figure*}[h]
    \centering
      \begin{subfigure}[b]{0.3\textwidth}
         \centering
         \includegraphics[width=\textwidth]{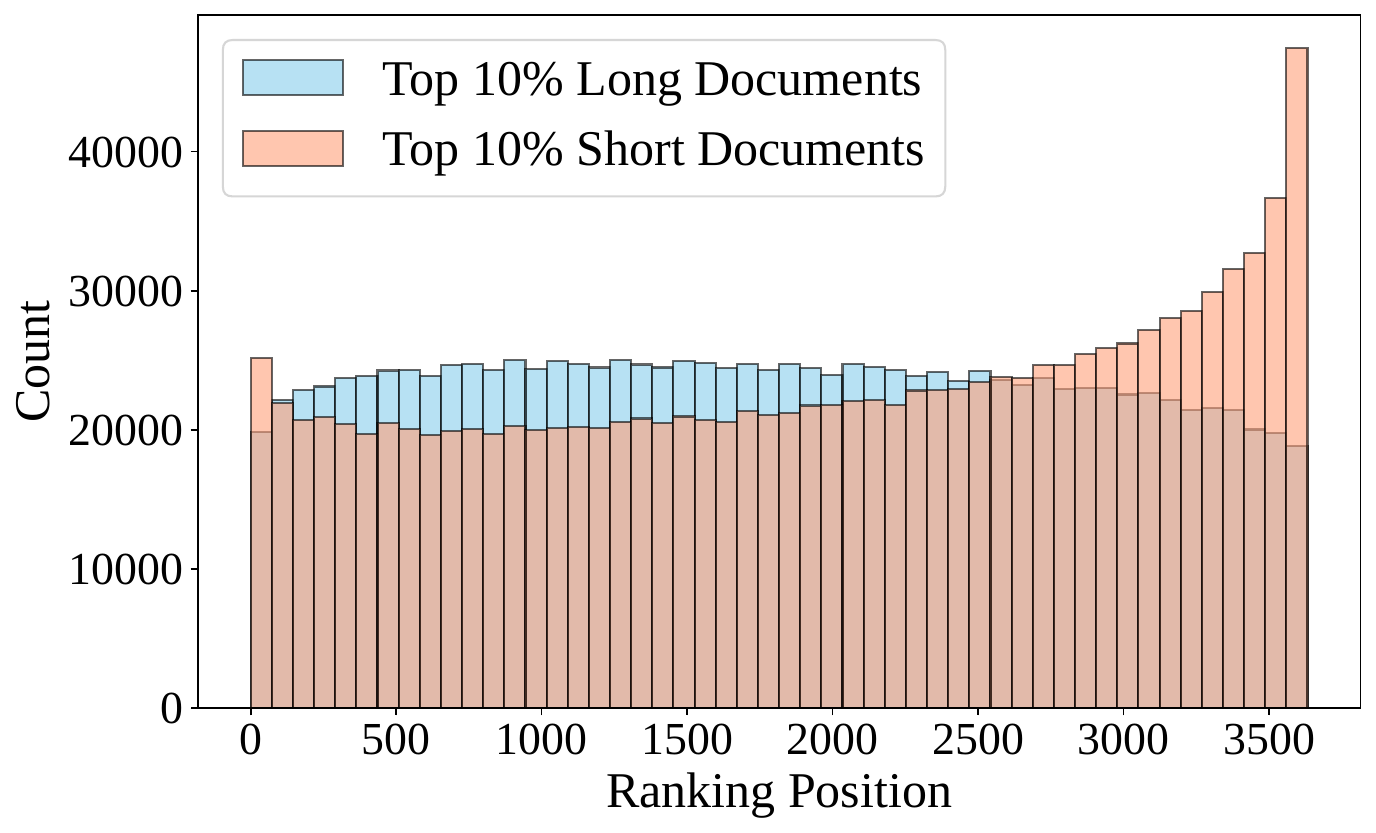}
         \caption{ANCE}
     \end{subfigure}
     \hfill
      \begin{subfigure}[b]{0.3\textwidth}
         \centering
         \includegraphics[width=\textwidth]{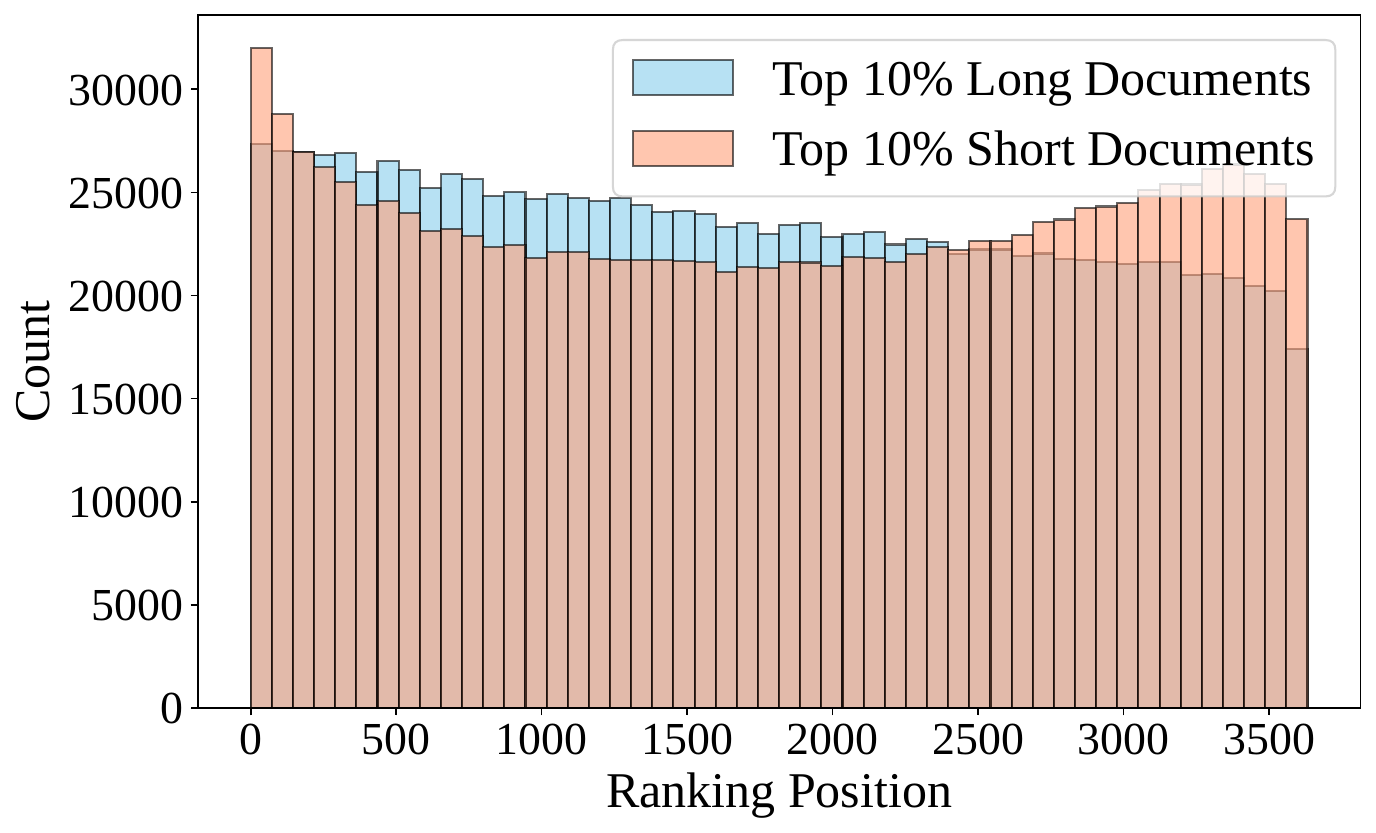}         \caption{GIST}
     \end{subfigure}
     \hfill
      \begin{subfigure}[b]{0.3\textwidth}
         \includegraphics[width=\textwidth]{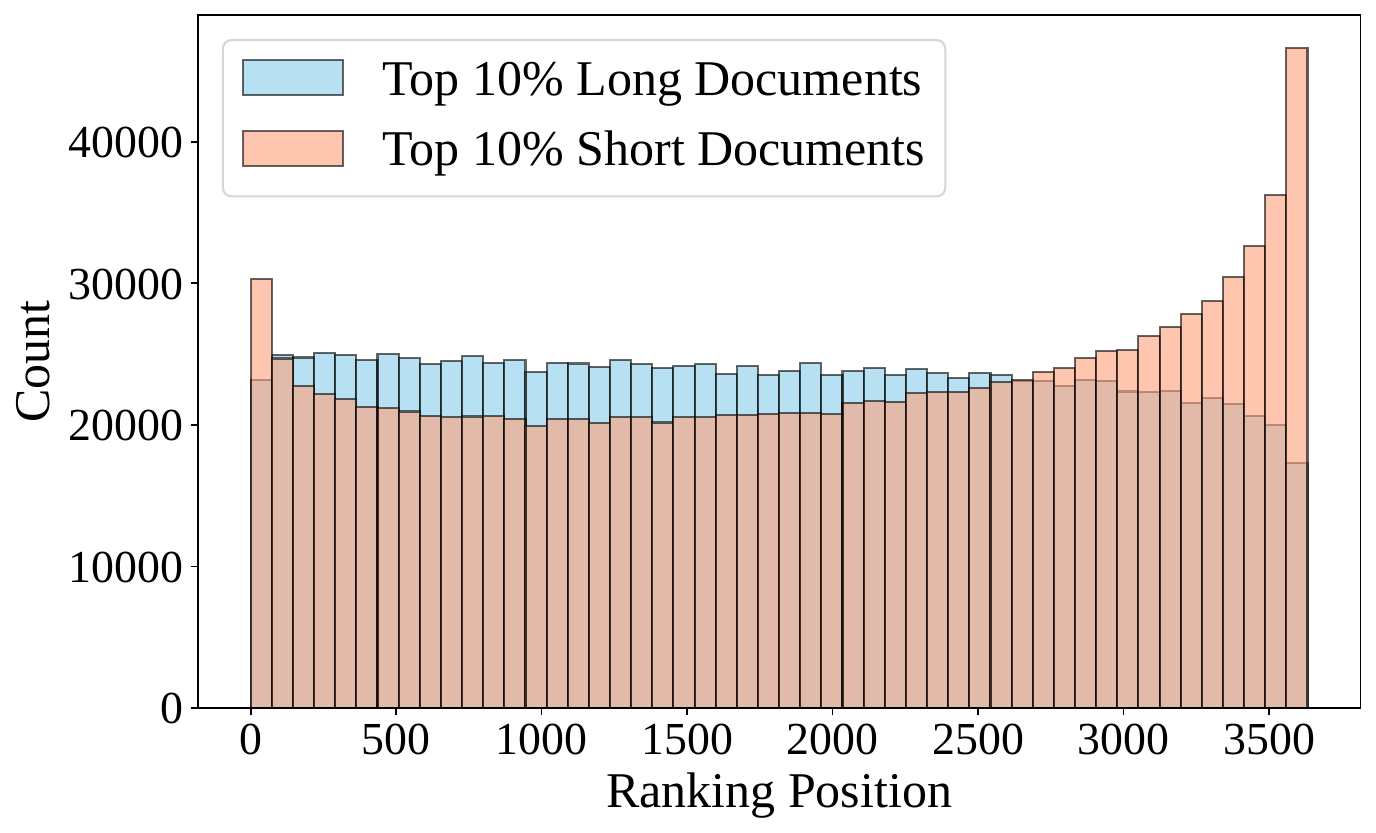}
         \centering
         \caption{E5-4K}
     \end{subfigure}
     \hfill
    \vspace{-8pt}
    \caption{{The count of relevant documents at different ranking positions, where a smaller ranking position indicates a document is more relevant to the query.}}
    \label{fig:ranking_position}
\end{figure*}

\textbf{Impact on Retrieval Tasks.} As shown in Figure~\ref{fig:cause_of_degreation} (Middle), long documents are clustered in a smaller central space, while shorter queries are more dispersed, with richer contextual representations~\cite{ethayarajh2019contextual}. While centrally-positioned long documents have higher similarity with all embeddings, their representational space is more limited. This can lead to shorter noise documents (Green Square) appearing more relevant to the query (Purple Circle) due to their broader representational space. Analysis of the NFCorpus dataset, comparing rankings of the top 10\% longest and shortest relevant documents (Figure~\ref{fig:ranking_position}), shows that short documents follow an inverted U-shaped distribution, whereas long documents exhibit a more uniform distribution. This is because shorter documents have a larger representational space, making it easier for them to appear at the beginning or end of the ranking list.

\textbf{Impact on STS Tasks.} As shown in Figure~\ref{fig:cause_of_degreation} (Right), Length Collapse causes long-text embeddings to cluster in a narrower space, where unrelated embeddings may show higher similarity than relevant pairs. Noise sentences (Green Square) in the long-text space may appear more similar than related sentences (Purple Square). In contrast, short-text embeddings are more spread out, resulting in better performance due to greater separation between noise and relevant sentences.

\section{Mitigating Length Collapse via Temperature Scaling}\label{sec:method}
As discussed in \S~\ref{sec:theorem_analysis}, the self-attention matrix applies stronger low-pass filtering to longer texts, resulting in distributional differences between long and short text embeddings, which subsequently leads to a decline in long-text performance across various tasks in the MTEB benchmark. To address this issue and validate our analysis, a straightforward solution is to reduce the filtering rate difference between long and short texts, thereby mitigating the disparity in their embeddings. To achieve this, we propose a scaling technique called Temperature Scaling (\textbf{TempScale}), which directly adjusts the attention map by multiplying $\tau_s$ by a constant temperature $\tau$ less than 1, thereby reducing the difference in filtering rates $\sigma_a$ between long and short texts.

\begin{figure}[t]  
    \centering    
    \includegraphics[width=0.95\linewidth]{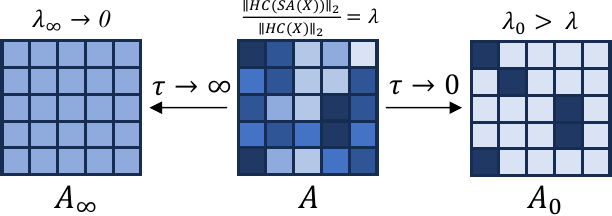}
    \caption{Two extreme cases of TempScale. The darker the color, the higher the attention score.}
    \label{fig:tempscale_case}
    \vspace{-5pt}
\end{figure}

Specifically, a larger $\sigma_s$ will result in a smaller factor $2\sqrt{1 + \frac{1}{e^{2\sigma^{2}_{s}}}}$ in Eqn.~\ref{ieqn:rate_with_len}, which means that as the text length $n$ increases, $\sigma_a$ will decrease more slowly, thereby reducing the difference in filtering rates between long and short texts. Therefore, by increasing $\sigma_s$, or equivalently decreasing $\tau_s$ based on Eqn.~\ref{eqn:tau_sigma}, we can mitigate the distributional difference between long and short text embeddings. 

Formally, let $\Mat{A}=\softmax\left( \frac{\Mat{X}\Mat{W}_{Q}(\Mat{X}\Mat{W}_{K})^\top}{\sqrt{d}} \right)$ denote a self-attention matrix. To decrease the $\tau_s$ of $\Mat{A}$ based on Eqn.~\ref{eqn:rewritten_attention}, we apply a temperature coefficient $\tau$ to the logits before performing the softmax operation. Specifically, for each row $\Mat{p}_i$ in the attention score matrix $\frac{\Mat{X}\Mat{W}_{Q}(\Mat{X}\Mat{W}_{K})^\top}{\sqrt{d}}$, we compute the scaled logits by dividing by a temperature $\tau \in (0,1]$, and then apply the softmax function to obtain the attention weights:
\begin{align}
    \Mat{A}=\softmax\left( \frac{\Mat{X}\Mat{W}_{Q}(\Mat{X}\Mat{W}_{K})^\top}{{\color{blue}\tau} \sqrt{d}} \right), \label{eqn:temp_scaling}
\end{align}
where a lower $\tau$ results a smaller $\tau_s$.

\textbf{Another Intuitive Explanation.} We illustrate the effects of temperature scaling using two extreme cases to highlight how TempScale works from a different perspective. As shown in Figure~\ref{fig:tempscale_case}, when scaling the matrix $\Mat{A}$ with a relatively large $\tau$, the elements of $\Mat{A}$ become nearly equal, causing the matrix to filter out all high-frequency information and resulting in identical token embeddings. In contrast, when scaling with a smaller temperature, $\Mat{A}$ no longer acts as a weighted sum of token representations, but selects a specific token's representation, preserving more high-frequency information. Viewing the attention matrix as an adjacency matrix, a higher temperature leads to a denser graph, enhancing information exchange between nodes but losing high-frequency details~\cite{oonograph,cai2020note}. A lower temperature, on the other hand, creates a sparser graph, preserving more high-frequency information and preventing over-smoothing between nodes.


\section{Experiments}\label{sec:experiments}
In this section, we first conduct experiments to validate the effectiveness of our TempScale on MTEB and LongEmbed. Then we analyze how different tasks can benefit from TempScale to validate our theoretical analysis.

\begin{table*}[t]
\centering
\resizebox{0.9\textwidth}{!}{%
\begin{tabular}{@{}lcccccccc@{}}

\toprule[1.5pt]
                 & \textbf{Class.}                      & \textbf{Clust.}                      & \textbf{Summ.}                       & \textbf{STS}                         & \textbf{BeirRetr.}                   & \textbf{Rerank.}                     & \textbf{LongEmbdRetr.}               & \textbf{Avg.}                        \\ 
\textbf{Num. Datasets} $(\rightarrow)$    & \textbf{8}                           & \textbf{11}                          & \textbf{1}                           & \textbf{10}                          & \textbf{5}                           & \textbf{4}                           & \textbf{4}                           & \textbf{43}                          \\ \midrule \midrule
\multicolumn{9}{c}{\textit{window=512}}                                                                                                                                                                                                                          \\ \midrule
ANCE             & 55.27 &	33.04&	29.58&	66.32&	26.95&	49.09&	34.02&	42.04 \\
+TempScale  & 55.44 &	33.28&	29.59&	66.47&	27.39&	49.25&	34.07&	42.21 \\
Relative Improv. (\%) & \phantom{\down} 0.29 \textcolor{green}{\up} &	\phantom{\down} 0.73 \textcolor{green}{\up}	& \phantom{\down} 0.06 \textcolor{green}{\up} &	\phantom{\down} 0.22 \textcolor{green}{\up} &	\phantom{\down} 1.63 \textcolor{green}{\up} &	\phantom{\down} 0.32 \textcolor{green}{\up} &	\phantom{\down} 0.17 \textcolor{green}{\up} &	\phantom{\down} 0.49 \textcolor{green}{\up} \\ \midrule
GTR              & 55.10&	38.65&	29.67&	70.11&	33.00&	54.23&	37.33&	45.44 \\
+TempScale & 55.59&	39.52&	29.83&	70.26&	33.56&	54.22&	37.33&	45.76 \\
Relative Improv. (\%) & \phantom{\down} 0.88 \textcolor{green}{\up} &	\phantom{\down} 2.26 \textcolor{green}{\up} &	\phantom{\down} 0.54 \textcolor{green}{\up} &	\phantom{\down} 0.21 \textcolor{green}{\up} &	\phantom{\down} 1.69 \textcolor{green}{\up} & \phantom{\down} -0.01 \textcolor{red}{\down} &	\phantom{\down} 0.01 \textcolor{green}{\up} &	\phantom{\down} 0.80 \textcolor{green}{\up} \\ \midrule
GIST             & 64.75&	44.77&	31.14&	75.61&	37.55&	58.55&	38.21&	50.08 \\
+TempScale & 65.12&	44.66&	32.17&	75.61&	37.82&	58.60&	38.35&	50.33 \\
Relative Improv. (\%) & \phantom{\down} 0.56 \textcolor{green}{\up} &	\phantom{\down} -0.25 \textcolor{red}{\down} &	\phantom{\down} 3.31 \textcolor{green}{\up} &	\phantom{\down} -0.01 \textcolor{red}{\down} &	\phantom{\down} 0.71 \textcolor{green}{\up}	& \phantom{\down} 0.08 \textcolor{green}{\up} &	\phantom{\down} 0.36 \textcolor{green}{\up} &	\phantom{\down} 0.68 \textcolor{green}{\up} \\ \midrule
BGE              & 64.79&	45.80&	31.03&	75.88&	38.14&	58.87&	37.46&	50.28 \\
+TempScale & 64.91&	45.79&	31.87&	75.68&	38.46&	58.97&	38.73&	50.63 \\
Relative Improv. (\%) & \phantom{\down} 0.19 \textcolor{green}{\up} &	\phantom{\down} -0.01 \textcolor{red}{\down} &	\phantom{\down} 2.71 \textcolor{green}{\up} &	\phantom{\down} -0.26 \textcolor{red}{\down} &	\phantom{\down} 0.84 \textcolor{green}{\up}	 & \phantom{\down} 0.17 \textcolor{green}{\up} &	\phantom{\down} 3.41 \textcolor{green}{\up} &	\phantom{\down} 1.01 \textcolor{green}{\up} \\ \midrule
\multicolumn{9}{c}{\textit{window=4k}}                                                                                                                                                                                                                           \\ \midrule
E5               & 61.72&	38.82&	30.58&	71.77&	29.77&	53.12&	56.01&	48.83 \\
+TempScale & 62.15&	40.62&	31.26&	72.17&	30.28&	53.47&	56.88&	49.55 \\
Relative Improv. (\%) & \phantom{\down} 0.7 \textcolor{green}{\up} &	\phantom{\down} 4.62 \textcolor{green}{\up} &	\phantom{\down} 2.25 \textcolor{green}{\up} &	\phantom{\down} 0.55 \textcolor{green}{\up} &	\phantom{\down} 1.69 \textcolor{green}{\up}	& \phantom{\down} 0.65 \textcolor{green}{\up} &	\phantom{\down} 1.56 \textcolor{green}{\up}	& \phantom{\down} 1.72 \textcolor{green}{\up} \\ \midrule

Avg Improv. (\%) & \phantom{\down} 0.53 \textcolor{green}{\up} &	\phantom{\down} 1.47 \textcolor{green}{\up} &	\phantom{\down} 1.77 \textcolor{green}{\up} &	\phantom{\down} 0.14 \textcolor{green}{\up} &	\phantom{\down} 1.31 \textcolor{green}{\up} & 	\phantom{\down} 0.24 \textcolor{green}{\up} &	\phantom{\down} 1.10 \textcolor{green}{\up} &	 \phantom{\down} 0.94 \textcolor{green}{\up} \\

\bottomrule[1.5pt]
\end{tabular}%
}
\caption{Average of the main metric (see Appendix~\ref{sec:datasets}) per task on MTEB English subsets and LongEmbd. Relative Improv. means percentage increase over the performance without TempScale and improvements are highlighted with \textcolor{green}{\up} while decreasing values are denoted by \textcolor{red}{\down}.}
\label{tab:main_table}
\end{table*}

\subsection{TempScale Benefits Embedding Models}
\textbf{Experiment Settings.} We evaluate embedding models on long-context retrieval using 4 real-world tasks from LongEmbed~\cite{zhu2024longembed}, focusing on factuality in both short-query and long-document settings. For other tasks, we select 39 datasets from MTEB~\cite{muennighoff2023mteb}, covering classification, clustering, summarization, STS, retrieval, and reranking. Specifically, for classification, we only downloaded 10 datasets due to open-source restrictions. For retrieval, we select the $5$ datasets with the longest documents to better assess Length Collapse. To comprehensively evaluate TempScale, we select several representative PLM-based embedding models, including (1) ANCE~\cite{xiong2021approximate}; (2) GTR~\cite{ni2022large}; (3) GIST~\cite{solatorio2024gistembed}; (4) BGE~\cite{xiao2023c}; (5) E5~\cite{zhu2024longembed}. These models are fine-tuned from various pretrain language models, including BERT~\cite{kenton2019bert}, RoBERTa~\cite{liu2019roberta}, and T5~\cite{chung2024scaling}. More descriptions of the datasets and models can be found in Appendix~\ref{sec:models}. When evaluating the embedding models, we set the same $\tau$ on the softmax function for the attention modules across all layers within the range of $\{0.1, 0.5, 0.6, 0.7, 0.8, 0.9, 1.0\}$. The metrics used for different tasks are consistent with MTEB and can be found in Appendix~\ref{sec:datasets}.

\textbf{Results.} We select the optimal temperature $\tau$ for each model based on their performance across a single task and present the results in Table~\ref{tab:main_table}. The findings demonstrate that our proposed method, TempScale, improves performance across a range of general tasks, with an average increase of 0.94\%. For long-context retrieval tasks, the average improvement is 1.10\%. Text length varies widely across tasks—for example, STS uses very short texts, whereas LongEmbed includes much longer ones (1000+ tokens). TempScale proves effective for both preventing long-text collapse and enhancing short-text embeddings, which improves downstream task performance. Additionally, larger context windows yield greater performance gains, with E5 showing the highest improvement of 1.72\%. This may be due to larger windows offering more data for adjustment. Beyond the excellent performance of TempScale, more importantly, it validates the correctness of our theoretical analysis.

\subsection{Further Analysis}
\textbf{How do the Classification and Clustering Tasks Benefit from TempScale?} In \S~\ref{sec:cause_of_degradation}, we attribute the performance decline in classification and clustering tasks to the distributional differences between long and short texts, which lead the model to assign greater weight to long text embeddings during classifier training. TempScale addresses this by adjusting the longer texts to align with the same space as short texts, as described in Figure~\ref{fig:intro_cosine}, ensuring that both contribute equally during training. More experiments can be found in Appendix~\ref{sec:discussions_on_class}.

\textbf{How do Retrieval Tasks Benefit from TempScale?} In \S~\ref{sec:cause_of_degradation}, we attribute the performance decline in retrieval tasks to the fact that short texts have more contextual embeddings. Therefore, TempScale improves the performance of long texts by adjusting the temperature to increase high-frequency information, enabling a more contextual distribution. To validate this, we select NFCorpus and SciFact to investigate the impact. As shown in Table~\ref{tab:longdoc}, we record the average ranking position of the longest positive documents after applying TempScale at different temperatures, where a lower value indicates the documents are ranked higher. The experimental results show that a lower temperature causes the model to rank relevant long documents higher. Our method enables long texts to become more contextual, thereby reducing the bias introduced by length. In summary, TempScale enhances the performance of long documents by introducing TempScale. Similar phenomena on more models and datasets are provided in Appendix~\ref{sec:more_long_queries}.

\begin{table*}[h!]
    \vspace{-3pt}
    \centering
    \begin{subtable}[t]{0.4\linewidth}
        \centering
        \label{tab:longdoc_nfcorpus}
        \resizebox{\linewidth}{!}{
            \begin{tabular}{lrrr}
                \toprule
                \textbf{Temperature $\tau$}       & 1.0  & 0.9   & 0.8  \\
                \midrule
                ANCE & 1,306.2 & 1,291.6 & 1,278.1 \\
                GTR & 1,132.8 & 1,120.5 & 1,113.1 \\
                GIST & 1,111.0 & 1,112.8 & 1,113.8 \\
                BGE & 998.3 & 978.9 & 965.1 \\
                E5 & 1,193.3 & 1,172.3 & 1,162.8 \\
                \bottomrule
            \end{tabular}
        }
        \caption{NFCorpus}
        \vspace{-5pt}
    \end{subtable}
    \hspace{0.05\linewidth}
    \begin{subtable}[t]{0.33\linewidth}
        \centering
        \label{tab:longdoc_scifact}
        \resizebox{\linewidth}{!}{
            \begin{tabular}{lccc}
                \toprule
                \textbf{Temperature $\tau$}       & 1.0  & 0.9   & 0.8  \\
                \midrule
                ANCE & 80.7 & 78.7 & 81.5 \\
                GTR  & 81.7 & 76.6 & 72.4 \\
                GIST & 14.4 & 12.4 & 11.3 \\
                BGE  & 13.3 & 12.3 & 12.4 \\
                E5   & 66.8 & 47.5 & 38.9 \\
                \bottomrule
            \end{tabular}
        }
        \caption{SciFact}
        \vspace{-5pt}
    \end{subtable}
    \caption{Average ranking position with 20\% longest document across different temperature $\tau$.}
    \label{tab:longdoc}
\end{table*}

\textbf{How do the STS Tasks Benefit from TempScale?} TempScale enhances performance by giving long sentence embeddings the same spatial representation as short texts. To verify this, we record the cosine similarity between random sentences, related sentences, and unrelated sentences at different temperatures, as shown in Table~\ref{tab:temperature_cos_sts}. The results show that as the $\tau$ in TempScale decreases, the similarity between related sentences remains relatively high, while the similarity between unrelated sentences is further reduced. Specifically, although the similarity between random sentences increases with $\tau$ in STS13 and STS14, the increase in the unrelated part is not as large as that of the random sentences. This indicates that the distance between unrelated sentences is increasing.   

\textbf{Can the Temperature be Set Based on the Length of the Texts?} As shown in Table~\ref{tab:longdoc}, the ranking of relevant long documents in retrieval tasks improves as the temperature decreases. This indicates that using a smaller temperature $\tau$ for long texts can better align the embeddings of long and short texts within the same distribution. For longer texts, better performance can be achieved by setting a smaller $\tau$, and the discussions can be found in Appendix~\ref{sec:more_long_queries}.

\begin{table*}[h]
\centering
\renewcommand{\arraystretch}{1.2}
\resizebox{\textwidth}{!}{
\begin{tabular}{c|c|c|c|c|c|c|c|c|c}
\hline
\multirow{2}{*}{\textbf{Temperature}} & \multicolumn{3}{c|}{\textbf{STS12}} & \multicolumn{3}{c|}{\textbf{STS13}} & \multicolumn{3}{c}{\textbf{STS14}} \\ \cline{2-10} 
                                      & \textbf{Random} & \textbf{Related} & \textbf{Unrelated} & \textbf{Random} & \textbf{Related} & \textbf{Unrelated} & \textbf{Random} & \textbf{Related} & \textbf{Unrelated} \\ \hline
1.0                                     & 0.9097          & 0.9611           & 0.7973             & 0.8707          & 0.9470           & 0.7931             & 0.8877          & 0.9494           & 0.7889             \\ 
0.9                                   & 0.9096          & 0.9612           & 0.7937             & 0.8721          & 0.9481           & 0.7942             & 0.8886          & 0.9499           & 0.7893             \\
0.8                                   & 0.9097          & 0.9612           & 0.7926             & 0.8741          & 0.9490           & 0.7969             & 0.8897          & 0.9501           & 0.7912             \\ \hline
\end{tabular}
}
\caption{Average cosine similarities across different temperature settings.}~\label{tab:temperature_cos_sts}
\vspace{-10pt}
\end{table*}

\section{Related Work}\label{sec:related_work}
\textbf{Text Embedding Models.}
Embeddings are generated by pre-trained language models (PLMs), which produce fixed-dimensional embeddings regardless of the input text length, laying the foundation of numerous NLP applications. Early works on text embeddings lack context awareness and are thus commonly labeled as word embedding models~\citep{pennington2014glove}. Modern embedding models~\citep{wang2022text,xiao2023c} incorporate context awareness into language models through self-attention mechanisms, serving as the foundation for the latest embedding models. These models, based on Transformer architectures like BERT~\citep{kenton2019bert} and RoBERTa~\citep{liu2019roberta}, are first pre-trained on large-scale weakly supervised text pairs using contrastive loss~\citep{gao2021simcse}, then fine-tuned on small scale but high-quality datasets. More recently, \citep{muennighoff2024generative} investigates the integration of generative and embedding tasks in large language models, introducing transformer-based GritLM, which enhances performance in both areas. Among the diverse range of pre-trained transformers~\citep{wolf2020transformers}, self-attention plays a crucial role, and this paper focuses on how this module contributes to length collapse.

\textbf{Context Window Extension for Embedding Models.}
Despite transformer-based embedding models excelling in generating vector representations, they are typically constrained by a narrow context window of around 512 input tokens~\citep{wang2022text,xiao2023c,ni2022large}. This limitation significantly restricts their use in scenarios that require processing long inputs, such as extensive Wikipedia entries or meeting transcripts~\citep{saad2024benchmarking,zhu2024longembed}. Previous work attributes this poor performance to limited context windows and attempts to extend the window size. Current initiatives to develop long-context embedding models generally begin with acquiring a long-context backbone model, either by pre-training from scratch with long inputs~\citep{gunther2023jina,nussbaum2024nomic,chen2024bge} or utilizing existing models~\citep{wang2024improving,zhu2024longembed}. This is followed by training the backbone model to generate embeddings. However, in this paper, we discover that regardless of the model's context window size, the model consistently performs worse on longer texts than on shorter texts due to length collapse. Therefore, we aim to improve the performance of long texts across all context window size models by analyzing and addressing length collapse, rather than simply expanding context window size.

\section{Discussion}\label{sec:discussion}
\textbf{LLM-based Embedding Models:} While our main focus is on PLM-based models, we also observe similar signs of Length Collapse in LLM-based models, which represent another major class of current mainstream models. We provide a detailed discussion of these observations in Appendix~\ref{sec:llm_length_collapse}. \textbf{Contrastive Learning:} Length Collapse increases similarity between long and short text embeddings. Although contrastive learning mitigates high similarity, performance degradation in long texts is also due to distribution differences between long and short texts, not just similarity. Further analysis of why contrastive learning fails to fully mitigate Length Collapse is provided in Appendix~\ref{sec:contrastive_learning}. \textbf{Comparison with Other Methods:} We compare our method with post-processing approaches (Flow Function~\cite{li2020sentence}, Whitening~\cite{su2021whitening}) and long-text methods (\citet{chiang2022overcoming}, YaRN~\cite{pengyarn}) in Appendix~\ref{sec:post_processing} and~\ref{sec:long_method}. Despite similar forms, these methods fail to address Length Collapse effectively.

\section{Conclusion}\label{sec:conclusion}
In this paper, we identify the phenomenon of Length Collapse, where the embeddings of longer texts tend to cluster together, and provide a Fourier domain analysis to explain this behavior. Our theoretical findings suggest that Self-Attention inherently performs stronger low-pass filtering as the text length increases, leading to patch uniformity issues in longer sentences. Furthermore, the distributional differences between short and long text embeddings contribute to the performance decline of longer texts. To address this, we propose TempScale, which effectively balances the filtering rates for short and long texts. Our extensive experiments validate the accuracy of our analysis and the effectiveness of our method, significantly improving the performance of general embedding models on the MTEB and LongEmbed benchmarks.

\section*{Limitations}
Limitations include: \textbf{1) LLM-based embedding model:} Although we have observed a Length Collapse in LLM-based embedding models, further analysis is needed to investigate how unidirectional attention mechanisms specifically contribute to this phenomenon. Additionally, LLMs also exhibit a performance drop on long texts during text generation, which can be seen as Length Collapse. However, in this work, we only focus on PLM-based models across various tasks in MTEB. Future work could attempt to explain the reasons behind the performance decline of LLMs on long text generation from the perspective of low-pass filtering; \textbf{2) Tuning method:} The work in this paper relies on existing models and pre-trained parameters without using a training dataset. In future work, we will focus on tuning the temperature for additional improvements; \textbf{3) Analysis on more modules:} We primarily investigate the impact of the self-attention module on Length Collapse in this paper. Moving forward, we plan to explore the effects of additional modules in transformers such as LayerNorm and FFN.

\section*{Acknowledgements}
This work was funded by the National Key R\&D Program of China (2023YFA1008704), the National Natural Science Foundation of China (62472426,62376275). Supported by fund for building world-class universities (disciplines) of Renmin University of China. Work partially done at Beijing Key Laboratory of Research on Large Models and Intelligent Governance, and Engineering Research Center of Next-Generation Intelligent Search and Recommendation, MOE. 

\bibliography{custom}
\appendix

\clearpage
\newpage
\section{Background Information about Fourier Analysis} \label{sec:background_fourier}
In this appendix, we provide additional background information on Fourier analysis. Specifically, consider the discrete Fourier transform (DFT) in the real-valued domain, denoted as $\mathcal{F}: \mathbb{R}^n \rightarrow \mathbb{C}^n$. The DFT can be expressed in matrix form as shown below:
\begin{align*} \label{eqn:dft_matrix}
\scalebox{0.85}{%
  $\Mat{DFT} = \frac{1}{\sqrt{n}} \begin{bmatrix}
  1 & 1 & \cdots & 1 \\
  1 & e^{2 \pi \mathrm{j}} & \cdots & e^{2 \pi \mathrm{j}(n-1)} \\
  \vdots & \vdots & \ddots & \vdots \\
  1 & e^{2 \pi \mathrm{j}(k-1)} & \cdots & e^{2\pi \mathrm{j} (k-1) \cdot (n-1)} \\
  \vdots & \vdots & \ddots & \vdots \\
  1 & e^{2 \pi \mathrm{j}(n-1)} & \cdots & e^{2\pi \mathrm{j}(n-1)^2}
  \end{bmatrix}$%
}
\end{align*}
and inverse discrete Fourier transform is $\Mat{DFT}^{-1} = \Mat{DFT}^{\top} = \Mat{DFT}$.
In signal processing, we can regard matrices as multi-channel signals. For example, $\Mat{X} \in \real^{n \times d}$ means $d$-channel $n$-length signals. When the DFT and inverse DFT are applied to multi-channel signals, each channel is transformed independently. That is, $\FT(\Mat{X}) = \begin{bmatrix}\FT(\Mat{x}_1) & \cdots & \FT(\Mat{x}_d)\end{bmatrix} = \Mat{DFT}\cdot\Mat{X}$.

Hereby, we can independently operate $\DC{\cdot}$ and $\HC{\cdot}$ on the echo channel using the matrices in Eqn. \ref{eqn:dft_matrix}. Then we can write $\DC{\cdot}$ as below:
\begin{align*}
\DC{\Mat{x}} &= \Mat{DFT}^{-1} \diag(1, 0, \cdots, 0) \Mat{DFT} \Mat{x} \\
&= \frac{1}{n} \Mat{1} \Mat{1}^T \Mat{x}.
\end{align*}

Conversely, we can denote $\HC{\cdot}$ as:
{\small{
\begin{align*}
\HC{\Mat{x}} &= \Mat{DFT}^{-1} \diag(0, 1, \cdots, 1) \Mat{DFT} \Mat{x} \\
&= \Mat{DFT}^{-1} (\Mat{I} - \diag(1, 0, \cdots, 0)) \Mat{DFT} \Mat{x} \\
&= (\Mat{I} - \frac{1}{n} \Mat{1} \Mat{1}^T) \Mat{x}.
\end{align*}
}}

\section{Detailed Proofs}

\subsection{Proof of Theorem \ref{thm:lambda2_sa_rate}} \label{prf:lambda2_sa_rate}
We start our analysis by providing a lemma.

\begin{lemma} ~\label{thm:f_2norm}
    The following holds $\lVert \Mat{AB} \rVert_F \leq \lVert \Mat{A} \rVert_{2} \lVert \Mat{B} \rVert_{F}$ and $\lVert \Mat{AB} \rVert_F \leq \lVert \Mat{A} \rVert_{F} \lVert \Mat{B} \rVert_{2}$.
\end{lemma}

\begin{proof}
        Denote $\Mat{B} = \begin{pmatrix} \Mat{b_1} & \cdots & \Mat{b_n} \end{pmatrix}$ and we have $\Mat{AB}=\begin{pmatrix} \Mat{Ab_1} & \cdots & \Mat{Ab_n} \end{pmatrix}$. From the definition of the spectral norm, we have:$\lVert \Mat{A} \rVert_2 \geq \frac{\lVert\Mat{Ab_i}\rVert_2}{\lVert\Mat{b_i}\rVert_2}.$ Taking the average of the right-hand side, we obtain: $
\lVert \Mat{A} \rVert_2^2 \geq \sum_{i=1}^n \frac{\lVert \Mat{b_i} \rVert_2^2}{\lVert \Mat{B} \rVert_F^2} \frac{\lVert \Mat{Ab_i} \rVert_2^2}{\lVert \Mat{b_i} \rVert_2^2}.$ This implies: $\lVert \Mat{A} \rVert_2^2 \lVert \Mat{B} \rVert_F^2 \geq \sum_{i=1}^n \lVert \Mat{Ab_i} \rVert_2^2 = \lVert \Mat{AB} \rVert_F^2.$ Finally, the last step utilizes the result $\lVert \Mat{A} \rVert^{2}_{F} = \sum_{i,j}|a_{ij}|^{2}=\sum_{j=1}^{n} \lVert \Mat{a_j} \rVert^{2}_{2}$. Because both the spectral norm and the Frobenius norm of a matrix remain unchanged under transposition, we have $\lVert \Mat{AB} \rVert_{F} = \lVert \Mat{B^{\top}A^{\top}} \rVert_{F} \leq \lVert \Mat{B^{\top}}\rVert_{2} \lVert \Mat{A^{\top}} \rVert_{F} = \lVert \Mat{A} \rVert_{F} \lVert \Mat{B} \rVert_{2}$.

\end{proof}

\begin{theorem}
(Filter Rate of SA) 
Let $\sigma_a$ be the largest singular value of $\HC{\Mat{A}}$. Define $\SA(\Mat{X}) = \Mat{A}\Mat{X}\Mat{W}_V$ as the output of a self-attention module, then
\begin{align}
\lVert \HC{\SA(\Mat{X})} \rVert_F \le \sigma_a \lVert\Mat{W}_V\rVert_2 \lVert \HC{\Mat{X}} \rVert_F.
\end{align}
\end{theorem}

\begin{proof}
First, we write $\Mat{X} = \DC{\Mat{X}} + \HC{\Mat{X}} = \frac{1}{n}\Mat{11^{\top}X} + \Mat{H}$, where $\Mat{H}=\HC{\Mat{X}}$ represents the remaining part of the original signals.

{\scriptsize{
\begin{align}
    \HC{\SA(\Mat{X})} & = \left( \Mat{I} - \frac{1}{n} \Mat{11}^T \right) \Mat{AXW}_{V} \\
    & = \left( \Mat{I} - \frac{1}{n} \Mat{11}^T \right) \Mat{A}(\frac{1}{n}\Mat{11}^{\top}\Mat{X} +\Mat{H})\Mat{W}_{V}  \\
    & = \frac{1}{n}\left( \Mat{I} - \frac{1}{n} \Mat{11}^T \right) \Mat{A11}^{\top}\Mat{X} \Mat{W}_{V} \\ & + \left( \Mat{I} - \frac{1}{n} \Mat{11}^T \right) \Mat{AHW}_{V}  \\
    & = \left( \Mat{I} - \frac{1}{n} \Mat{11}^T \right) \Mat{AHW}_{V}    
\end{align}
}}
\normalsize

Therefore,
{\small{
\begin{align}
    \lVert \HC{\SA(\Mat{X})} \rVert_{F} & = \left\lVert \left( \Mat{I} - \frac{1}{n} \Mat{11}^T \right) \Mat{AHW}_{V} \right\rVert_{F} \\
    \label{eqn:rate_holder} & \leq \left\lVert \left( \Mat{I} - \frac{1}{n} \Mat{11}^T \right)\Mat{A} \right\rVert_{2} \lVert \Mat{W}_{V}\rVert_{2} \lVert \Mat{H} \rVert_{F} \\
    & = \sigma_a \lVert \Mat{W}_{V}\rVert_{2} \lVert \Mat{H} \rVert_{F}
\end{align}}}

The Eqn. \ref{eqn:rate_holder} leverages inequality in Lemma \ref{thm:f_2norm}.

\end{proof}

\subsection{Proof of Theorem \ref{thm:length_rate}} \label{prf:length_rate}

\begin{theorem}
(Filter Rate of Different Input Length $n$)
Let $\Mat{XW}_{Q}$ and $\Mat{XW}_{K}$ be a Gaussian matrix, where elements $q_{ij} \sim \mathcal{N}(0, \sigma_q^2)$ and $k_{ij} \sim \mathcal{N}(0, \sigma_k^2), \forall i, j.$ Let $x_{ij}=\Mat{q}_{i}^{\top} \Mat{k}_{j} / \sqrt{d}$ the attention score of pair $i,j$, whose variance can be expressed as $\sigma_s^2=\sigma_q^2 \sigma_k^2+C_{cross}$, where $C_{cross}$ is the cross-covariance of the squared queries and keys~\cite{goodman1960exact}.  Then we have

\begin{align}
    \sigma_{a} \leq \sqrt{\frac{n}{2\sqrt{1+\frac{1}{e^{2\sigma^{2}_{s}}}}(n-1)^{\frac{3}{2}}+1}},
\end{align}

where $\sigma_{a}$ decreases with $n$ increasing.
\end{theorem}

\begin{proof}
First, we have
\begin{align}
   \sigma_a & = \left\lVert \left(\Mat{I}-\frac{1}{n}\Mat{11}^T \right)\Mat{A} \right\rVert_{2} \\
    & \leq \left\lVert \Mat{I}-\frac{1}{n}\Mat{11}^T \right\rVert_2 \lVert \Mat{A} \rVert_{2} \\
    \label{eqn:sigma_F} & \leq \lVert \Mat{A} \rVert_{F}
\end{align}

The Eqn.~\ref{eqn:sigma_F} leverages $\left\lVert \Mat{I}-\frac{1}{n}\Mat{11}^T \right\rVert_2=1$ and $\lVert \Mat{A} \rVert_{2} \leq \lVert \Mat{A} \rVert_{F}$. Now we need to upper bound $\lVert \Mat{A} \rVert_{F}$. Generally, the product of two independent Gaussian variables has a density in the form of a modified Bessel function of the second kind~\cite{nahshanlinear}. When the vector dimensions are sufficiently large, the Central Limit Theorem implies that the distribution of the dot product between $\Mat{q}_i$ and $\Mat{k}_j$ can be approximated by a Gaussian distribution with zero mean and variance $\sigma^2_{s}$. As mentioned in Theorem \ref{thm:length_rate}, the variance of $\Mat{q}_i^\top \Mat{k}_j$ can be expressed as \(\sigma^2 = \sigma^2_q\sigma^2_k + C_{\rm cross}\), where \(C_{\rm cross} = \text{Cov}(\Mat{q}^2, \Mat{k}^2) - \text{Cov}(\Mat{q}, \Mat{k})^2\) is the cross-covariance of the squared queries and keys \cite{goodman1960exact}. Thus we can suppose that each element $p_j \sim \mathcal{N}(0, \sigma_{s})$ in the matrix $\Mat{XW_{Q}W_{K}^{\top}X^T}$ is independent, where $j \in (1,\cdots,n)$:

{\tiny{
\begin{align}
    \lVert \Mat{A} \rVert_{F} &= \sqrt{n\sum_{j=1}^{n} \left( 
    \frac{e^{x_j}}{\sum_{i=1}^{n}{e^{x_i}}} \right)^{2}}  \\
    &= \sqrt{n\frac{\sum_{j=1}^{n} e^{2x_j}}{2\sum_{i=1}^{n} \sum_{j=1,j \neq i}^{n} e^{x_i+x_j} + \sum_{j=1}^{n} e^{2x_j}}} \\
    \label{eqn:log_normal_add}&= \sqrt{\frac{ne^{\ln{n}+2\sigma_{s}^{2} - \frac{1}{2}\ln{\frac{e^{4\sigma_{s}^{2}}-1}{n}}}}{2e^{\ln{n(n-1)}+\sigma^2_{s} - \frac{1}{2}  \ln{\frac{e^{2\sigma_s^{2}}- 1}{n(n-1)}}} + e^{\ln{n}+2\sigma_{s}^{2} - \frac{1}{2}\ln{\frac{e^{4\sigma_{s}^{2}}-1}{n}}}}} \\
    &= \sqrt{\frac{n}{2\sqrt{1+\frac{1}{e^{2\sigma^{2}_{s}}}}(n-1)^{\frac{3}{2}}+1}}.
\end{align}}}

The derivation in Eqn.~\ref{eqn:log_normal_add} primarily relies on the theorem from \citet{fenton1960sum}, which addresses the sum of log-normal variables. First, we have $e^x \sim \text{LogNormal}(0, \sigma_s^2)$ and $e^{2x} \sim \text{LogNormal}(0, 4\sigma_s^2)$. Additionally, we assume that $e^{x_i}$ and $e^{x_j}$ are independent, leading to $e^{x_i + x_j} \sim \text{LogNormal}(0, 2\sigma_s^2)$. Now, considering the sum of log-normal variables, \citet{fenton1960sum}'s theorem provides that, for moderate values of $\sigma^2$, the sum of zero-mean i.i.d. log-normal variables can be approximated by another log-normal distribution with mean $\mu_\Sigma$ and variance $\sigma_\Sigma^2$, where:
\begin{align*}
    \sigma_\Sigma^2 = \ln\left( \frac{1}{n} \left(e^{\sigma^2} - 1\right) +1 \right); \\ \quad \mu_\Sigma = \ln n + (\sigma^2 - \sigma_\Sigma^2)/2.
\end{align*}

For moderate values of $n$ and $\sigma^2$, the variance $\sigma_{\Sigma}^2$ can be approximated as:
\begin{align*}
\sigma_{\Sigma}^2 \approx \ln\left( \frac{1}{n} \left(e^{\sigma^2} - 1\right) \right).
\end{align*}
Thus, the sum $\sum_{j=1}^{n} e^{2x_j}$ follows a log-normal distribution:

\resizebox{0.5\textwidth}{!}{
$\sum_{j=1}^{n} e^{2x_j} \sim \text{LogNormal}\left(\ln n + 2\sigma_s^2 - \frac{1}{2} \ln\left( \frac{1}{n} \left(e^{4\sigma^2} - 1\right) \right), \ln\left( \frac{1}{n} \left(e^{4\sigma^2} - 1\right)\right)\right).$}

Similarly, the sum $\sum_{i=1}^{n} \sum_{j=1, j \neq i}^{n} e^{x_i + x_j}$ follows:

\resizebox{0.5\textwidth}{!}{
$\sum_{i=1}^{n} \sum_{j=1, j \neq i}^{n} e^{x_i + x_j} \sim \text{LogNormal}\left( \ln n(n-1) + \sigma_s^2 - \frac{1}{2} \ln\left( \frac{e^{2\sigma_s^2} - 1}{n(n-1)} \right),  \ln\left( \frac{e^{2\sigma_s^2} - 1}{n(n-1)} \right)\right).$
}

From these, Eqn.~\ref{eqn:log_normal_add} follows naturally. 
Finally, we have 
\begin{align*}
    \sigma_{a} \leq \sqrt{\frac{n}{2\sqrt{1+\frac{1}{e^{2\sigma^{2}_{s}}}}(n-1)^{\frac{3}{2}}+1}},
\end{align*}
where $\sigma_{a}$ decreases with $n$ increasing. 

\end{proof}

\subsection{Proof of Corollary \ref{thm:length_collapse}} \label{prf:length_collapse}


\begin{theorem}
(Length Collapse in Text Embeddings)
Let $x_1, x_2 \in \mathbb{R}^d$ be text embeddings of two documents with length $n$, decomposed as:
$$
x_i = DC{x_i} + HC{x_i}
$$
Assume $\|HC{x_i}\|_2 \leq \beta(n)$ where $\beta(n) = O(1/\sqrt{n})$, and $\|DC{x_i}\|_2 \geq \alpha > 0$ is a constant. Then the cosine similarity satisfies:
$$
\lim_{n \to \infty} \cos(x_1, x_2) = 1
$$
\end{theorem}

\begin{proof}
    Given the two texts embeddings $\Mat{x}_1$ and $\Mat{x}_2$, we have
    {\small
    \begin{align}
        cos(\Mat{x}_1, \Mat{x}_2) & = \frac{\left( \HC{\Mat{x}_1} + \DC{\Mat{x}_1} \right)  \left( \HC{\Mat{x}_2^T} + \DC{\Mat{x}_2^T} \right)}{ \lVert \Mat{x}_1 \rVert_2 \lVert \Mat{x}_2 \rVert_2} \\
           \label{eqn:cos1} & = \frac{\HC{\Mat{x}_1} \HC{\Mat{x}_2^T}+\DC{\Mat{x}_1} \DC{\Mat{x}_2^T} }{\lVert \Mat{x}_1 \rVert_2 \lVert \Mat{x}_2 \rVert_2} \\
            & \label{eqn:cos2} \geq \frac{\alpha^2}{\sqrt{\alpha^2+\alpha_1^2} \sqrt{\alpha^2 + \alpha_2^2}},
    \end{align}}
where $\alpha_1$ and $\alpha_2$ represent the maximum values in the frequency domain of $\HC{\Mat{x}_1}$ and $\HC{\Mat{x}_2}$ and $\alpha$ represent the value of $\DC{\Mat{x}_1}$ and $\DC{\Mat{x}_2}$, respectively, after applying the discrete Fourier transform. Eqn.~\ref{eqn:cos1} leverages that $\HC{\cdot}$ and $\DC{\cdot}$ are orthogonal. Eqn.~\ref{eqn:cos2} leverages the assumption that the mean of word embeddings in natural language texts maintains a relatively consistent representation. Thus $\HC{\Mat{x}_1}$ and $\HC{\Mat{x}_2}$ have the same value $\alpha$ after applying the discrete Fourier transform. Finally, according to Theorem~\ref{thm:length_rate}, $\alpha_1$ and $\alpha_2$ will gradually decrease with $n$ grows, leading to a higher cosine similarity between $\Mat{x}_1$ and $\Mat{x}_2$.
\end{proof}

\section{More Experiments and Analysis} \label{sec:more_details}

\subsection{Rewriting Process} \label{sec:rewriting_process}
To investigate the differences in embedding distributions across texts of varying lengths, we use the Llama3 (\textit{i.e., }\texttt{Llama-3.1-8B-Instruct})~\cite{dubey2024llama} model to rewrite the texts. Specifically, we used two prompts, ``\textit{Please express the given text in one sentence. No more than 10 tokens. \{\{original text\}\}}'' and ``\textit{Please use few sentences to summarize the given text. \{\{original text\}\}}'', to summarize the texts. This rewriting allows the texts to retain the same semantics while having shorter lengths. By studying the differences in texts of varying lengths, we conclude that the cause of the Length Collapse phenomenon is that longer texts cluster to each other in embedding space.

\begin{figure*}[t]
     \centering
      \begin{subfigure}[b]{0.18\textwidth}
         \centering
         \includegraphics[width=\textwidth]{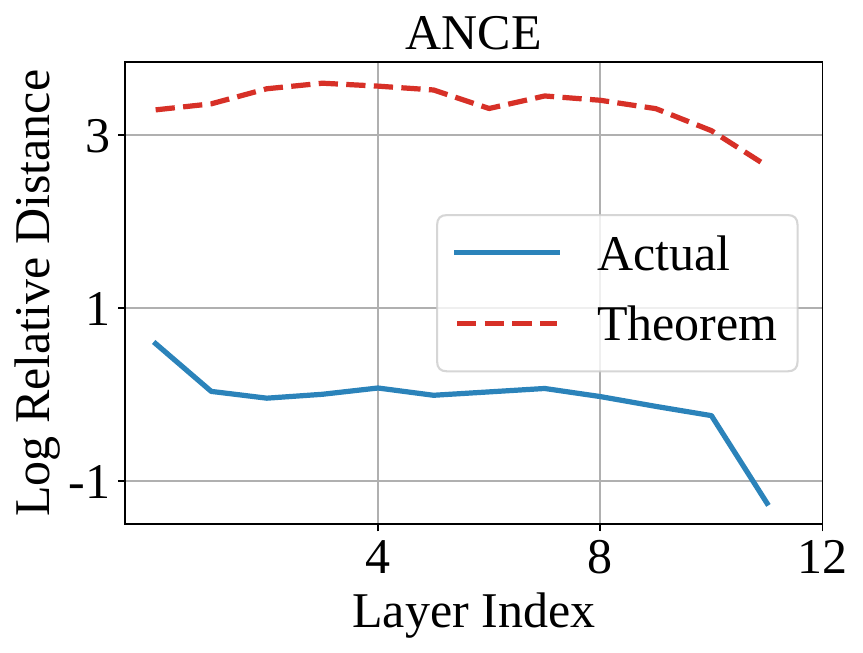}
         \label{fig:theorem2_ance}
     \end{subfigure}
     \hfill
     \begin{subfigure}[b]{0.18\textwidth}
         \centering
         \includegraphics[width=\textwidth]{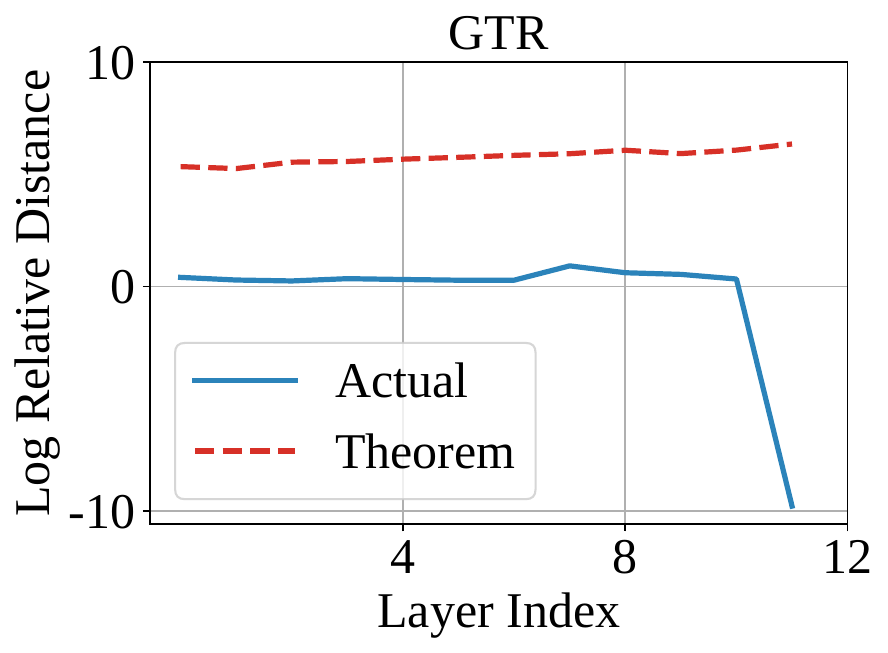}
         \label{fig:theorem2_gtr}
     \end{subfigure}
     \hfill
     \begin{subfigure}[b]{0.18\textwidth}
         \centering
         \includegraphics[width=\textwidth]{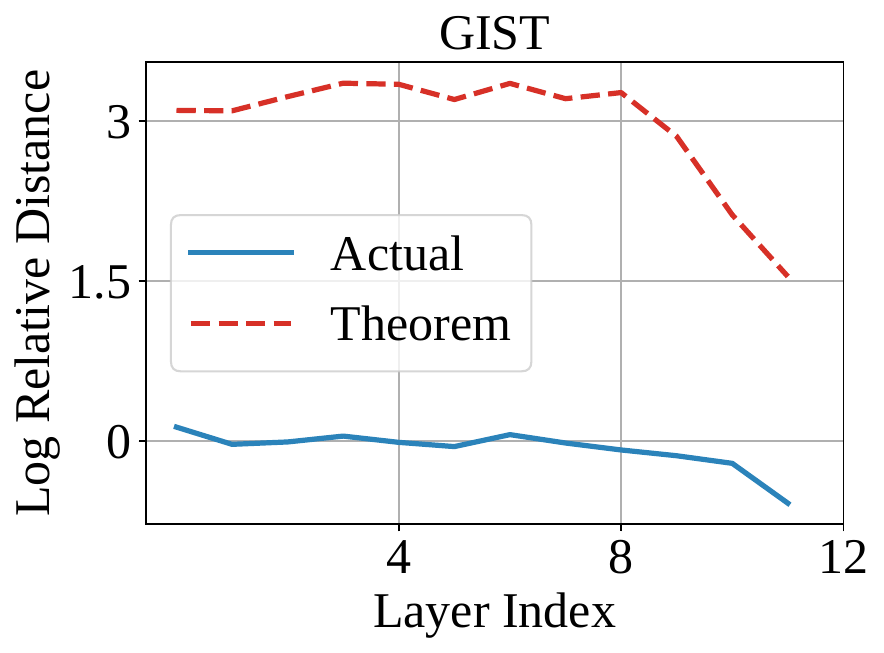}
         \label{fig:theorem2_gist}
     \end{subfigure}
     \hfill
     \begin{subfigure}[b]{0.18\textwidth}
         \centering
         \includegraphics[width=\textwidth]{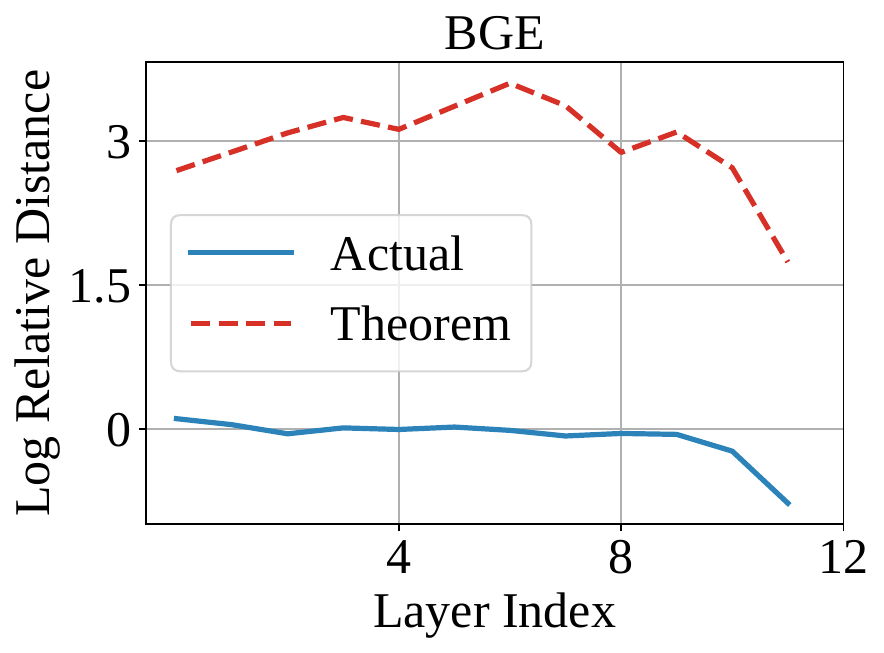}
         \label{fig:theorem2_bge}
     \end{subfigure}
     \hfill
     \begin{subfigure}[b]{0.18\textwidth}
         \centering
         \includegraphics[width=\textwidth]{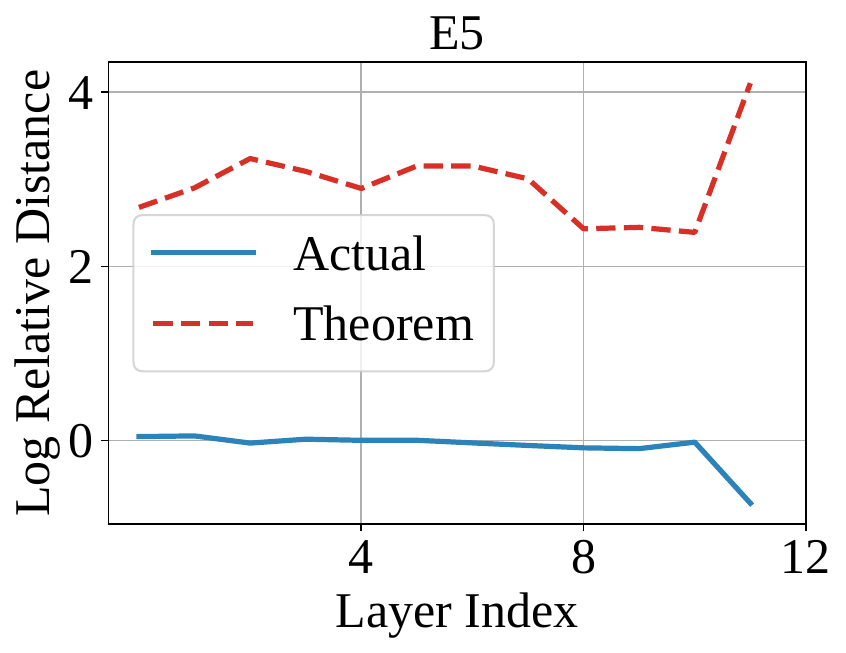}
         \label{fig:theorem2_e5}
     \end{subfigure}
     \vspace{-20pt}
     \caption{Visualization of the intensity of high-frequency components and their theoretical upper bounds. The blue line is defined by $\log(\lVert \HC{\Mat{X}_{l+1}} \rVert_{F}/ \lVert \HC{\Mat{X}_{l}} \rVert_{F} )$, and the red line is estimated using the results in Theorem~\ref{thm:lambda2_sa_rate}.}
     \label{fig:theorem2}
\end{figure*}

\subsection{Details on Figure~\ref{fig:theorem2}}~\label{sec:details_figure_theorem2}
To verify Theorem~\ref{thm:lambda2_sa_rate}, we illustrate the high-frequency intensity of each layer's output along with its theoretical upper limit. Our visualization is based on the ofﬁcial checkpoint of 12-layer ANCE, GIST, and BGE. We use a logarithmic scale for the purpose of a better view. Let $\Mat{X}_l$ denote the output of the $l$-th layer. For red line, we directly calculate $\log(  \lVert \HC{\Mat{X}_{l+1}} \rVert_F / \lVert \HC{\Mat{X}_l} \rVert_F  )$ at each layer. In practice, models typically employ multi-head attention, so we replace $\lVert \Mat{W}_V \rVert_2$ in Eqn.~\ref{eqn:theorem2} with $\sigma_1 H$, where $\sigma_1 = \max_{h=1}^H \lVert \Mat{W}_{V}^h \rVert_2$. For the blue line, we obtain the coefﬁcient  $\sigma_1$ with respect to network parameters and apply the logarithmic scale. To summarize, Figure~\ref{fig:theorem2} implies a convergence rate, which is consistent with our Theorem~\ref{thm:lambda2_sa_rate}.

\begin{figure*}[h]
     \centering
      \begin{subfigure}[b]{0.18\textwidth}
         \centering
         \includegraphics[width=\textwidth]{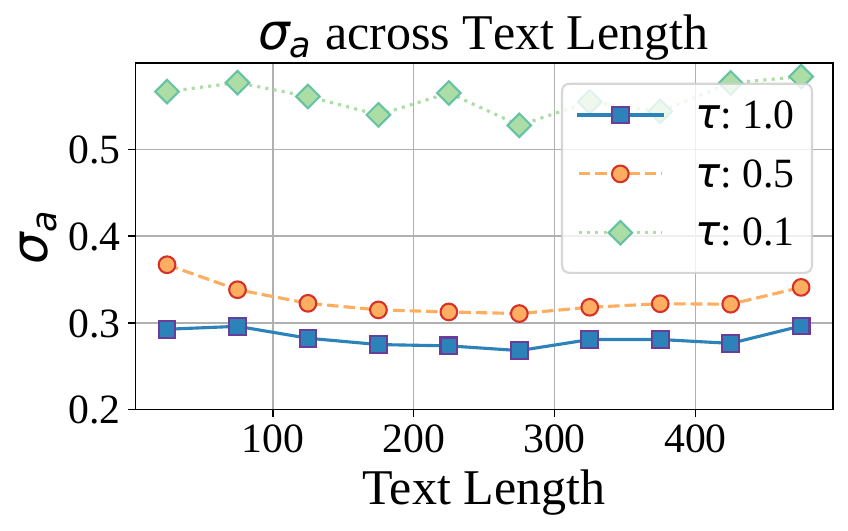}
         \caption{ANCE}
     \end{subfigure}
     \hfill
     \begin{subfigure}[b]{0.18\textwidth}
         \centering
         \includegraphics[width=\textwidth]{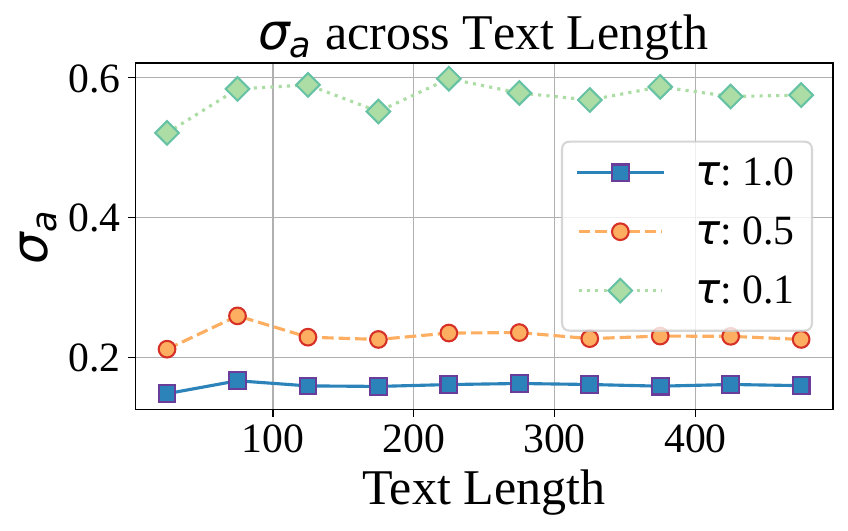}
         \caption{GTR}
     \end{subfigure}
     \hfill
     \begin{subfigure}[b]{0.18\textwidth}
         \centering
         \includegraphics[width=\textwidth]{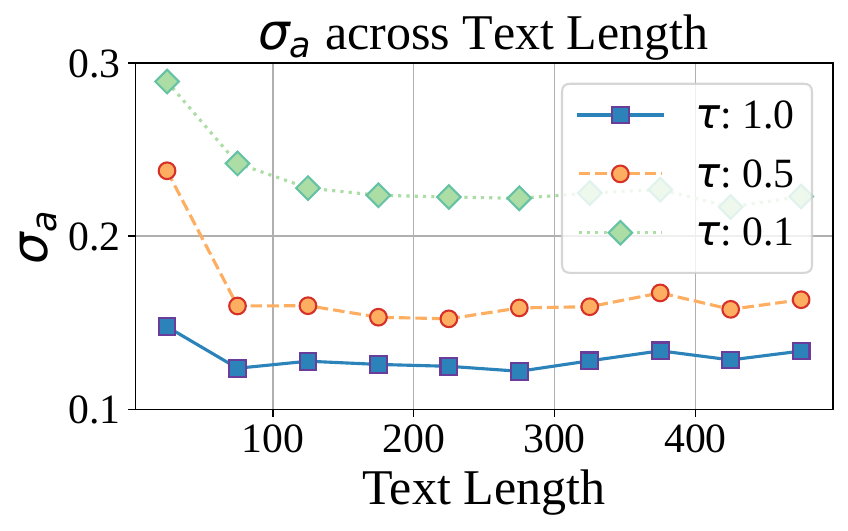}
         \caption{GSIT}
     \end{subfigure}
     \hfill
     \begin{subfigure}[b]{0.18\textwidth}
         \centering
         \includegraphics[width=\textwidth]{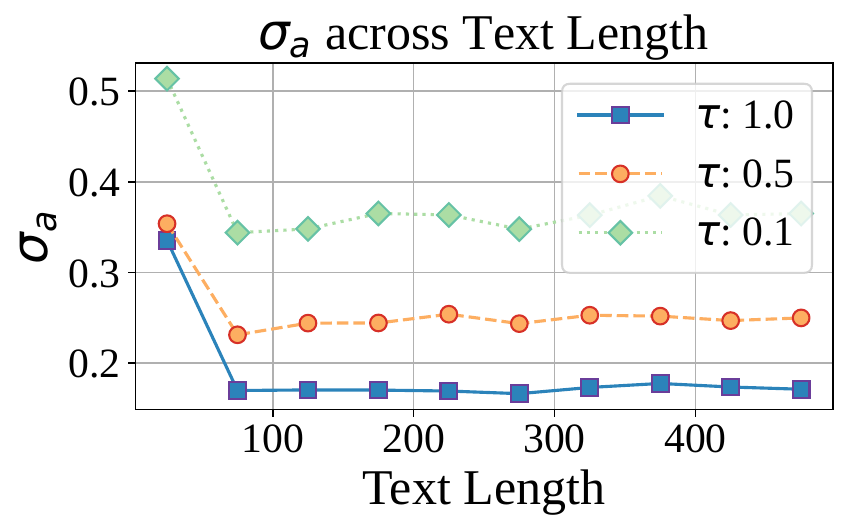}
         \caption{BGE}
     \end{subfigure}
     \hfill
     \begin{subfigure}[b]{0.18\textwidth}
         \centering
         \includegraphics[width=\textwidth]{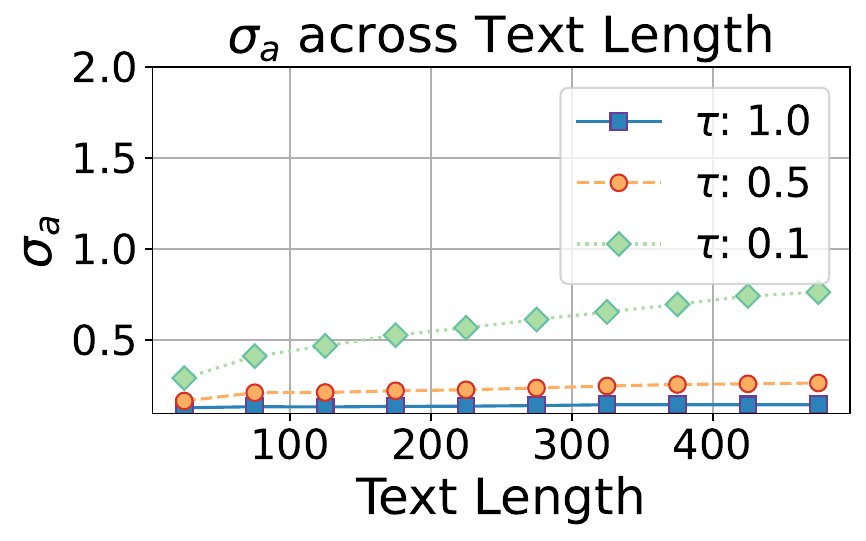}
         \caption{E5}
     \end{subfigure}

     \caption{$\sigma_a$ of $\HC{\Mat{A}}$ before the last layer across different text length under different $\tau$ setting.}
     \label{fig:more_results_on_theorem3}
     \vspace{-5pt}
\end{figure*}

\subsection{Details on Figure~\ref{fig:more_results_on_theorem3}}~\label{sec:details_figure_theorem3}
To verify Theorem~\ref{thm:length_rate}, we visualize the value of $\sigma_{a}$
across different text lengths. Our visualization is based on the texts from NFCorpus. We sample 100 samples for each bin from 0 to 500 with a bin size of 50. The $\sigma_{a}$ value is computed as the average of the $\sigma_{a}$ based on the attention of all heads before the output of the final layer. To summarize, Figure~\ref{fig:more_results_on_theorem3} implies that $\sigma_a$ shows a decreasing trend as $n$ increases, which is consistent with our Theorem~\ref{thm:length_rate}. Moreover, $\sigma_{a}$ also increases as $\tau$ decreases, further validating our proposed method, TempScale. E5 model shows an increasing $\sigma_a$ with length, which may be due to anomalies in the deep layer attention patterns.

\subsection{More Analysis about Assumption in Theorem~\ref{thm:length_collapse}} \label{sec:details_theorem4_assumption}

\begin{figure*}[h]
     \centering
      \begin{subfigure}[b]{0.19\textwidth}
         \centering
         \includegraphics[width=\textwidth]{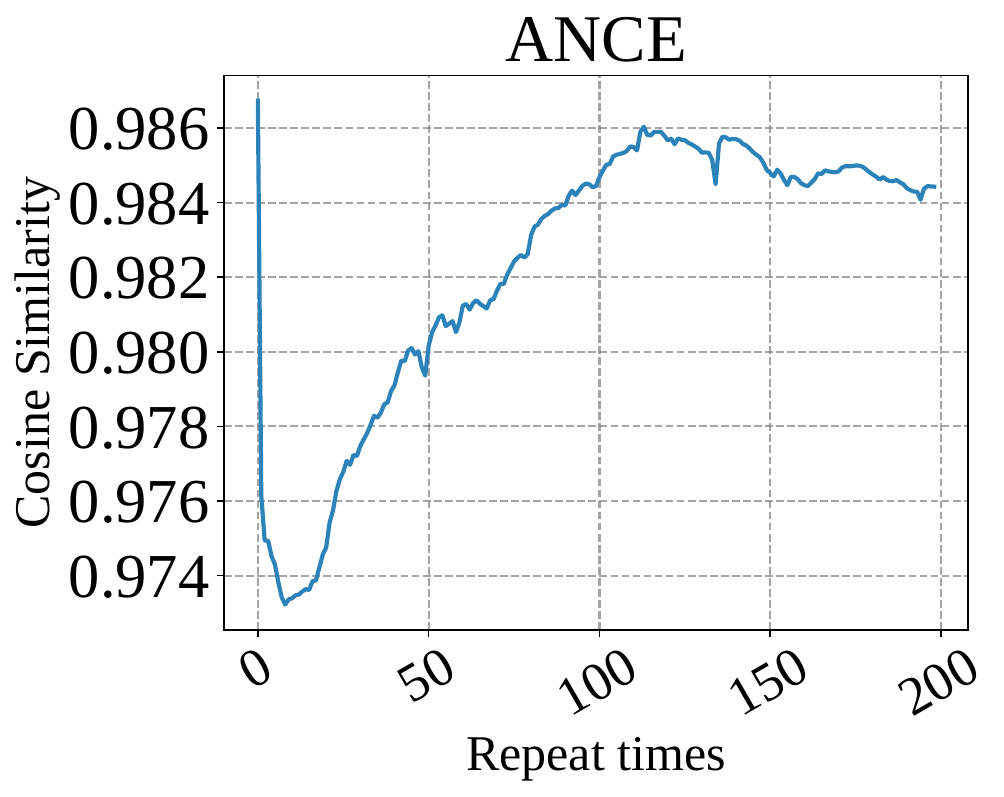}
     \end{subfigure}
     \hfill
     \begin{subfigure}[b]{0.185\textwidth}
         \centering
         \includegraphics[width=\textwidth]{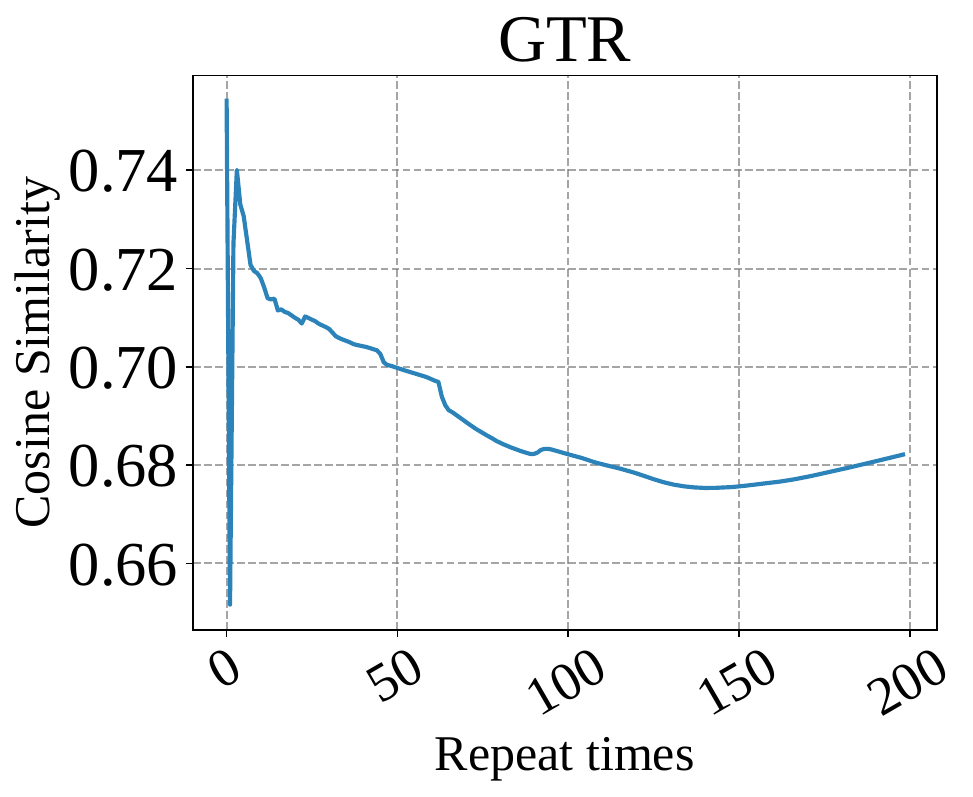}
     \end{subfigure}
     \hfill
     \begin{subfigure}[b]{0.185\textwidth}
         \centering
         \includegraphics[width=\textwidth]{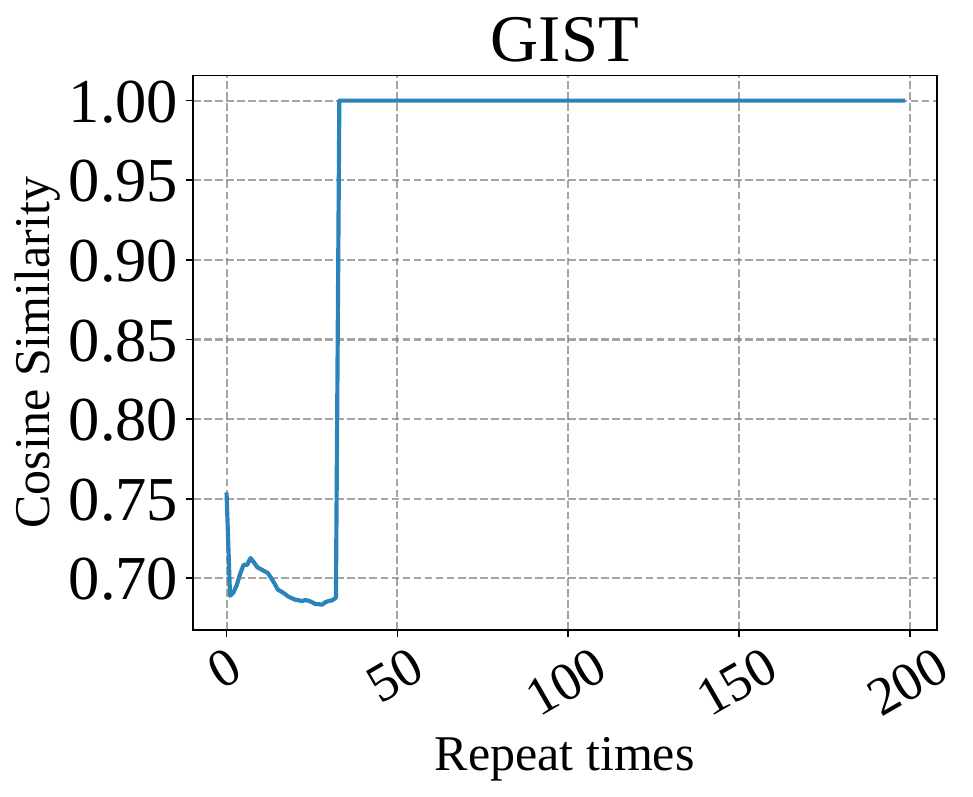}
     \end{subfigure}
     \hfill
     \begin{subfigure}[b]{0.18\textwidth}
         \centering
         \includegraphics[width=\textwidth]{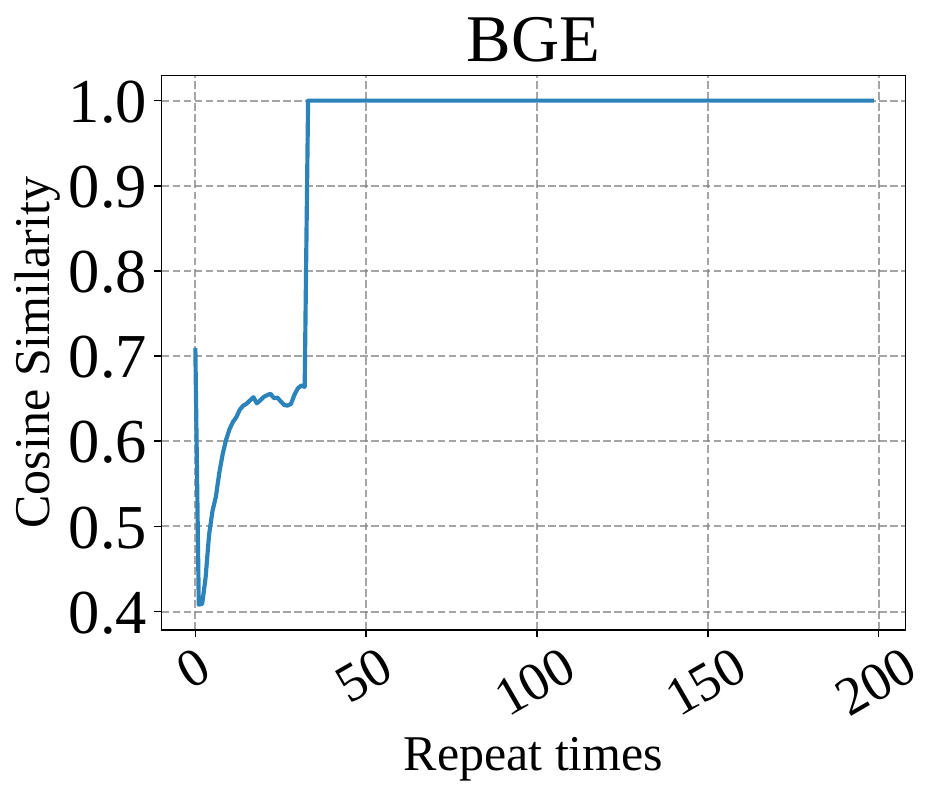}
     \end{subfigure}
     \hfill
     \begin{subfigure}[b]{0.185\textwidth}
         \centering
         \includegraphics[width=\textwidth]{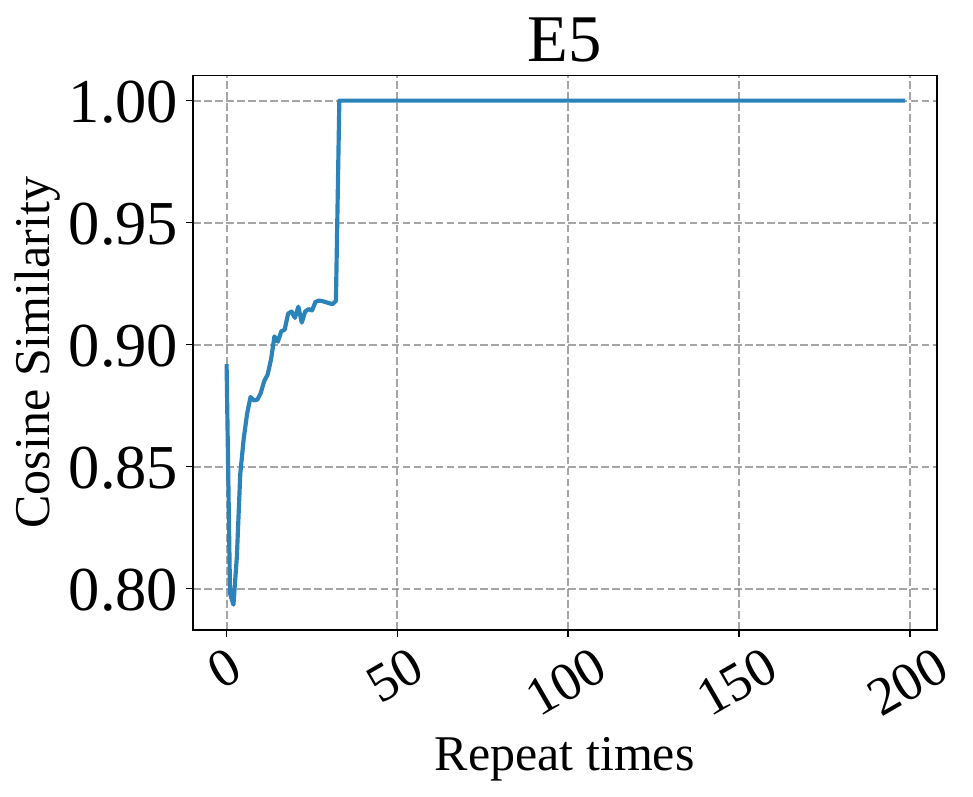}
     \end{subfigure}

     \caption{The cosine similarity of embeddings of texts generated by repeating words ``dog'' and ``cat''.}
     \label{fig:theorem4_assumption}
     \vspace{-10pt}
\end{figure*}

In Corollary~\ref{thm:length_collapse}, we hypothesize and verify that all-natural language sequences tend to have a relatively consistent representation. As a result, different texts tend to exhibit consistent low-pass signals after losing high-frequency information, leading to ultimately consistent text embeddings. This leads to an increase in cosine similarity for longer texts. However, as shown in Figure~\ref{fig:theorem4_assumption}, we repeat the words ``dog'' and ``word'' $n$ times and calculate the similarity between these two texts. The results show that even when the sequences do not overlap, the text embeddings tend to converge to similar representations as the sequence length increases. This further demonstrates that Length Collapse causes completely different sequences to converge toward similarity.

\subsection{More Discussion about Other Works}~\label{sec:more_discussionts}
\textbf{Other Components in Transformer.} After discussing how self-attention contributes to Length Collapse, we proceed to examine the influence of other modules in the transformer, such as multi-head, residual, and FFN. Fortunately, previous work~\cite{wanganti} has talked about whether these components can effectively alleviate the low-pass ﬁltering drawbacks. The proof demonstrates that while these components help preserve high-frequency signals, they do not alter the fact that the MSA block, as a whole, functions solely with the representational power of a low-pass filter. Furthermore, the ability of these models to preserve high-frequency signals is solely determined by their internal parameters and architecture, independent of the input text length. As a result, these modules do not impact our analysis of Length Collapse in practical models.

\textbf{Difference from Over-Smoothing in Deeper Layers.} In previous research~\cite{wanganti}, it has been noted that a self-attention module acts as a low-pass filter, causing input feature maps to gradually lose high-frequency signals as the model layers go deeper. Furthermore, other studies~\cite{oonograph,cai2020note} indicate that the node features of Graph Convolutional Networks (GCNs) can become exponentially trapped in the null space of the graph Laplacian matrix. The root cause of this phenomenon is that both graph Laplacian matrices and self-attention matrices consistently exhibit a dominant eigenvector, commonly referred to as the DC component. While these studies address over-smoothing in deeper layers, we focus on how the low-pass filtering process changes as the input sequence lengthens, specifically examining over-smoothing in longer sequences.

\subsection{More discussions on Classification and Clustering Tasks}~\label{sec:discussions_on_class}
\begin{figure*}[h]
\centering
    \begin{minipage}{0.48\textwidth}
        \includegraphics[width=\linewidth]{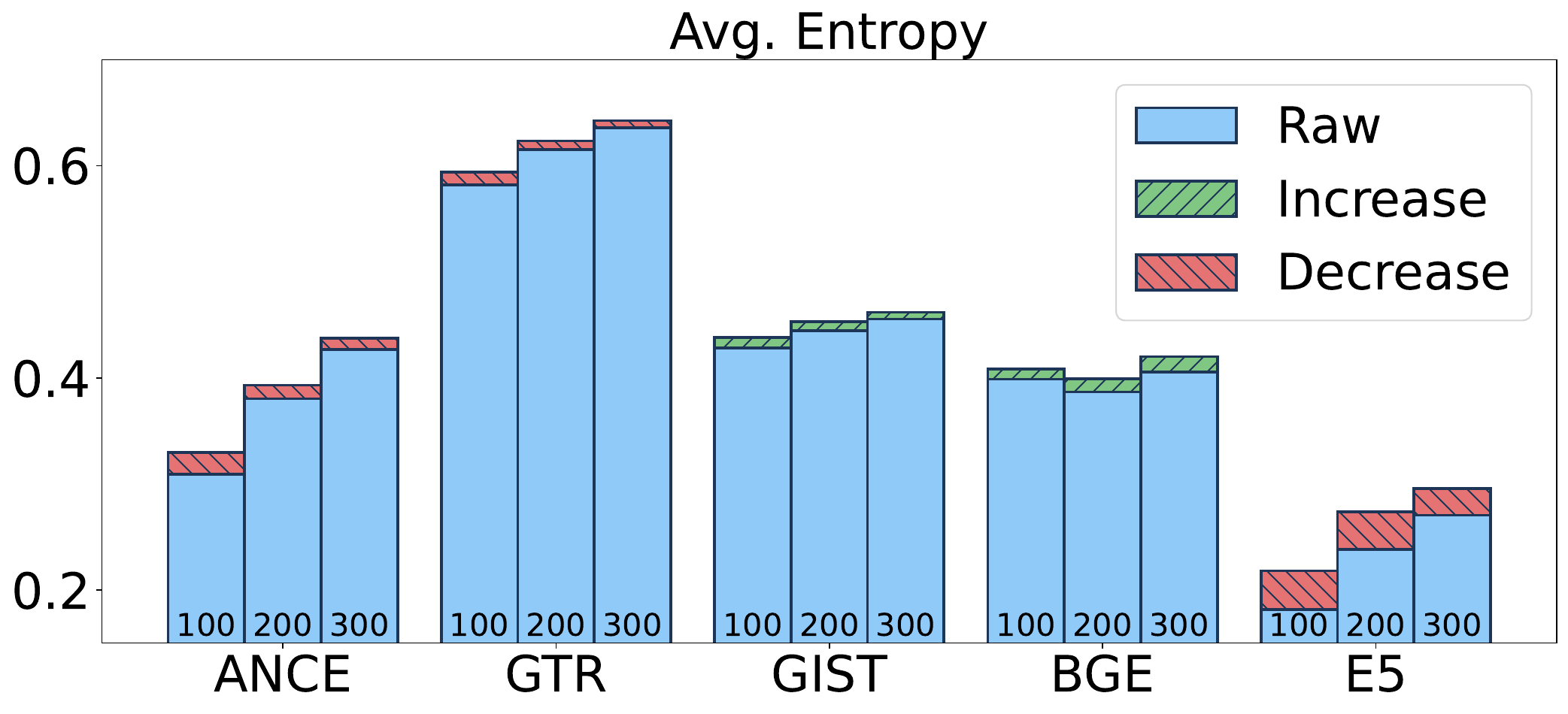}
        \label{fig:entropy}
    \end{minipage} \hfill
    \centering
    \begin{minipage}{0.48\textwidth}
        \includegraphics[width=\linewidth]{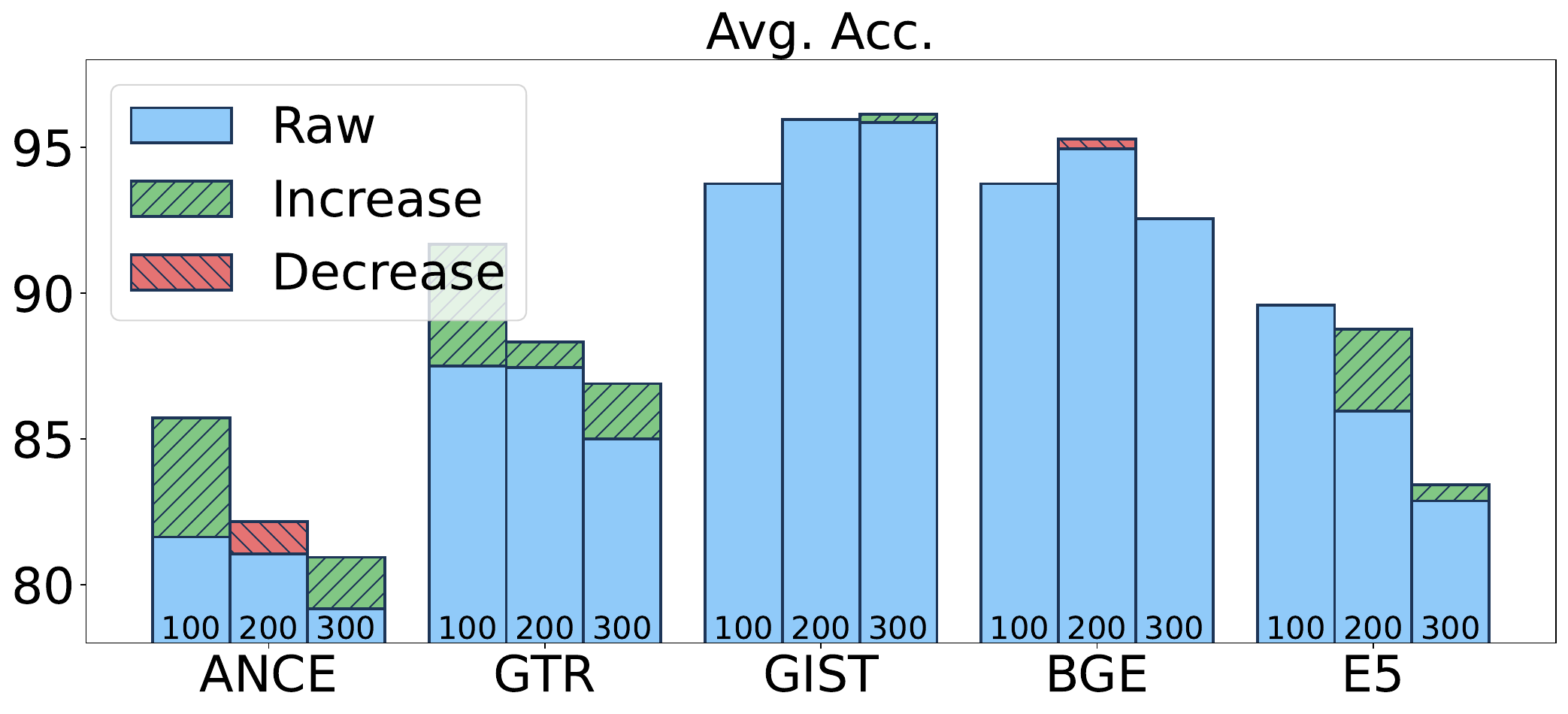}
        \label{fig:classification}
    \end{minipage}
    \vspace{-20pt}
    \caption{Probability entropy and classification accuracy of models across different length intervals on AmazonPolarity dataset. Each model has bars representing intervals of 100 in length, with 500 text samples per interval, covering a range from 0 to 300. Bars represent raw outputs, with green and red hatching indicating increases and decreases after TempScale, respectively.}
    \label{fig:data_group}
    \vspace{-10pt}
\end{figure*}

For data grouping tasks like classification and clustering, in the case of $N$-class tasks, we can think of the model as learning $N$ classification boundaries. The farther the text embedding is from the boundary, the closer the output probability approaches 1 or 0. As shown in Figure~\ref{fig:data_group} (left), we plot the entropy of the model output probabilities across different length intervals. The model outputs higher entropy for longer texts, which may be because the embeddings of longer texts are positioned closer to the center of the space as described in Figure~\ref{fig:intro_tsnet}, resulting in a shorter distance to various classification boundaries. Meanwhile, in Figure~\ref{fig:data_group} (right), we can also observe that accuracy and entropy follow the same trend: the model achieves higher accuracy when it has lower entropy. In other words, the model performs better if the text embeddings are farther from the boundary. After applying TempScale, a decreased entropy will result in increased accuracy. This further supports the relationship between entropy and accuracy in classification tasks. Moreover, if the model exhibits a more severe Length Collapse phenomenon, meaning a greater performance drop on longer texts, the more performance improvement it experiences after applying TempScale. This suggests that one possible reason TempScale is effective on shorter text datasets is that it adjusts the distribution differences between texts of varying lengths. As shown in Figure~\ref{fig:intro_cosine}, after applying TempScale, the distribution of text embeddings across different length intervals becomes more uniform, which facilitates the model in learning length-agnostic parameters.

\subsection{More Results on Longer Texts} \label{sec:more_long_queries}

\begin{figure*}[h]
     \centering
      \begin{subfigure}[b]{0.23\textwidth}
         \centering
         \includegraphics[width=\textwidth]{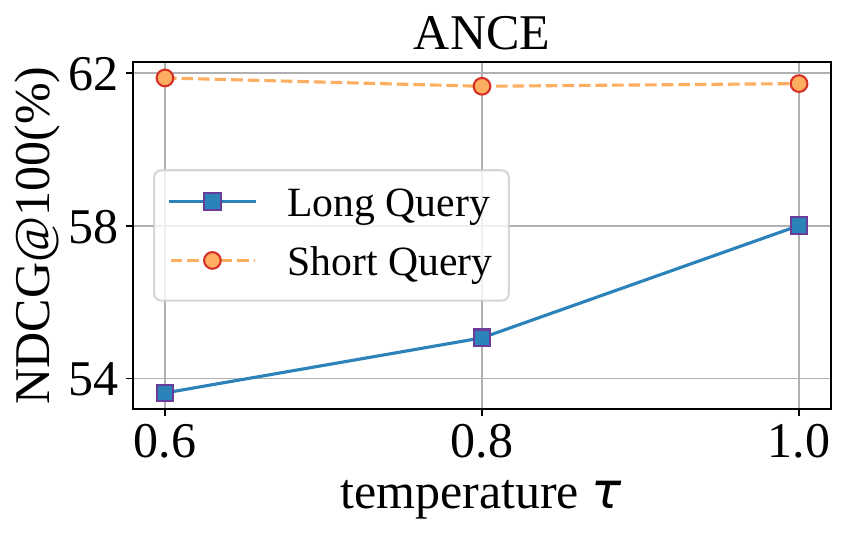}
     \end{subfigure}
     \hfill
     \begin{subfigure}[b]{0.23\textwidth}
         \centering
         \includegraphics[width=\textwidth]{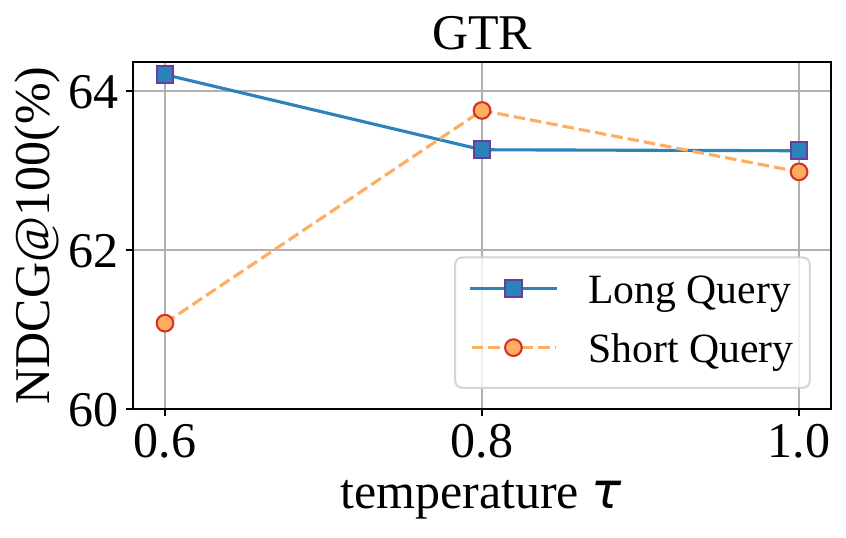}
     \end{subfigure}
     \hfill
     \begin{subfigure}[b]{0.23\textwidth}
         \centering
         \includegraphics[width=\textwidth]{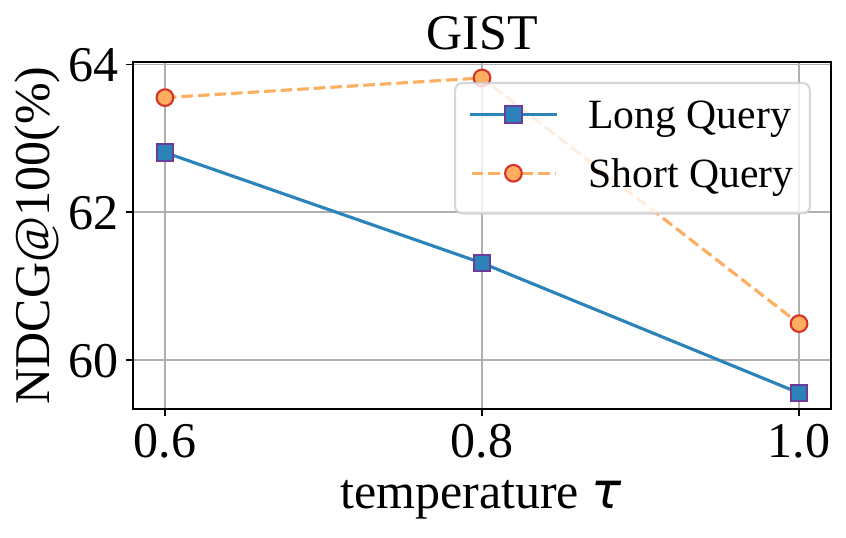}
     \end{subfigure}
     \hfill
     \begin{subfigure}[b]{0.23\textwidth}
         \centering
         \includegraphics[width=\textwidth]{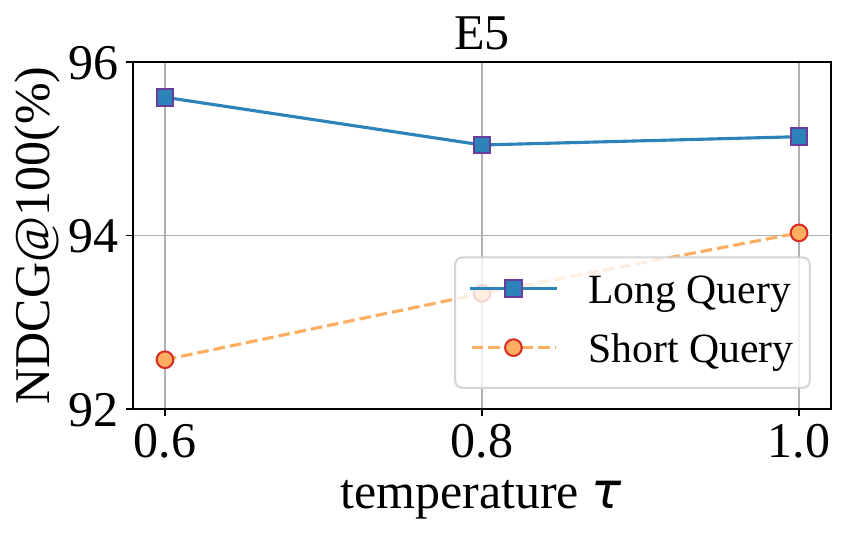}
     \end{subfigure}
     \caption{Results about performance difference between long query and short query across varying temperature $\tau$ on SummScreenFD dataset.}
     \vspace{-5pt}
     \label{fig:more_results_long_query}
\end{figure*}

\begin{figure*}[t!]
    \vspace{-5pt}
    \centering
    \begin{minipage}{0.3\textwidth}
        \includegraphics[width=\linewidth]{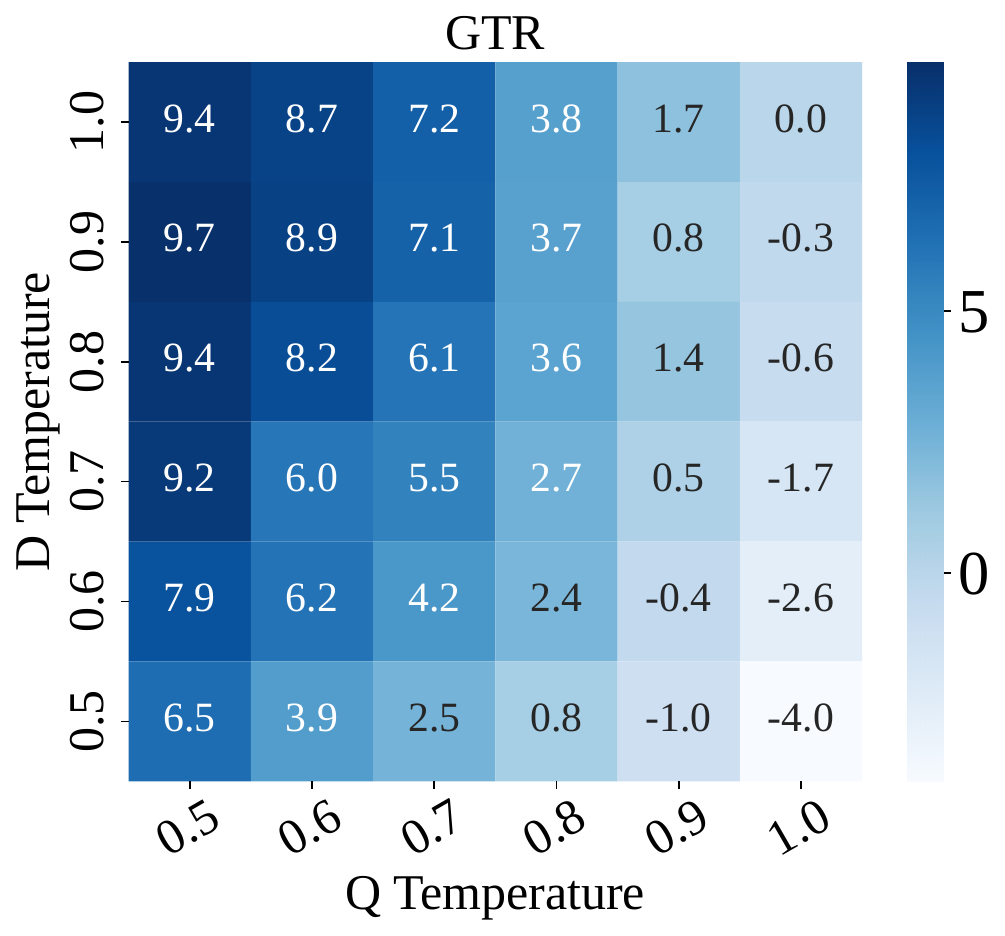}
        \label{fig:heatmap_GTR}
    \end{minipage} \hfill
    \centering
    \begin{minipage}{0.3\textwidth}
        \includegraphics[width=\linewidth]{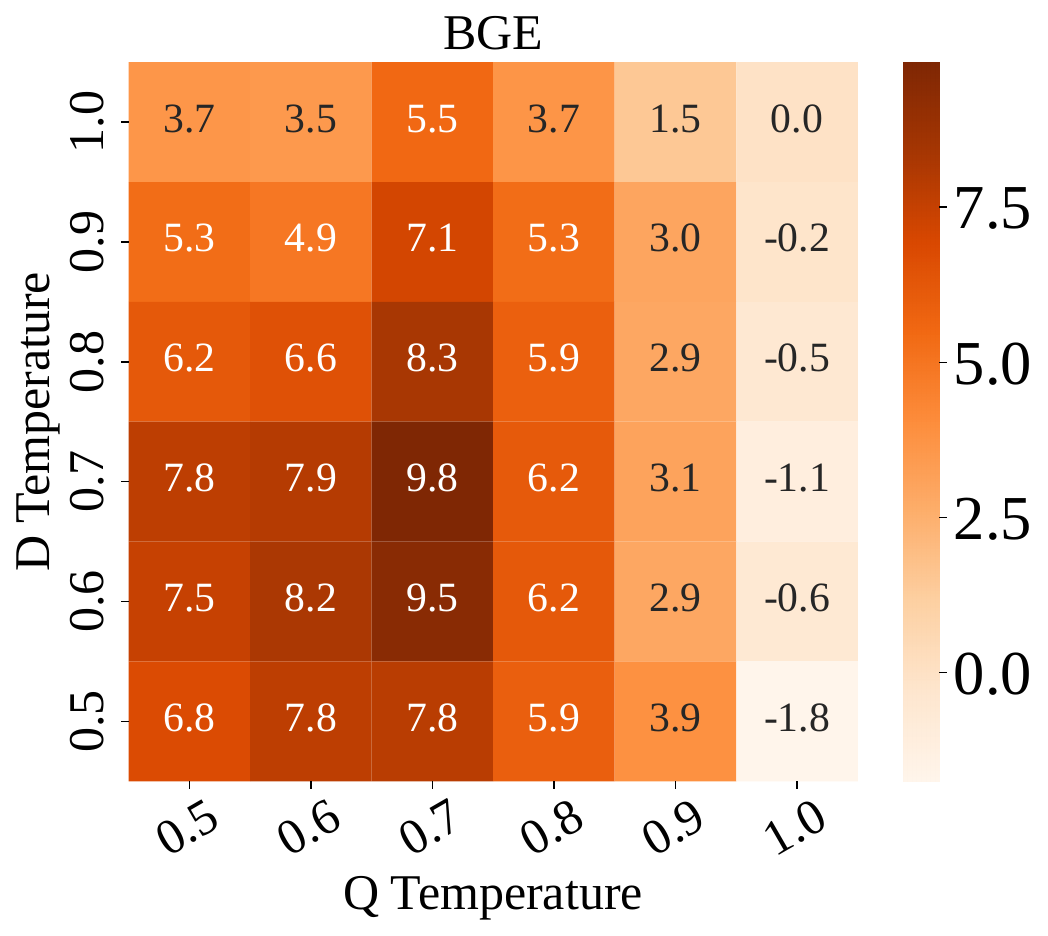}
        \label{fig:heatmap_GIST}
    \end{minipage} \hfill
    \centering
    \begin{minipage}{0.3\textwidth}
        \includegraphics[width=\linewidth]{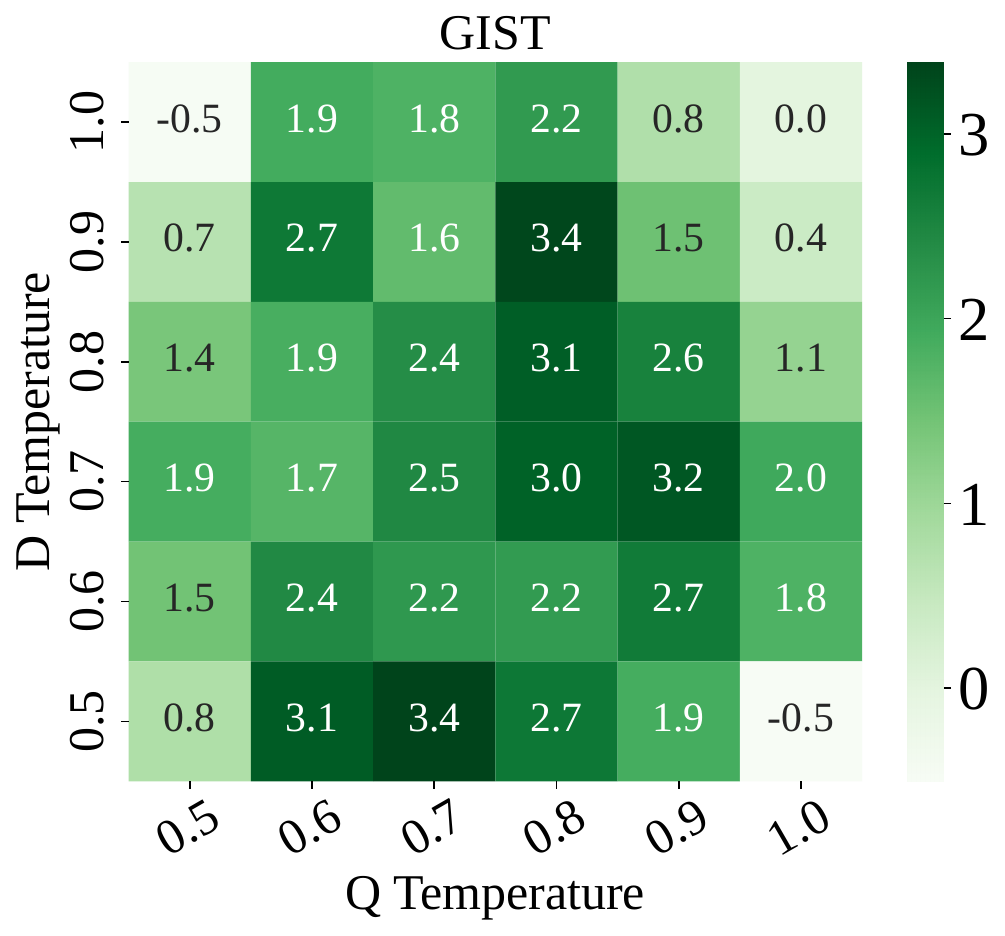}
        \label{fig:heatmap_BGE}
    \end{minipage} \hfill
    \vspace{-10pt}
    \caption{Relative performance compared to the raw results with varying query (Q Temperature) and document (D Temperature) temperatures using different models on QMSum.}
    \label{fig:heatmap}
\end{figure*}

\begin{figure*}[h]
    \vspace{-5pt}
     \centering
      \begin{subfigure}[b]{0.23\textwidth}
         \centering
         \includegraphics[width=\textwidth]{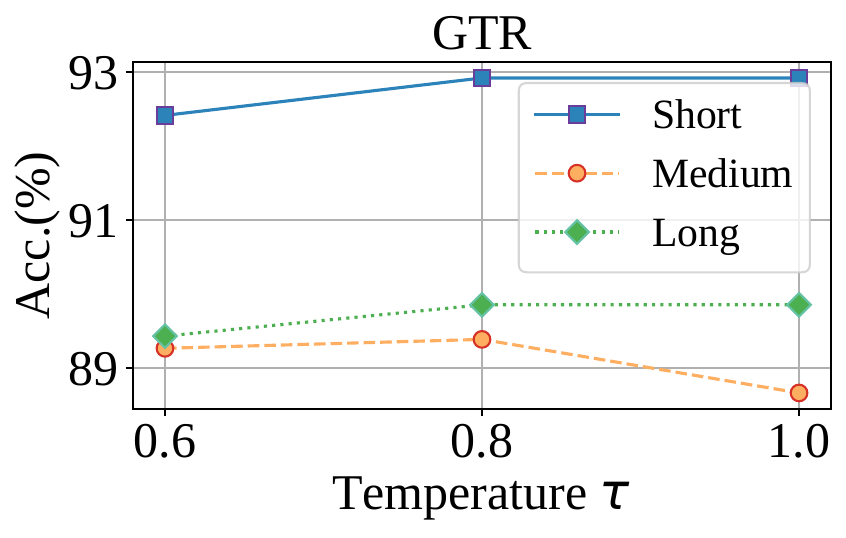}
     \end{subfigure}
     \hfill
     \begin{subfigure}[b]{0.23\textwidth}
         \centering
         \includegraphics[width=\textwidth]{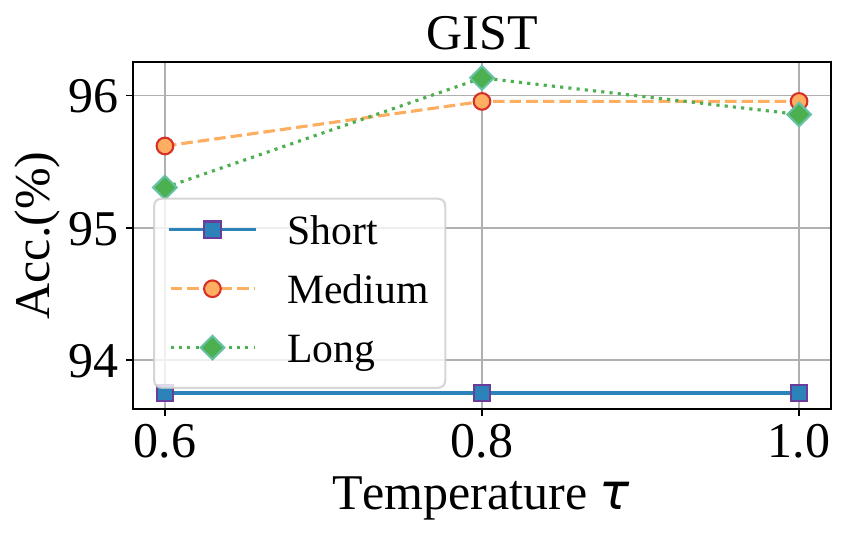}
     \end{subfigure}
     \hfill
     \begin{subfigure}[b]{0.23\textwidth}
         \centering
         \includegraphics[width=\textwidth]{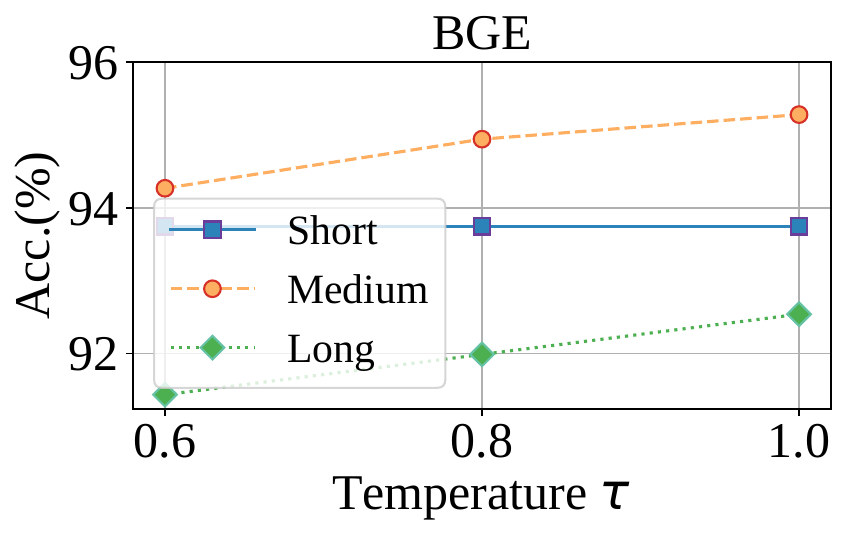}
     \end{subfigure}
     \hfill
     \begin{subfigure}[b]{0.23\textwidth}
         \centering
         \includegraphics[width=\textwidth]{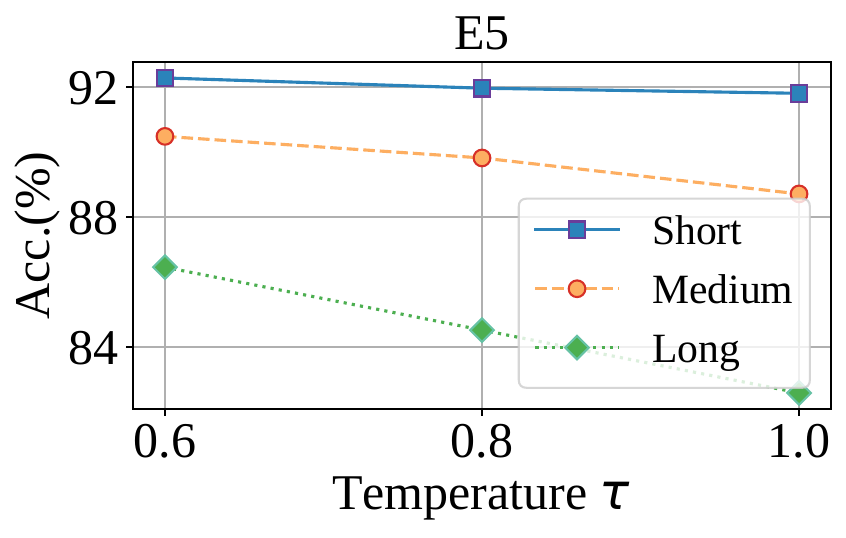}
     \end{subfigure}

     \caption{Results about performance difference between long query and short query across varying temperature $\tau$ on AmazonPolarity dataset.}
     \vspace{-10pt}
     \label{fig:more_results_classification_long_text}
\end{figure*}

\begin{figure*}[h]
    \centering
    \begin{minipage}{0.3\textwidth}
        \includegraphics[width=\linewidth]{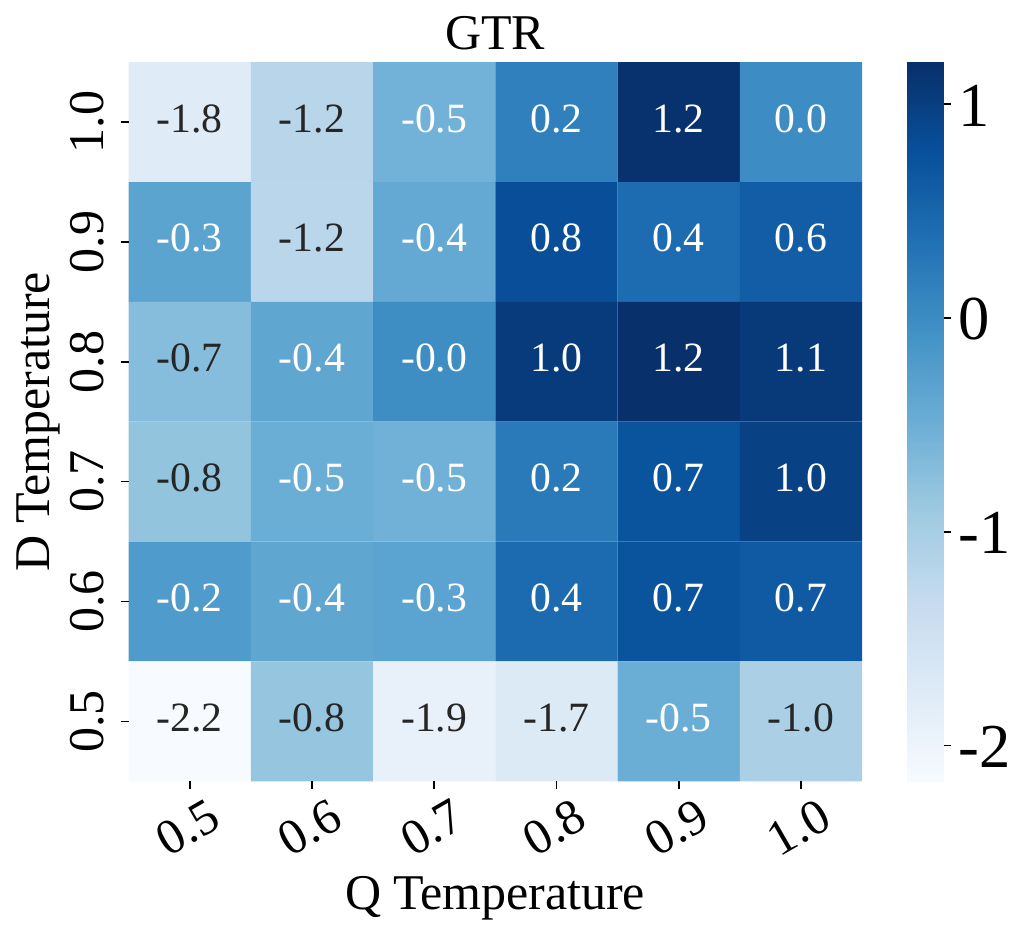}
        \label{fig:heatmap_GTR_summ}
    \end{minipage} \hfill
    \centering
    \begin{minipage}{0.3\textwidth}
        \includegraphics[width=\linewidth]{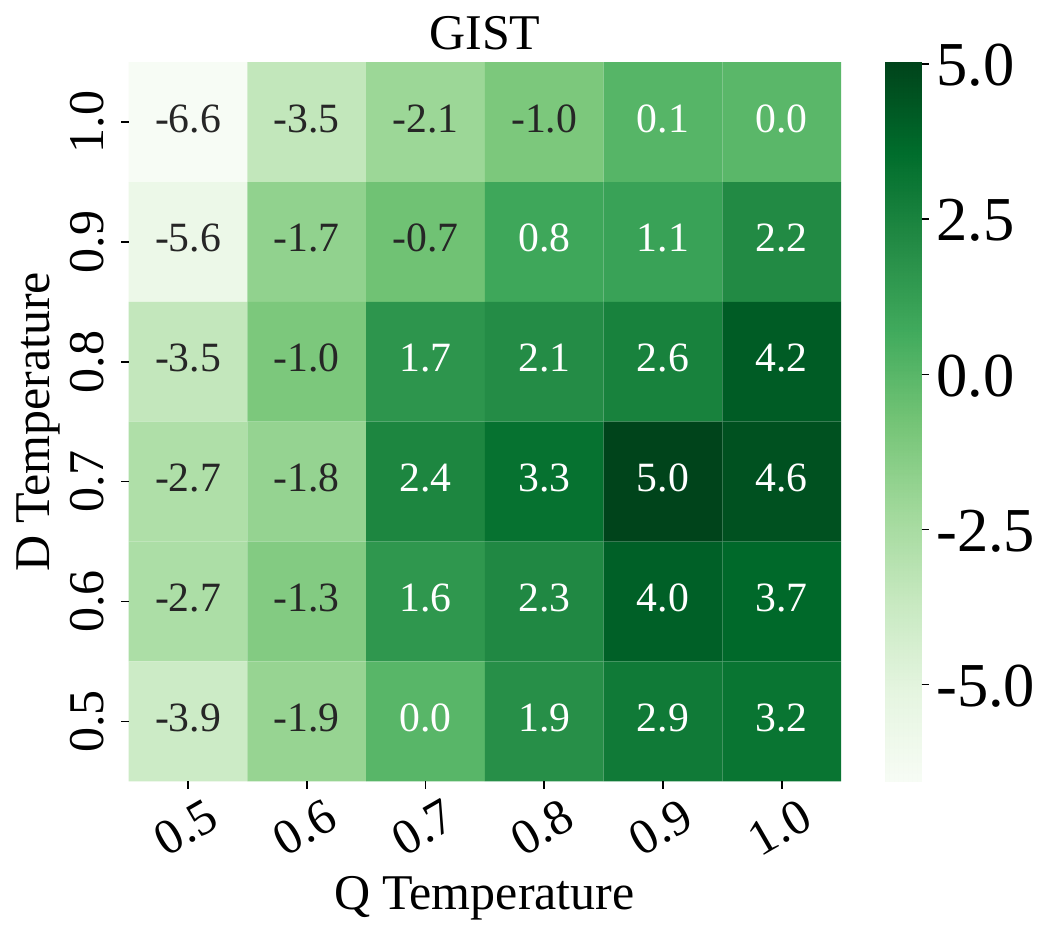}
        \label{fig:heatmap_GIST_summ}
    \end{minipage} \hfill
    \centering
    \begin{minipage}{0.3\textwidth}
        \includegraphics[width=\linewidth]{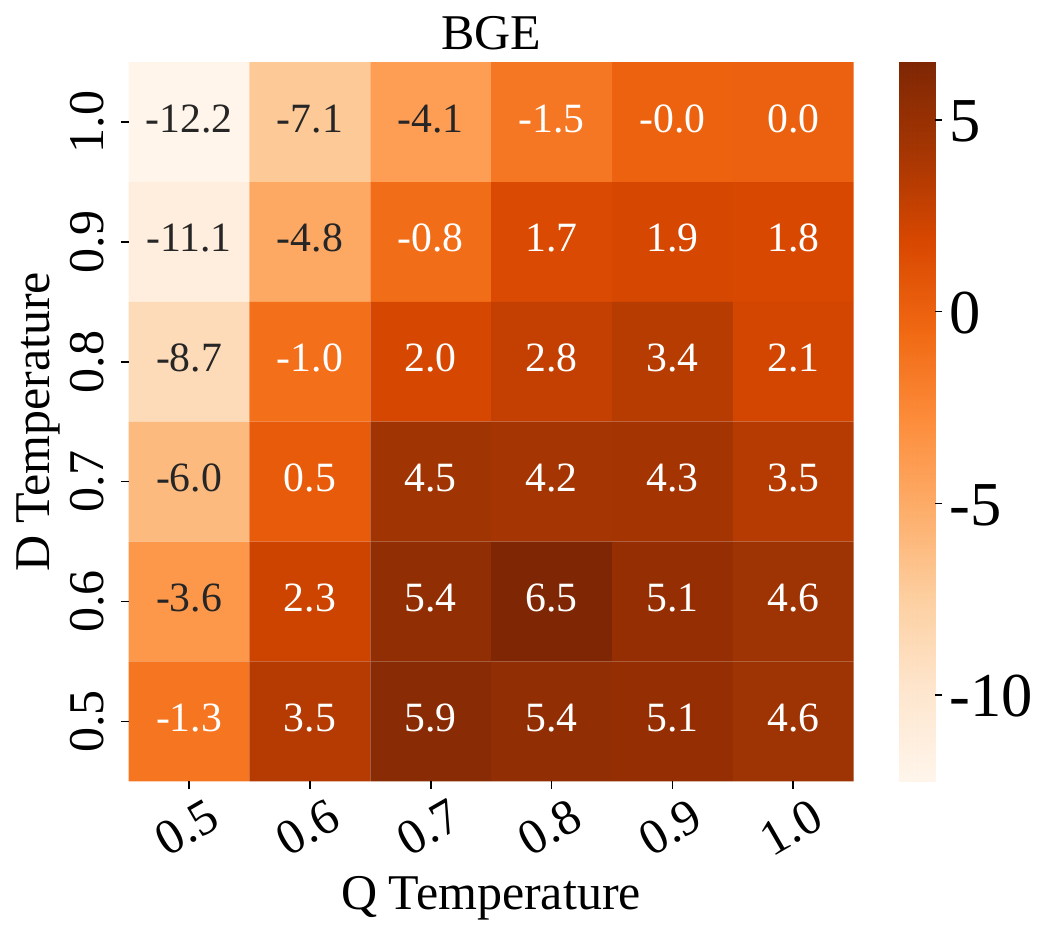}
        \label{fig:heatmap_BGE_summ}
    \end{minipage}
    \vspace{-20pt}
    \caption{Relative performance compared to the raw results with varying query (Q Temperature) and document (D Temperature) temperatures using different models on SummScreenFD}
    \vspace{-10pt}
    \label{fig:more_heatmap}
\end{figure*}

\textbf{Can the temperature be set based on the length of the text?} In the previous experiments, we use the same temperature $\tau$ for scaling all texts in the same task. However, our analysis indicates that texts of different lengths have varying filtering rates, so a natural idea is to use different temperatures for texts of different lengths. As shown in Figure~\ref{fig:more_results_classification_long_text}, we plot the performance trend for texts under the same settings in Figure~\ref{fig:data_group} as the temperature varies. The results indicate that a higher temperature is optimal for short texts, while a lower temperature is preferable for long texts, as confirmed by the results in Table~\ref{tab:longdoc}. When performing retrieval tasks, we can also set different temperatures for queries and documents to achieve better performance. As shown in Figure~\ref{fig:heatmap}, on the QMSum dataset, we can consistently achieve better performance by setting a lower temperature for queries. Moreover, as shown in Figure~\ref{fig:more_heatmap}, compared to the QMSum dataset, SummScreenFD requires a lower temperature for scaling the document due to its longer length. This further supports the conclusion that longer texts require a lower temperature for scaling. In Figure~\ref{fig:more_results_long_query}, we observe a similar phenomenon across other models. Except for ANCE, the performance of other models on long queries decreases as the temperature decreases. This suggests that long queries require a lower temperature to mitigate the Length Collapse phenomenon. However, ANCE's performance degradation with decreasing temperature is likely attributable to its inherent limitations in processing long texts.

\section{Length Collapse in LLM-based Embedding Models}~\label{sec:llm_length_collapse}
To explore whether Length Collapse also occurs in long-context LLMs, we select three widely used LLM-based embedding models from the MTEB benchmark—bge-multilingual-gemma2~\cite{chen2024bge}, NV-Embed-v2~\cite{lee2024nv}, and e5-mistral-7b-instruct~\cite{wang2022text}—all of which rank within the top 20 on the MTEB leaderboard. We conduct experiments using three commonly used long-text datasets, including nq and hotpotqa in LongRAG~\cite{jiang2024longrag} and LongAlpaca-12k~\cite{chenlonglora}, which are selected based on relevance to the keyword "long" and chosen for their evenly distributed text lengths. Specifically, we analyze shifts in embedding space across different text length intervals by calculating the average Euclidean distance of each embedding from the central embedding (the mean of all embeddings) and computing cosine similarity between each pair of embeddings as in Table~\ref{tab:llm_distance} and Table~\ref{tab:llm_cosine}.

\begin{table*}[h!]
\centering
\begin{tabular}{llccc}
\toprule
\textbf{Dataset} & \textbf{Model} & \textbf{0-1000} & \textbf{1000-2000} & \textbf{2000-3000} \\
\midrule
\multirow{3}{*}{nq} & bge-multilingual-gemma2 & 0.94 & 0.92 & 0.91 \\
& NV-Embed-v2 & 0.98 & 0.91 & 0.89 \\
& e5-mistral-7b-instruct & 0.68 & 0.64 & 0.64 \\
\midrule
\multirow{3}{*}{hotpotqa} & bge-multilingual-gemma2 & 0.93 & 0.88 & 0.87 \\
& NV-Embed-v2 & 0.95 & 0.84 & 0.79 \\
& e5-mistral-7b-instruct & 0.70 & 0.61 & 0.57 \\
\midrule
\multirow{3}{*}{LongAlpaca-12k} & bge-multilingual-gemma2 & 0.71 & 0.48 & 0.49 \\
& NV-Embed-v2 & 0.89 & 0.86 & 0.87 \\
& e5-mistral-7b-instruct & 0.69 & 0.53 & 0.53 \\
\bottomrule
\end{tabular}
\caption{Euclidean distance results for different datasets and LLM-based embedding models.}~\label{tab:llm_distance}
\end{table*}

\begin{table*}[h!]
\centering
\begin{tabular}{llccc}
\toprule
\textbf{Dataset} & \textbf{Model} & \textbf{0-1000} & \textbf{1000-2000} & \textbf{2000-3000} \\
\midrule
\multirow{3}{*}{nq} & bge-multilingual-gemma2 & 0.39 & 0.42 & 0.45 \\
& NV-Embed-v2 & 0.32 & 0.47 & 0.59 \\
& e5-mistral-7b-instruct & 0.75 & 0.77 & 0.78 \\
\midrule
\multirow{3}{*}{hotpotqa} & bge-multilingual-gemma2 & 0.51 & 0.52 & 0.59 \\
& NV-Embed-v2 & 0.48 & 0.59 & 0.69 \\
& e5-mistral-7b-instruct & 0.75 & 0.81 & 0.85 \\
\midrule
\multirow{3}{*}{LongAlpaca-12k} & bge-multilingual-gemma2 & 0.76 & 0.90 & 0.89 \\
& NV-Embed-v2 & 0.48 & 0.65 & 0.68 \\
& e5-mistral-7b-instruct & 0.78 & 0.87 & 0.87 \\
\bottomrule
\end{tabular}
\caption{Pair cosine similarities results for different datasets and LLM-based embedding models.}~\label{tab:llm_cosine}
\end{table*}

The experimental results show that as text length increases, the embeddings from different LLM-based embedding models exhibit a gradual convergence trend, with the average Euclidean distance between embeddings decreasing and pairwise cosine similarity increasing. This indicates that even mainstream long-context LLM embedding models tend to experience embedding convergence (Length Collapse) and reduced distinctiveness in long-text processing due to the low-pass filtering effect.

\section{More Discussions about TempScale}~\label{sec:discussions_tempscale}
\subsection{Relationship with Contrastive Learning}~\label{sec:contrastive_learning}
\textbf{Background and Motivation.} Contrastive learning is widely used to address embedding space anisotropy, reducing high similarity between random text samples of varying lengths by maximizing distances among negative pairs and aligning positive pairs. While it improves embedding quality in many applications, its impact on Length Collapse remains unexplored. This section examines how contrastive learning affects Length Collapse, assessing its contributions and limitations.

As shown in previous works~\cite{wang2020understanding}, the InfoNCE loss, when scaled to a large number of negative samples, can be decomposed into two primary components: \textbf{Alignment} and \textbf{Uniformity}. Alignment ensures that positive pairs—texts with similar content—are close in the embedding space, while Uniformity spreads embeddings of negative pairs to prevent them from clustering excessively. Mathematically, as the number of negative samples $M \to \infty$, the normalized InfoNCE loss can be expressed as:

{\tiny
\begin{align*}
\lim_{M\rightarrow\infty}L(f,\tau)-\log M&=-\frac{1}{\tau}E_{\left(x, x^{+}\right)\sim p_{\text{pos}}}\left[f(x)^{T} f\left(x^{+}\right)\right]\\
&+E_{x\sim p_{\text{data}}}\left[\log E_{x^{-}\sim p_{\text{data}}^{-}}\left[e^{f(x)^{T} f\left(x^{-}\right)/\tau}\right]\right].
\end{align*}}

These properties make contrastive learning especially effective in tasks such as retrieval, where maximizing inter-sample variance is crucial. 

\subsubsection{Why Contrastive Learning Cannot Fully Address Length Collapse}
While contrastive learning alleviates high similarity issues across all text lengths, it does not entirely resolve the \textbf{Length Collapse} issue inherent in PLM-based models. Length Collapse arises from self-attention’s tendency to push embeddings for longer texts toward a concentrated representation space. This characteristic is unaffected by contrastive learning’s alignment or uniformity mechanisms because contrastive learning optimizes the relative positioning of positive and negative pairs rather than mitigating the length-induced clustering trend.

\subsection{Comparison with Other Post-processing Techniques}~\label{sec:post_processing}
In addition to TempScale, several post-processing methods have been proposed to address high similarity in embeddings, such as the Flow Function~\cite{li2020sentence} and Whitening~\cite{su2021whitening}. For comparative analysis, we implemented the last2avg version of the Flow Function, as the full flow version requires additional training, and the Whitening method is based on its original formulation. The results, summarized in Table~\ref{tab:post_processing}, highlight that these methods do not perform as effectively in our specific application. These results indicate that Flow and Whitening, originally developed for standard BERT embeddings, are less effective with the fine-tuned Transformer model used in this paper, which includes additional normalization layers. While TempScale reduces similarity in long-sequence embeddings, achieving distributional consistency across different sequence lengths is more critical for performance. Therefore, lowering similarity alone may not significantly improve downstream tasks.

\begin{table*}[h]
\centering
\resizebox{\textwidth}{!}{%
\begin{tabular}{lcccccccc}
\toprule
\textbf{Model} & \textbf{Rerank.} & \textbf{Summ.} & \textbf{Class.} & \textbf{Clust.} & \textbf{LongEmbdRetr.} & \textbf{STS} & \textbf{BeirRetr.} & \textbf{Avg.} \\ \midrule
\textbf{ANCE} & 49.09 & 29.58 & 55.27 & 33.04 & 34.02 & 66.32 & 36.87 & 43.45 \\ 
\textbf{+Flow} & 48.49 & 29.57 & 56.34 & 32.11 & 31.92 & 66.02 & 36.57 & 43.00 \\ 
\textbf{+Whitening} & 25.16 & 21.76 & 25.82 & 3.21 & 1.29 & 2.78 & 0.79 & 11.54 \\ \midrule
\textbf{GTR} & 54.23 & 29.67 & 55.10 & 38.65 & 37.33 & 70.11 & 44.98 & 47.15 \\ 
\textbf{+Flow} & 37.49 & 28.09 & 41.98 & 21.87 & 7.65 & 35.55 & 1.35 & 24.85 \\ 
\textbf{+Whitening} & 24.63 & 20.69 & 25.71 & 3.06 & 1.35 & -3.02 & 0.63 & 10.44 \\ \midrule
\textbf{GIST} & 58.55 & 31.14 & 64.75 & 44.77 & 38.21 & 75.61 & 52.77 & 52.26 \\ 
\textbf{+Flow} & 58.16 & 30.56 & 65.68 & 44.67 & 35.83 & 74.98 & 51.40 & 51.61 \\ 
\textbf{+Whitening} & 24.08 & 24.26 & 20.06 & 9.83 & 1.78 & -4.31 & 0.81 & 10.93 \\ \midrule
\textbf{BGE} & 58.87 & 31.03 & 64.79 & 45.80 & 37.46 & 75.88 & 55.29 & 52.73 \\ 
\textbf{+Flow} & 58.48 & 30.73 & 64.77 & 45.28 & 36.95 & 74.43 & 54.52 & 52.16 \\ 
\textbf{+Whitening} & 24.75 & 19.61 & 16.50 & 4.23 & 2.39 & 1.60 & 0.47 & 9.94 \\ \midrule
\textbf{E5-4K} & 53.12 & 30.58 & 61.72 & 41.01 & 56.01 & 71.77 & 47.22 & 51.63 \\ 
\textbf{+Flow} & 52.48 & 29.82 & 62.33 & 39.95 & 44.71 & 68.55 & 30.68 & 46.93 \\ 
\textbf{+Whitening} & 24.94 & 22.14 & 22.73 & 4.04 & 1.85 & 1.96 & 0.56 & 11.17 \\
\bottomrule
\end{tabular}}
\caption{Performance comparison of different post-processing methods across various tasks} \label{tab:post_processing}
\end{table*}

\subsection{Comparison with Other Similar Long-Text Methods}~\label{sec:long_method}
To thoroughly evaluate our TempScale approach, we compare it with two related attention scaling methods: $\text{softmax}\left(\frac{\text{logn}}{\tau \sqrt{d}} QK^T \right) V$ (Method1)~\cite{chiang2022overcoming} and YaRN~\cite{pengyarn}. In YaRN, the scaling formula for the attention matrix is given as $\text{softmax}\left(\frac{1}{\tau \sqrt{d}} QK^T \right) V$, where $\frac{1}{\tau} = 0.1 \ln{s} + 1$, and $s = \frac{L^{\prime}}{L}$, with $L^{\prime}$ representing the extended context window length and $L$ the original context window length. These methods propose different scaling strategies, but they differ in their theoretical motivations, applications, and experimental outcomes.

While Method1, YaRN, and TempScale share a similar structural approach, they tackle different challenges. Method1 is for binary classification tasks, specifically determining if the first string in a sequence of binary strings is a ``1''. It focuses on resolving straightforward decision-making problems in binary data. YaRN, by contrast, extends the context window of LLMs without requiring retraining, particularly adjusting the attention mechanism to handle larger contexts. TempScale addresses Length Collapse in embedding models for long texts, improving performance without retraining by preserving distinct representations as text length increases. Moreover, Method1 offers a solution using a specially designed Transformer example, but its motivation is limited by the specificity of this example. In contrast, YaRN introduces an empirically derived scaling formula. TempScale is based on a rigorous low-pass filtering analysis, giving it a strong theoretical foundation and an intuitive explanation.

\begin{table*}[h!]
\centering
\resizebox{\textwidth}{!}{
\begin{tabular}{l|cccccccc}
\toprule
\textbf{Model} & \textbf{Rerank.} & \textbf{Summ.} & \textbf{Class.} & \textbf{Clust.} & \textbf{LongEmbdRetr.} & \textbf{STS} & \textbf{BeirRetr.} & \textbf{Avg.} \\
\midrule
\textbf{ANCE} & 49.09 & 29.58 & 55.27 & 33.04 & 34.02 & 66.32 & 36.87 & 43.45 \\
\textbf{+Method1} & 41.80 & 29.21 & 48.21 & 20.72 & 6.23 & 53.40 & 7.74 & 29.61 \\
\textbf{+YaRN($\lambda=0.0001$)} & 49.09 & 29.58 & 55.62 & 33.01 & 34.02 & 66.32 & 36.86 & 43.50 \\
\textbf{+YaRN($\lambda=0.001$)} & 49.08 & 29.57 & 55.62 & 33.05 & 33.96 & 66.32 & 36.87 & 43.49 \\
\textbf{+YaRN($\lambda=0.01$)} & 49.10 & 29.31 & 55.65 & 32.95 & 33.92 & 66.27 & 36.87 & 43.44 \\
\textbf{+YaRN($\lambda=0.1$)} & 48.95 & 29.30 & 55.45 & 32.65 & 32.31 & 65.72 & 35.77 & 42.88 \\
\textbf{+YaRN($\lambda=1$}) & 45.71 & 29.28 & 52.80 & 26.16 & 10.59 & 60.15 & 13.34 & 34.00 \\ \midrule
\textbf{GTR} & 54.23 & 29.67 & 55.10 & 38.65 & 37.33 & 70.11 & 44.98 & 47.15 \\
\textbf{+Method1} & 38.12 & 28.44 & 40.34 & 14.46 & 3.61 & 47.09 & 0.82 & 24.70 \\
\textbf{+YaRN($\lambda=0.0001$)} & 54.23 & 29.68 & 55.06 & 38.30 & 37.42 & 70.11 & 44.98 & 47.11 \\
\textbf{+YaRN($\lambda=0.001$)} & 54.23 & 29.71 & 55.08 & 38.29 & 37.56 & 70.11 & 44.97 & 47.13 \\
\textbf{+YaRN($\lambda=0.01$)} & 54.23 & 29.71 & 55.04 & 38.63 & 37.39 & 70.06 & 45.00 & 47.15 \\
\textbf{+YaRN($\lambda=0.1$)} & 54.12 & 29.34 & 54.86 & 34.07 & 36.53 & 69.86 & 43.06 & 45.98 \\
\textbf{+YaRN($\lambda=1$)} & 55.19 & 29.00 & 47.27 & 19.06 & 3.99 & 61.87 & 1.46 & 31.12 \\ \midrule
\textbf{GIST} & 58.55 & 31.14 & 64.75 & 44.77 & 38.21 & 75.61 & 52.77 & 52.26 \\
\textbf{+Method1} & 58.64 & 28.26 & 51.05 & 28.11 & 5.42 & 62.43 & 7.51 & 34.49 \\
\textbf{+YaRN($\lambda=0.0001$)} & 58.55 & 31.14 & 64.26 & 44.75 & 38.21 & 75.61 & 52.78 & 52.19 \\
\textbf{+YaRN($\lambda=0.001$)} & 58.55 & 31.12 & 64.28 & 44.75 & 38.20 & 75.61 & 52.78 & 52.18 \\
\textbf{+YaRN($\lambda=0.01$)} & 58.53 & 31.26 & 64.24 & 44.78 & 38.05 & 75.60 & 52.79 & 52.18 \\
\textbf{+YaRN($\lambda=0.1$)} & 58.45 & 30.36 & 63.86 & 44.62 & 37.34 & 75.31 & 52.10 & 51.72 \\
\textbf{+YaRN($\lambda=1$)} & 55.75 & 26.76 & 58.79 & 36.95 & 11.85 & 69.00 & 23.59 & 40.38 \\ \midrule
\textbf{BGE} & 58.87 & 31.03 & 64.79 & 45.80 & 37.46 & 75.88 & 55.29 & 52.73 \\
\textbf{+Method1} & 37.78 & 29.17 & 37.80 & 14.36 & 2.10 & 43.72 & 0.84 & 23.68 \\
\textbf{+YaRN($\lambda=0.0001$)} & 58.86 & 31.04 & 64.78 & 45.77 & 37.45 & 75.88 & 55.29 & 52.72 \\
\textbf{+YaRN($\lambda=0.001$)} & 58.86 & 30.96 & 64.78 & 45.75 & 37.47 & 75.87 & 55.28 & 52.71 \\
\textbf{+YaRN($\lambda=0.01$)} & 58.85 & 31.02 & 64.73 & 45.61 & 37.26 & 75.86 & 55.22 & 52.65 \\
\textbf{+YaRN($\lambda=0.1$)} & 58.86 & 30.90 & 64.51 & 45.19 & 36.55 & 75.55 & 54.96 & 52.36 \\
\textbf{+YaRN($\lambda=1$)} & 52.65 & 29.09 & 50.51 & 25.50 & 2.36 & 64.20 & 2.14 & 32.35 \\ \midrule
\textbf{E5-4K} & 53.12 & 30.58 & 61.72 & 41.01 & 56.01 & 71.77 & 47.22 & 51.63 \\
\textbf{+Method1} & 40.90 & 24.11 & 42.85 & 13.64 & 3.24 & 41.62 & 1.05 & 23.92 \\
\textbf{+YaRN($\lambda=0.0001$)} & 53.12 & 30.57 & 61.78 & 40.77 & 56.01 & 71.77 & 47.22 & 51.61 \\
\textbf{+YaRN($\lambda=0.001$)} & 53.11 & 30.55 & 61.78 & 40.75 & 55.98 & 71.77 & 47.22 & 51.59 \\
\textbf{+YaRN($\lambda=0.01$)} & 53.07 & 30.42 & 61.73 & 40.61 & 55.53 & 71.71 & 47.21 & 51.47 \\
\textbf{+YaRN($\lambda=0.1$)} & 52.64 & 30.20 & 61.51 & 40.19 & 48.39 & 70.68 & 46.36 & 50.00 \\
\textbf{+YaRN($\lambda=1$)} & 45.00 & 29.01 & 50.81 & 17.99 & 2.91 & 57.27 & 1.46 & 29.21 \\
\bottomrule
\end{tabular}
}
\caption{Average main metric on MTEB and LongEmbd across Method1 and YaRN.}
\label{tab:model_compare}
\end{table*}

To effectively compare the performance of the above methods, we tested them on the MTEB benchmark. Since we are not modifying the context window length, we adapt YaRN as $\text{softmax}\left(\frac{\tau}{\sqrt{d}} QK^T \right) V$, with $\tau = \lambda \log n + 1$. We explore values for $\lambda$ in the range $\{0.0001, 0.001, 0.01, 0.1, 1\}$. This setup is reasonable, as our findings indicate that longer texts generally benefit from a smaller temperature scale. The experimental results are as shown in Table~\ref{tab:model_compare}.

The experimental results indicate that neither Method1 nor YaRN effectively adapts to the embedding model scenario. (Although YaRN shows slight improvement when the $\lambda$ value is small, in this case, Method 2 degenerates into TempScale.) Some potential reasons are as follows: A plausible explanation is that while both methods perform finer scaling on the attention matrix, applying different temperature adjustments across varying text lengths may lead to embeddings from different lengths falling into distinct distributions, which is unfavorable for downstream tasks. Moreover, in generation tasks, Method1 and YaRN succeed likely because of differences in output requirements. For these tasks, the model outputs a probability distribution and samples from it. Minor perturbations generally don’t affect the token output significantly; even if one sequence’s token distribution changes, it doesn't impact the output of other sequences. In contrast, any substantial change in a single embedding for an embedding model can directly affect the entire downstream performance. For instance, in classification tasks, a classifier model relies on embeddings as input, and in retrieval tasks, a change in embedding impacts document ranking. Overall, for embedding tasks, simply applying a single temperature adjustment better maintains the overall embedding distribution, helping mitigate Length Collapse and achieve better results across various downstream applications.

\section{Datasets and Evaluation Metrics} \label{sec:datasets}
Table \ref{tab:dataset-table} provides an overview of the datasets used in our experiments. Next, we give a brief description of the tasks involved in the experiments and the corresponding datasets and evaluation metrics they include.

\begin{table*}[t]
\centering
\resizebox{\textwidth}{!}{%
\begin{tabular}{l|l|c|c|c|c|c|c|c|c}
\toprule[1.5pt]
\textbf{Type} & \textbf{Name}                          & \textbf{Categ.} & \textbf{\#Lang.} & \textbf{\begin{tabular}[c]{@{}c@{}}Train\\ Samples\end{tabular}} & \textbf{\begin{tabular}[c]{@{}c@{}}Dev\\ Samples\end{tabular}} & \textbf{\begin{tabular}[c]{@{}c@{}}Test\\ Samples\end{tabular}} & \textbf{\begin{tabular}[c]{@{}c@{}}Train avg.\\ chars\end{tabular}} & \textbf{\begin{tabular}[c]{@{}c@{}}Dev avg.\\ chars\end{tabular}} & \textbf{\begin{tabular}[c]{@{}c@{}}Test avg.\\ chars\end{tabular}} \\ \midrule \midrule

\multirow{5}{*}{BEIR Retrival} & NFCorpus                               & s2p             & 1              & 0                                                                & 0                                                              & 3,956                                                            & 0                                                                   & 0                                                                 & 1,462.7                                                             \\
 & SciFact                                & s2p             & 1              & 0                                                                & 0                                                              & 5,483                                                            & 0                                                                   & 0                                                                 & 1,422.3                                                             \\ 
 
  & SciFact   & s2p  & 1 & 0  & 0  & 5,483 & 0 & 0   & 1,422.3   \\                                      & SCIDOCS   & s2p  & 1 & 0  & 0  & 26,657 & 0 & 0   & 1161.9   \\                   
  & FiQA2018   & s2p  & 1 & 0  & 0  & 58,286 & 0 & 0   & 760.4   \\   
  & Touche2020   & s2p  & 1 & 0  & 0  & 382,594 & 0 & 0   & 1720.1   \\   
 \midrule

\multirow{8}{*}{Classification} & AmazonPolarityClassification           & p2p             & 1              & 3,600,000                                                        & 0                                                              & 400,000                                                          & 431.6                                                               & 0                                                                 & 431.4                                                              \\
 & Banking77Classification                & s2s             & 1              & 10,003                                                           & 0                                                              & 3,080                                                            & 59.5                                                                & 0                                                                 & 54.2                                                               \\
 & EmotionClassification                  & s2s             & 1              & 16,000                                                           & 2,000                                                          & 2,000                                                            & 96.8                                                                & 95.3                                                              & 96.6                                                               \\
 & ImdbClassification                     & p2p             & 1              & 25,000                                                           & 0                                                              & 25,000                                                           & 1,325.1                                                              & 0                                                                 & 1,293.8                                                             \\
 & MassiveIntentClassification            & s2s             & 51             & 11,514                                                           & 2,033                                                          & 2,974                                                            & 35.0                                                                & 34.8                                                              & 34.6                                                               \\
 & MassiveScenarioClassification          & s2s             & 51             & 11,514                                                           & 2,033                                                          & 2,974                                                            & 35.0                                                                & 34.8                                                              & 34.6                                                               \\
 & ToxicConversationsClassification       & s2s             & 1              & 50,000                                                           & 0                                                              & 50,000                                                           & 298.8                                                               & 0                                                                 & 296.6                                                              \\
 & TweetSentimentExtractionClassification & s2s             & 1              & 27,481                                                           & 0                                                              & 3,534                                                            & 68.3                                                                & 0                                                                 & 67.8                                                               \\ \midrule

\multirow{11}{*}{Clustering} & ArxivClusteringP2P                     & p2p             & 1              & 0                                                                & 0                                                              & 732,723                                                          & 0                                                                   & 0                                                                 & 1,009.9                                                             \\
 & ArxivClusteringS2S                     & s2s             & 1              & 0                                                                & 0                                                              & 732,723                                                          & 0                                                                   & 0                                                                 & 74.0                                                               \\
 & BiorxivClusteringP2P                   & p2p             & 1              & 0                                                                & 0                                                              & 75,000                                                           & 0                                                                   & 0                                                                 & 1,666.2                                                             \\
 & BiorxivClusteringS2S                   & s2s             & 1              & 0                                                                & 0                                                              & 75,000                                                           & 0                                                                   & 0                                                                 & 101.6                                                              \\
 & MedrxivClusteringP2P                   & p2p             & 1              & 0                                                                & 0                                                              & 37,500                                                           & 0                                                                   & 0                                                                 & 1,981.2                                                             \\
 & MedrxivClusteringS2S                   & s2s             & 1              & 0                                                                & 0                                                              & 37,500                                                           & 0                                                                   & 0                                                                 & 114.7                                                              \\
 & RedditClustering                       & s2s             & 1              & 0                                                                & 420,464                                                        & 420,464                                                          & 0                                                                   & 64.7                                                              & 64.7                                                               \\
 & RedditClusteringP2P                    & p2p             & 1              & 0                                                                & 0                                                              & 459,399                                                          & 0                                                                   & 0                                                                 & 727.7                                                              \\
 & StackExchangeClustering                & s2s             & 1              & 0                                                                & 417,060                                                        & 373,850                                                          & 0                                                                   & 56.8                                                              & 57.0                                                               \\
 & StackExchangeClusteringP2P             & p2p             & 1              & 0                                                                & 0                                                              & 75,000                                                           & 0                                                                   & 0                                                                 & 1,090.7                                                             \\
 & TwentyNewsgroupsClustering             & s2s             & 1              & 0                                                                & 0                                                              & 59,545                                                           & 0                                                                   & 0                                                                 & 32.0                                                               \\ \midrule

\multirow{4}{*}{LongEmbd Retrival} & LEMBNarrativeQARetrieval               & s2p             & 1              & 0                                                                & 0                                                              & 10,804                                                           & 0                                                                   & 0                                                                 & 326,753.5                                                           \\
 & LEMBQMSumRetrieval                     & s2p             & 1              & 0                                                                & 0                                                              & 1,724                                                            & 0                                                                   & 0                                                                 & 53,335.8                                                            \\
 & LEMBSummScreenFDRetrieval              & s2p             & 1              & 0                                                                & 672                                                            & 0                                                               & 0                                                                   & 30,854.3                                                           & 0                                                                  \\
 & LEMBWikimQARetrieval                   & s2p             & 1              & 0                                                                & 0                                                              & 500                                                             & 0                                                                   & 0                                                                 & 37,445.6                                                            \\ \midrule

\multirow{4}{*}{Reranking} & AskUbuntuDupQuestions                  & s2s             & 1              & 0                                                                & 0                                                              & 2,255                                                            & 0                                                                   & 0                                                                 & 52.5                                                               \\
 & MindSmallReranking                     & s2s             & 1              & 231,530                                                          & 0                                                              & 107,968                                                          & 69.0                                                                & 0                                                                 & 70.9                                                               \\
 & SciDocsRR                              & s2s             & 1              & 0                                                                & 19,594                                                         & 19,599                                                           & 0                                                                   & 69.4                                                              & 69.0                                                               \\
 & StackOverflowDupQuestions              & s2s             & 1              & 23,018                                                           & 3,467                                                          & 3,467                                                            & 49.6                                                                & 49.8                                                              & 49.8                                                               \\ \midrule

\multirow{10}{*}{STS} & BIOSSES                                & s2s             & 1              & 200                                                              & 200                                                            & 200                                                             & 156.6                                                               & 156.6                                                             & 156.6                                                              \\
 & SICK-R                                 & s2s             & 1              & 19,854                                                           & 19,854                                                         & 19,854                                                           & 46.1                                                                & 46.1                                                              & 46.1                                                               \\
 & STS12                                  & s2s             & 1              & 4,468                                                            & 0                                                              & 6,216                                                            & 100.7                                                               & 0                                                                 & 64.7                                                               \\
 & STS13                                  & s2s             & 1              & 0                                                                & 0                                                              & 3,000                                                            & 0                                                                   & 0                                                                 & 54.0                                                               \\
 & STS14                                  & s2s             & 1              & 0                                                                & 0                                                              & 7,500                                                            & 0                                                                   & 0                                                                 & 54.3                                                               \\
 & STS15                                  & s2s             & 1              & 0                                                                & 0                                                              & 6,000                                                            & 0                                                                   & 0                                                                 & 57.7                                                               \\
 & STS16                                  & s2s             & 1              & 0                                                                & 0                                                              & 2,372                                                            & 0                                                                   & 0                                                                 & 65.3                                                               \\
 & STS17                                  & s2s             & 11             & 0                                                                & 0                                                              & 500                                                              & 0                                                                   & 0                                                                 & 43.3                                                               \\
 & STS22                                  & p2p             & 18             & 0                                                                & 0                                                              & 8,060                                                            & 0                                                                   & 0                                                                 & 1,992.8                                                             \\
 & STSBenchmark                           & s2s             & 1              & 11,498                                                           & 3,000                                                          & 2,758                                                            & 57.6                                                                & 64.0                                                              & 53.6                                                               \\ \midrule

\multirow{1}{*}{Summarization} & SummEval                               & p2p             & 1              & 0                                                                & 0                                                              & 2,800                                                            & 0                                                                   & 0                                                                 & 359.8                                                              \\ \bottomrule[1.5pt]
\end{tabular}%
}
\caption{Statistics of the experimental datasets used in the work.}~\label{tab:dataset-table}
\end{table*}

\begin{table*}[t]
\centering
\resizebox{0.8 \textwidth}{!}{%
\begin{tabular}{l|l}
\toprule
\textbf{Model Name}              & \multicolumn{1}{c}{\textbf{Publicly Available Link}}                                                       \\ \midrule \midrule
ANCE & {\color[HTML]{0563C1} https://huggingface.co/sentence-transformers/msmarco-roberta-base-ance-firstp} \\
GTR                      & {\color[HTML]{0563C1} https://huggingface.co/sentence-transformers/gtr-t5-base}                      \\

GIST          & {\color[HTML]{0563C1} https://huggingface.co/avsolatorio/GIST-small-Embedding-v0}                    \\
BGE                 & {\color[HTML]{0563C1} https://huggingface.co/BAAI/bge-base-en-v1.5}                                  \\
E5                       & {\color[HTML]{0563C1} https://huggingface.co/dwzhu/e5-base-4k}                                       \\
\bottomrule
\end{tabular}
}
\caption{Embedding models used in the experiments.}~\label{tab:model-table}
\end{table*}

\subsection{Classification}

In general, we use the provided embedding model to obtain a training set and a test set. The embeddings of the training set are used to train a logistic regression classifier with a maximum of 100 iterations, which is then scored on the test set. The main evaluation metrics are accuracy, average precision, and the F1 score.

\textbf{AmazonPolarity}~\cite{zhang2015character} consists of Amazon customer reviews, each labeled as either ``positive'' or ``negative.''

\textbf{Banking77} ~\cite{casanueva2020efficient} dataset consists of online banking user queries labeled with one of 77 specific intents.

\textbf{Emotion} ~\cite{saravia2018carer} comprises Twitter messages categorized by six fundamental emotions: anger, fear, joy, love, sadness, and surprise.

\textbf{IMDb} ~\cite{maas2011learning} consists of extensive movie reviews categorized as either positive or negative.

\textbf{MassiveIntent} ~\cite{fitzgerald2022massive} is a multilingual dataset featuring a diverse array of utterances from Amazon Alexa, each labeled with one of 60 different intents across 51 languages.

\textbf{MassiveScenario} ~\cite{fitzgerald2022massive} dataset comprises a diverse collection of Amazon Alexa user utterances, each labeled with one of 60 thematic intents, and supports 51 languages.

\textbf{ToxicConversations}~\footnote{\href{https://www.kaggle.com/competitions/jigsaw-unintended-bias-in-toxicity-classification/overview}{ToxicConversations}}
, sourced from a Kaggle competition, comprises comments from the Civil Comments platform, complete with annotations indicating whether each comment is toxic.

\textbf{TweetSentimentExtraction}~\footnote{\href{https://www.kaggle.com/competitions/tweet-sentiment-extraction/overview}{TweetSentimentExtraction}}, a dataset from a Kaggle competition, focuses on classifying tweets into three categories: neutral, positive, and negative sentiments.

\subsection{Clustering }

Clustering aims at grouping a given set of sentences or texts into meaningful clusters by training a mini-batch k-means model on the text embeddings. The model is scored using the v-measure ~\cite{rosenberg2007v}. Since the V-measure does not depend on the cluster labels, the arrangement of the labels will not affect the score.

\textbf{ArxivClusteringS2S, ArxivClusteringP2P, BiorxivClusteringS2S, BiorxivClusteringP2P, MedrxivClusteringP2P, MedrxivClusteringS2S}~\cite{muennighoff2023mteb}. These datasets are tailored for MTEB, utilizing titles or a combination of titles and abstracts from arXiv, bioRxiv, and medRxiv, with clustering labels derived from human-assigned categories, emphasizing both main and secondary classification levels.

\textbf{RedditClustering}~\cite{geigle2021tweac}, a dataset that consists of titles from 199 subreddits, is organized into 25 splits, each featuring 10 to 50 classes, with every class containing between 100 and 1,000 sentences.

\textbf{RedditClusteringP2P}~\cite{muennighoff2023mteb}, developed for the MTEB, consists of Reddit posts combined with their titles, organized into ten splits featuring 10 and 100 clusters each, with a total of 1,000 to 100,000 posts, aimed at clustering based on subreddit affiliation.

\textbf{StackExchangeClustering}~\cite{geigle2021tweac}, a dataset consisting of titles from 121 Stack Exchange communities, is organized into 25 subsets, each containing 10 to 50 categories, with 100 to 1,000 sentences per category.

\textbf{StackExchangeClusteringP2P}~\cite{muennighoff2023mteb}, designed for MTEB, comprises 10 splits of posts from StackExchange, each containing 5,000 to 10,000 entries, clustered by subreddit based on the combined content of titles and posts.

\textbf{TwentyNewsgroupsClustering}~\footnote{\url{https://scikit-learn.org/0.19/datasets/twenty_newsgroups.html}} consists of article titles from 20 different newsgroups, designed for clustering tasks, and includes 10 splits with each split featuring between 1,000 and 10,000 titles across the 20 categories.

\subsection{Reranking}

Reranking involves inputting a query along with a series of relevant and irrelevant reference texts, then sorting the results based on their relevance to the query. The provided model embeds the reference texts, which are compared to the query using cosine similarity. Each query is scored, and the average score across all queries is used to generate the final ranking. The evaluation metrics are MRR@k and MAP, with MAP serving as the primary metric.

\textbf{AskUbuntuDupQuestions}~\footnote{\url{https://github.com/taolei87/askubuntu}} dataset comprises questions sourced from AskUbuntu, accompanied by manually annotated labels that indicate whether pairs of questions are similar or dissimilar.

\textbf{MindSmall}~\cite{wu2020mind} dataset is a comprehensive English resource designed for research in news recommendation, focusing on ranking news articles based on the title of a currently read article to suggest related content.

\textbf{SciDocsRR}~\cite{cohan2020specter} is a dataset designed for ranking related scientific papers using their titles as the primary basis for assessment.

\textbf{StackOverflowDupQuestions}~\cite{liu2018linkso} dataset focuses on identifying whether questions tagged with Java, JavaScript, and Python on Stack Overflow are duplicates of existing queries.

\subsection{Retrieval}
In the retrieval task, each dataset consists of a corpus, queries, and a mapping of each query to relevant documents. The task goal is to find these relevant documents based on a given query. When evaluating, we first use the provided model to embed queries and corpus documents and then calculate the cosine similarity to obtain relevance scores and rank the corpus documents for each query based on these scores. The evaluation metrics consist of nDCG@k, MRR@k, MAP@k, precision@k, and recall@k, with nDCG@10 as the primary metric.

\subsubsection{Beir Retrival}
\textbf{NFCorpus}~\cite{boteva2016full} is a dataset that includes natural language queries sourced from NutritionFacts, paired with annotated medical documents from PubMed, utilizing the original splits from various types of content from NF, such as videos, blogs, and Q\&A posts.

\textbf{SciFact}~\cite{wadden2020fact} dataset evaluates scientific claims by matching them with evidence sourced from research literature, specifically utilizing a set of 300 test queries and the complete document collection from the original dataset.

\textbf{SCIDOCS}~\cite{specter2020cohan} is a benchmark dataset for evaluating scientific document embeddings, featuring seven tasks such as citation prediction, document classification, and recommendation.

\textbf{FiQA2018}~\cite{maia201818} is a financial question answering dataset built by crawling StackExchange posts under the Investment topic from 2009 to 2017, containing a knowledge base of 57,640 answer posts and 17,110 training question-answer pairs, with 531 testing question-answer pairs.

\textbf{Touche2020}~\cite{bondarenko2020overview} is a benchmark for argument retrieval, designed to support research in finding and evaluating arguments on various topics.

\subsubsection{LongEmbd Retrival}
LongEmbed~\cite{zhu2024longembed} includes 4 real-world retrieval tasks curated from long-form QA and summarization. The document in LongEmbed is much longer compared to BEIR. Thus, it can effectively evaluate the capability of the embedding model on long texts.

\textbf{LEMBNarrativeQARetrieval}~\cite{kovcisky2018narrativeqa} is a question-answering dataset featuring lengthy narratives, averaging 50,474 words, that challenge models to comprehend and extract information about characters and events dispersed throughout the stories.

\textbf{LEMBQMSumRetrieval}~\cite{zhong2021qmsum} dataset focuses on generating summaries of meetings based on specific queries, necessitating the extraction and synthesis of relevant information from various segments of the conversation that cover multiple topics and participants.

\textbf{LEMBSummScreenFDRetrieval}~\cite{chen2021summscreen} dataset consists of pairs of transcripts from TV series and their corresponding human-crafted summaries, requiring the integration of dispersed plot elements into concise narrative descriptions.

\textbf{LEMBWikimQARetrieval}~\cite{ho-etal-2020-constructing} dataset is a complex question-answering resource that includes questions requiring up to five reasoning steps, designed using specific templates to encourage deep understanding rather than simple retrieval of information.

\subsection{Semantic Textual Similarity (STS)}
The task goal is to determine the similarity between a pair of sentences, where continuous scores serve as labels, with higher values indicating greater similarity. The provided model embeds the sentences, and their similarity is calculated using cosine similarity. The primary evaluation metric is the Spearman correlation~\cite{reimers2016task}.

\textbf{STS12, STS13, STS14, STS15, STS16, STS17, STS22, STSBenchmark}~\cite{agirre2012semeval, agirre2013sem, bandhakavi2014generating, biccici2015rtm, nakov-etal-2016-semeval}~\footnote{\url{https://alt.qcri.org/semeval2017/task1/}}~\footnote{\url{https://competitions.codalab.org/competitions/33835}}~\footnote{\url{https://github.com/PhilipMay/stsb-multi-mt/}} are collections of sentence pairs designed to evaluate semantic textual similarity, with the former set focused on monolingual English pairs and the latter two incorporating cross-lingual comparisons across multiple languages.

\textbf{BIOSSES}~\cite{souganciouglu2017biosses} comprises 100 pairs of sentences specifically focused on the biomedical domain.

\textbf{SICK-R}~\cite{dadas-etal-2020-evaluation}, which stands for Sentences Involving Compositional Knowledge, comprises 100,000 diverse sentence pairs that exhibit rich lexical, syntactic, and semantic characteristics.

\subsection{Summarization}
 The input consists of a set of summaries written by humans and machines. The goal is to score the machine-generated summaries. Use the provided model to embed the summaries. Calculate the distance between each machine summary and all human summary embeddings. Retain similar scores to the model score for each individual machine-generated summary. Calculate the Spearman correlation based on cosine similarity~\cite{reimers2016task} as the main metric.

\textbf{SummEval}~\cite{fabbri2021summeval} consists of summaries produced by advanced summarization models trained on CNN and DailyMail articles.

\section{Embedding Models} \label{sec:models}
Table~\ref{tab:model-table} provides the models used in the experiments and their publicly available links. Below is a brief introduction to these models.

\textbf{ANCE}~\cite{xiong2021approximate} enhances dense retrieval by selecting challenging negative samples from the entire corpus and asynchronously updating the Approximate Nearest Neighbor (ANN) index with each training iteration, using a context window size of 512.

\textbf{GTR}~\cite{ni2022large} improves dual encoder performance for retrieval tasks by scaling up model size while keeping a fixed bottleneck embedding, leading to significant improvements in out-of-domain generalization, all within a context window size of 512.

\textbf{GIST}~\cite{solatorio2024gistembed} consistently improves performance across different model sizes by leveraging the strengths of large, resource-intensive models to enhance smaller ones, making advanced AI technologies more accessible and cost-effective, all within a context window size of 512.

\textbf{BGE}~\cite{xiao2023c} offers a range of well-trained embedding models based on a BERT-like architecture, enabling users to balance performance and efficiency for various applications while also allowing easy fine-tuning. In our experiments, we use the gte-base-en-v1.5 model, which operates with a context window size of 512.

\textbf{E5}~\cite{zhu2024longembed} is a long-context embedding model fine-tuned to support 4k token inputs while maintaining the original performance for shorter contexts, designed to advance research in long-context embedding technologies. It uses a context window size of 4k.

\end{document}